\documentclass[10pt,twocolumn,letterpaper]{article}

\usepackage{cvpr}              

\definecolor{cvprblue}{rgb}{0.21,0.49,0.74}
\usepackage[pagebackref,breaklinks,colorlinks,allcolors=cvprblue]{hyperref}
\usepackage{xfrac}

\def\paperID{12851} 
\def\confName{CVPR}
\def\confYear{2025}

\usepackage{caption}
\usepackage{subcaption}
\usepackage{graphicx}
\usepackage{amsmath}
\usepackage{amssymb}
\usepackage{booktabs}
\usepackage{subfiles}
\usepackage{tabularx}
\usepackage{pifont}
\usepackage{color, colortbl}
\usepackage{algpseudocode,algorithm,algorithmicx}
\usepackage[11pt]{moresize}
\usepackage{tikz}
\usepackage{comment}
\usepackage{flushend}
\usepackage{verbatim}
\usepackage{multirow}
\usepackage{rotating}
\usepackage{microtype}
\usepackage{ifthen}
\usepackage[normalem]{ulem}
\usepackage{epstopdf}
\usepackage{tikz}

\usepackage{arydshln} 
\usepackage{array} 
\usepackage{makecell} 

\algnewcommand\algorithmicforeach{\textbf{for each}}
\algdef{S}[FOR]{ForEach}[1]{\algorithmicforeach\ #1\ \algorithmicdo}

\algnewcommand\algorithmicnot{\textbf{not}}
\algdef{SE}[IF]{IfNot}{EndIf}[1]{\algorithmicif\ \algorithmicnot\ #1\ \algorithmicthen}{\algorithmicend\ \algorithmicif}%

\definecolor{myGreen}{HTML}{E7FEE9}
\definecolor{myBlue}{HTML}{DFE5EF}
\definecolor{myOrange}{HTML}{FEF4D2}
   
\newcommand{\textpartone}{{\color{red}En un lugar de la Mancha, de cuyo nombre no quiero acordarme, no ha mucho tiempo que vivía un hidalgo de los de lanza en astillero, adarga antigua, rocín flaco y galgo corredor ...}}
\newcommand{\textparttwo}{{\color{red} Una olla de algo más vaca que carnero, salpicón las más noches, duelos y quebrantos los sábados, lantejas los viernes, algún palomino de añadidura los domingos, consumían las tres partes de su hacienda ...}}
\newcommand{\textpartthree}{{\color{red} El resto della concluían sayo de velarte, calzas de velludo para las fiestas, con sus pantuflos de lo mesmo, y los días de entresemana se honraba con su vellorí de lo más fino ...}}

\newcommand{\textpartfour}{{\color{red} Tenía en su casa una ama que pasaba de los cuarenta y una sobrina que no llegaba a los veinte, y un mozo de campo y plaza que así ensillaba el rocín como tomaba la podadera ...}}

\newcounter{mycounter}
\newcommand{\enUnLugar}{
\protect\stepcounter{mycounter}
\ifcase \themycounter%
    \or \textpartone%
    \or \textparttwo%
    \or \textpartthree%
    \or \textpartfour \setcounter{mycounter}{0}
    \else Text ended%
\fi
}

\newcommand{\highlightedsubfigure}[4]{%
    \begin{tikzpicture}
        \node (img) {\begin{subfigure}[b]{#1}
            \centering
            \includegraphics[width=1.0\textwidth]{#2}
            \caption{#3}
        \end{subfigure}};
        \draw[#4,thick] (img.south west) rectangle (img.north east);
    \end{tikzpicture}
}

\newcolumntype{a}{:c|}
\newcolumntype{d}{:c}
\definecolor{tabfirst}{rgb}{0.4, 0.65, 0.3} 
\definecolor{tabsecond}{rgb}{0.7, 0.8, 0.65} 

\definecolor{nominal}{RGB}{27, 107, 162}
\definecolor{groundtruth}{RGB}{54, 161, 101}
\definecolor{repr_error}{RGB}{229, 126, 0}

\definecolor{proxy}{RGB}{187,136,187}

\newcommand{\ate}{\textcolor{groundtruth}{ATE\xspace}}
\newcommand{\gtfshort}{\textcolor{proxy}{GTF\xspace}}
\newcommand{\gtf}{\textcolor{proxy}{GTF ATE\xspace}}

\newcommand{\numDatasets}[0]{4\xspace}

\newcommand{\numSuccesfulExperimentsGlomap}[0]{14\xspace}  
\newcommand{\numExperimentsGlomap}[0]{15\xspace}  
\newcommand{\avgImprGtfGlomap}[0]{\textcolor{proxy}{26\%}}  
\newcommand{\avgImprGtGlomap}[0]{\textcolor{groundtruth}{32\%}}  

\newcommand{\numSuccesfulExperimentsDroid}[0]{9}  
\newcommand{\numExperimentsDroid}[0]{9}  
\newcommand{\avgImprGtfDroid}[0]{\textcolor{proxy}{6\%}}  
\newcommand{\avgImprGtDroid}[0]{\textcolor{groundtruth}{11\%}}
\title{Look Ma, No Ground Truth! \\ \textcolor{proxy}{G}round-\textcolor{proxy}{T}ruth-\textcolor{proxy}{F}ree Tuning of Structure from Motion and Visual SLAM}

\author{Alejandro Fontan$^{1,\dagger}$ , Javier Civera$^2$, Tobias Fischer$^1$ and Michael Milford$^1$\\
Queensland University of Technology$^1$, Universidad de Zaragoza$^2$\\
{\tt\small  $^\dagger$alejandro.fontan@qut.edu.au}
}

\begin{document}
\maketitle

\begin{abstract}
Evaluation is critical to both developing and tuning Structure from Motion (SfM) and Visual SLAM (VSLAM) systems, but is universally reliant on high-quality geometric ground truth -- a resource that is not only costly and time-intensive but, in many cases, entirely unobtainable. This dependency on ground truth restricts SfM and SLAM applications across diverse environments and limits scalability to real-world scenarios. In this work, we propose a novel ground-truth-free (\gtfshort) evaluation methodology that eliminates the need for geometric ground truth, instead using sensitivity estimation via sampling from both original and noisy versions of input images. Our approach shows strong correlation with traditional ground-truth-based benchmarks and supports \gtfshort\ hyperparameter tuning. Removing the need for ground truth opens up new opportunities to leverage a much larger number of dataset sources, and for self-supervised and online tuning, with the potential for a data-driven breakthrough analogous to what has occurred in generative AI.
\end{abstract}    
\section{Introduction}
\label{sec:introduction}

\begin{figure}[t]
\centering
\begin{subfigure}[b]{0.02\textwidth}
    \centering
     \rotatebox{90}{\parbox[t]{2.6cm}{\raggedleft \ate}}
\end{subfigure}
\hfill
\hspace{-0.8cm}
\begin{subfigure}[b]{0.38\textwidth}
    \centering
    \includegraphics[width=1.1\textwidth]{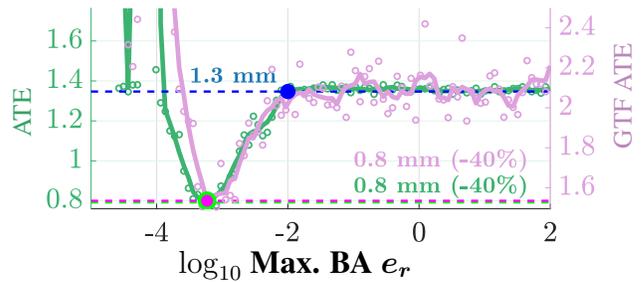}  
    \caption*{\large \textbf{$\log_{10}$ Max. BA $\boldsymbol{e_r}$}}
\end{subfigure}  
\hfill
\begin{subfigure}[b]{0.02\textwidth}
    \centering
    \rotatebox{90}{\parbox[t]{3.2cm}{\raggedleft \gtf}}
\end{subfigure}
\caption{\textbf{Illustration of ground-truth-free tuning in GLOMAP}. The \textcolor{groundtruth}{green} line fits \textcolor{groundtruth}{Absolute Trajectory Error (ATE)} results of GLOMAP as we vary one of its hyperparameters, specifically the maximum reprojection error for inliers in the Bundle Adjustment (Max. BA $e_r$) in radians. Note that, while the Max. BA $e_r$ \textcolor{nominal}{default value} in GLOMAP is \textcolor{nominal}{$10^{-2}$}, leading to an ATE of \textcolor{nominal}{$1.3$mm}, the optimal one for this particular sequence is \textcolor{groundtruth}{$\simeq 10^{-3}$}, for which the ATE improvement is \textcolor{groundtruth}{$\simeq 40\%$}, reaching \textcolor{groundtruth}{$0.8$mm}. \\\hspace{\textwidth} Now look at our proposed \textcolor{proxy}{GTF ATE} curve in \textcolor{proxy}{pink}, which \textcolor{proxy}{\textbf{without ground truth}}, is able to mimic the relative GLOMAP performance for different values of the hyperparameter, and hence also discerning its optimal setup.}
\label{fig:teaser_4}
\end{figure}

\newcommand{\atet}{\footnotesize ATE}
\newcommand{\rpet}{\footnotesize RPE}
\newcommand{\map}{\footnotesize Map.}
\newcommand{\qual}{\footnotesize Qua.}
\newcommand{\pho}{\footnotesize Pho.}

\begin{table}[ht]
\footnotesize
\centering
\setlength{\tabcolsep}{0.8pt} 
\rowcolors{2}{gray!15}{white} 
\resizebox{0.99\columnwidth}{!}{ 
\begin{tabular}{c c c c c c c c c c c c}
     & \rotatebox{90}{\footnotesize SVO~\cite{forster2014svo,forster2016svo}} & \rotatebox{90}{\cellcolor{white}\makecell{\footnotesize LSD / DSO / D3VO \\ \cellcolor{white} \footnotesize\cite{engel2014lsd,engel2016photometrically,yang2018challenges,yang2020d3vo}}} & \rotatebox{90}{\cellcolor{white}\makecell{\footnotesize ORB-SLAM \\ \cellcolor{white} \footnotesize\cite{mur2015orb,mur2017orb,campos2021orb}}} & \rotatebox{90}{\cellcolor{white}\makecell{\footnotesize Colmap / Glomap \\ \cellcolor{white} \footnotesize\cite{schoenberger2016sfm,pan2024global}}} & \rotatebox{90}{\footnotesize Tartan-VO~\cite{wang2021tartanvo}} & \rotatebox{90}{\footnotesize Droid-SLAM~\cite{teed2021droid}} & \rotatebox{90}{\cellcolor{white}\makecell{\footnotesize NICE / NICER \\ \cellcolor{white} \footnotesize SLAM~\cite{zhu2022nice,zhu2024nicer}}} & \rotatebox{90}{\footnotesize PVO~\cite{ye2023pvo}} & \rotatebox{90}{\footnotesize AnyFeature~\cite{fontanRSS2024}} &  
     \rotatebox{90}{\footnotesize MonoGS~\cite{matsuki2024gaussian}}  & \rotatebox{90}{\footnotesize DPV-SLAM~\cite{teed2024deep,lipson2025deep}}\\
    \midrule
    ICL-NUIM~\cite{handa2014benchmark}                      & \atet& \atet&  -   &  -   &   -  &  -   &  -   &  -   &  -   &  -   & \atet\\
    Replica~\cite{replica19arxiv}                           &   -  &   -  &  -   &  -   &   -  &  -   &\map  &  -   &  -   & \pho &   - \\
    VKITTI~\cite{Gaidon:Virtual:CVPR2016,cabon2020vkitti2}  &   -  &   -  &  -   &  -   &   -  &  -   &  -   & \atet&  -   &  -   &   - \\
    TartanAir~\cite{wang2020tartanair}                      &   -  &   -  &  -   &  -   & \atet& \atet&  -   &  -   &  -   &   -  & \atet\\
    Drunkards~\cite{recasens2023drunkard}                   &   -  &   -  &   -  &  -   &   -  &   -  &  -   &  -   &  -   &   -  &   - \\
    \midrule
    TUM-RGBD~\cite{sturm12iros}                             & \atet& \atet& \atet&   -  &   -  & \atet& \atet& \atet& \atet& \atet& \atet\\
    7-Scenes~\cite{shotton2013scene,glocker2013real}        &   -  &   -  &   -  &  -   &   -  &   -  & \atet&   -  &  -   &   -  &    - \\
    KITTI~\cite{geiger2013vision}                           &   -  & $t_r$& $t_r$&  -   & $t_r$&   -  &   -  & \atet& \atet&   -  & $t_r$\\
    EuRoC~\cite{burri2016euroc}                             & \atet& \atet& \atet&  -   & \atet& \atet&   -  &   -  & \atet&   -  & \atet\\
    ScanNet~\cite{dai2017scannet,yeshwanth2023scannet++}    &   -  &   -  &  -   &  -   &   -  &   -  & \atet&   -  &  -   &   -  &    - \\
    ETH3D~\cite{schops2017multi,schops2019bad}              &   -  &   -  & \atet& \atet&   -  & \atet&   -  &   -  & \atet&   -  &    - \\
    Rosario~\cite{pire2019rosario}                          &   -  &   -  & \atet&  -   &   -  &   -  &   -  &   -  &  -   &   -  &    - \\
    MADMAX~\cite{meyer2021madmax}                           &   -  &   -  & \atet&  -   &   -  &   -  &   -  &   -  &\qual &   -  &    - \\
    Lamar~\cite{sarlin2022lamar}                            &   -  &   -  &  -   & \atet&   -  &   -  &   -  &   -  &   -  &   -  &   -  \\
    Minimal Texture~\cite{fontan2023sid}                    &   -  &   -  & \atet&  -   &   -  &   -  &   -  &   -  &\qual &   -  &    - \\
    4Seasons~\cite{wenzel2020fourseasons,wenzel20244seasons}&   -  &   -  & \rpet&  -   &   -  &   -  &   -  &   -  &  -   &   -  & \rpet\\
    \midrule
    TUM Mono~\cite{engel2016photometrically}                &   -  & $e_a$& $e_a$&  -   &   -  &   -  &   -  &   -  &\qual &   -  &    - \\
    HAMLYN~\cite{recasens2021endo}                          &   -  &   -  &   -  &  -   &   -  &   -  &   -  &   -  &\qual &   -  &    - \\
    \bottomrule
\end{tabular}
}
\caption{\textbf{Benchmarking Metrics in SfM and VSLAM}$^\ddagger$. Despite substantial efforts towards diversity, current benchmarks still rely heavily on small, curated datasets, limiting the adaptability of localization pipelines to real-world scenarios. \textbf{Datasets} are listed in descending order as synthetic, real with ground truth, and real with pseudo-ground truth. \textbf{Metrics}: Absolute Trajectory Error (ATE) and Relative Pose Error (RPE)~\cite{sturm12iros}, translational and rotational errors ($t_r$)~\cite{geiger2013vision}, alignment error ($e_a$)~\cite{engel2016photometrically}, qualitative graph metrics (Qua.)~\cite{fontanRSS2024}, reconstruction metrics (Map.)~\cite{sucar2021imap}, and photometric rendering metrics (Pho.)~\cite{wang2004image,zhang2018unreasonable,matsuki2024gaussian}. $^\ddagger$\footnotesize Due to the intrinsic complexity of benchmarking SfM/VSLAM~\cite{zhang2018tutorial}, this table provides only a high-level overview. Please, refer to the respective publications for the full details.}
\label{tab:state-of-the-art}
\end{table}

Despite significant advances over the past decades, localization and 3D reconstruction from the images of a single moving camera still holds great potential for various downstream tasks, including, among others, view synthesis~\cite{tosi2024nerfs} and robotics~\cite{placed2023survey}. A critical long-term goal in this domain is achieving data scalability, which would unlock new applications and significantly enhance the performance of current systems. However, while cameras are inexpensive and easy to deploy, and access to vast video data is increasingly feasible~\cite{diba2020large,ahmadyan2021objectron,sener2022assembly101,grauman2022ego4d}, the field of visual localization—encompassing methods like Structure from Motion (SfM) and Visual SLAM (VSLAM)—has yet to achieve the same scalability and robustness breakthroughs seen in fields such as natural language processing~\cite{touvron2023llama} or generative AI~\cite{rombach2022high}. 

A major obstacle in advancing localization pipelines is the complexity of benchmarking tasks, such as hyperparameter optimization during development or performance comparison against existing solutions~\cite{zhang2018tutorial}. Accurate benchmarking requires objective evaluation against ground truth data, which serves as a crucial reference for assessing system performance~\cite{amigoni2007good,balaguer2007towards}. Moreover, real-world datasets are indispensable for meaningful benchmarking, as simulated data, while valuable for controlled experimentation, often falls short to capture the intricate complexities of real-world scenarios, including varying material properties, fine-grained structures, and dynamic reflections~\cite{replica19arxiv,wang2020tartanair,recasens2023drunkard}.

Unlike tasks such as object detection, tracking, or image segmentation—where ground truth is derived from large, human-annotated datasets~\cite{deng2009imagenet,lin2014microsoft,cordts2016cityscapes,dai2017scannet,yeshwanth2023scannet++}—localization pipelines require highly precise global positioning data. Outdoors, this typically involves sophisticated systems like RTK-GPS~\cite{geiger2013vision,wenzel2020fourseasons,agarwal2020ford,wenzel20244seasons}, while urban and indoor settings demand expensive and complex measurement setups~\cite{sturm12iros,burri2016euroc,schops2019bad,zhang2022hilti,helmberger2022hilti}. These challenges make the acquisition of ground truth data for localization resource-intensive and technically demanding~\cite{mallios2017underwater,meyer2021madmax}.

In some specialized domains, the difficulty of obtaining reliable ground truth is even more pronounced. Fields such as medical robotics~\cite{lamarca2020defslam,recasens2021endo,morlana2024topological}, extra-planetary exploration~\cite{meyer2021madmax,giubilato2022challenges}, and underwater robotics~\cite{humphries2023uncrewed,sauder2024self} operate in environments where ground truth data is either exceptionally difficult or impossible to obtain. Even when feasible, its collection often occurs under controlled conditions—such as overcast weather for underwater robotics—limiting the diversity of environmental scenarios.

As a consequence of all these challenges, and as shown in Table~\ref{tab:state-of-the-art}, current localization pipelines are frequently trained and evaluated on relatively small number of carefully curated datasets. This limitation constrains their scalability and hinders their ability to adapt to the diverse, unstructured conditions found in real-world applications~\cite{cadena2016past,davison2018futuremapping}. 

Unlike existing methods that rely heavily on expensive, carefully calibrated data, this paper addresses these challenges from a fresh perspective by proposing a novel \textcolor{proxy}{G}round-\textcolor{proxy}{T}ruth-\textcolor{proxy}{F}ree accuracy metric, \gtf, for evaluating SfM and VSLAM pipelines. Our approach assesses the precision of estimated camera trajectories by correlating them with sensitivity measurements derived from both original and noise-augmented input images.

\begin{figure}[t]
\centering
\begin{subfigure}[b]{0.48\textwidth}
    \centering
    \includegraphics[width=1.0\columnwidth]{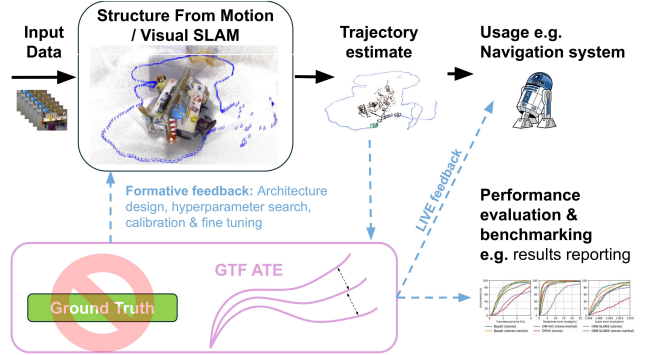}
\end{subfigure}
\caption{\textbf{Benchmarking Structure-from-Motion and Visual SLAM \emph{Without Ground Truth}.} The figure showcases the potential capabilities of our \textcolor{proxy}{G}round-\textcolor{proxy}{T}ruth-\textcolor{proxy}{F}ree \textcolor{proxy}{A}bsolute \textcolor{proxy}{T}rajectory \textcolor{proxy}{E}rror (\gtf), enabling formative feedback, live feedback, and comprehensive performance evaluation and benchmarking of SfM and VSLAM, all without relying on actual ground truth data.}
\label{fig:intro}
\end{figure}

The contributions of this work include an analytical formulation for precision comparison of linear systems using noise augmentation, the development of a comprehensive end-to-end, system-agnostic, and metric-agnostic evaluation methodology that \textit{eliminates the need for ground truth}, and, as shown in Figure~\ref{fig:teaser_4}, extensive experimental results demonstrating that our \textcolor{proxy}{ground-truth-free} metric strongly correlates with traditional \textcolor{groundtruth}{ground-truth-based} metrics across various datasets for tasks such as hyperparameter tuning. As depicted in Figure~\ref{fig:intro}, by reducing dependence on high-quality ground truth data, our method has the potential to significantly enhance the scalability of localization pipelines, paving the way for breakthroughs in real-world applications, akin to those seen in generative AI~\cite{rombach2022high}. 

\section{Related Work -- The Run for Benchmarking}
\label{sec:related}

Structure-from-Motion (SfM)~\cite{schoenberger2016sfm,schoenberger2016mvs,moulon2016openmvg,theia-manual,pan2024global,wang2024dust3r,duisterhof2024mast3r,zhang2024monst3r,wang20243d} aims at recovering the 3D structure of a scene from a typically sparse collection of images, as well as estimating their six-degrees-of-freedom camera poses. Visual SLAM (VSLAM)~\cite{davison2007monoslam,engel2014lsd,mur2015orb,schops2019bad,teed2021droid,ye2023pvo,fontanRSS2024,matsuki2024gaussian,teed2024deep,lipson2025deep}, a ``sister'' field, typically differs on having video input instead of temporally sparse images and targeting real-time and online processing. In both fields, the availability of standardized datasets and the selection of appropriate metrics have been instrumental in advancing the state of the art and developing effective and accurate pipelines~\cite{bonarini2006rawseeds,mountney2010three,pomerleau2011tracking,bao2011semantic,sturm12iros,geiger2013vision,glocker2013real,handa2014benchmark,Choi_2015_CVPR,engel2016photometrically,burri2016euroc,Gaidon:Virtual:CVPR2016,mallios2017underwater,maddern20171,schops2017multi,dai2017scannet,judd2018multimotion,judd2019oxford,canoeDataset,schops2019bad,pire2019rosario,replica19arxiv,cabon2020vkitti2,wenzel2020fourseasons,wang2020tartanair,meyer2021madmax,recasens2021endo,zhang2022hilti,helmberger2022hilti,sarlin2022lamar,barron2022mip,yeshwanth2023scannet++,fontan2023sid,singh2023online,recasens2023drunkard,wenzel20244seasons}. As SfM and VSLAM have evolved, the metrics used to evaluate them have adapted, reflecting the increasing complexity and scale of modern systems.

\subsection{\textbf{\fontsize{10.22pt}{16pt}\selectfont Ground Truth-less Benchmarking for SfM/VSLAM}}
Due to the mentioned difficulties in achieving large and realistic datasets with accurate ground truth, the reprojection error has been used several times as a metric, \emph{e.g.}, by \citet{schoenberger2016sfm}. Using the optimization goal as a metric, however, is not in general good practice, as it could be overfitted. Other works have used the $\chi^2$ or Mahalanobis error~\cite{kummerle2011g,olson2009evaluating}, that measures the consistency between the estimated errors and uncertainties, but not the errors' magnitude. \citet{recasens2023drunkard} proposed the Absolute Palindrome Trajectory Error, consisting a forward and backward passes through the image sequence. Such metric, however, is only valid for visual odometry and not for SfM/VSLAM and may also be affected by the well-known motion bias~\cite{yang2018challenges,fontan2023motion}.

\subsection{SfM/VSLAM Metrics With Ground Truth}

In urban environments, Wulf \etal~\cite{wulf2007ground} quantified errors between 3D scans and \textbf{reference maps} using Euclidean distance and angular differences. The ground-truth reference maps were obtained from highly accurate CAD data.
To address the limitations of global reference frames, Burgard \etal~\cite{burgard2009comparison,kummerle2009measuring} compared \textbf{relative displacements} between poses estimated by graph-based SLAM with \textit{true relative displacements}, obtained through manual matching of laser-range observations with the background knowledge of an expert familiar with the environment's topology.

Sturm \etal \cite{sturm12iros} introduced a benchmark for evaluating RGB-D SLAM using two key metrics: \textbf{Relative Pose Error (RPE)} and \textbf{Absolute Trajectory Error (ATE)}. RPE measures local accuracy by comparing estimated and true motion over fixed intervals, effectively assessing odometric drift \cite{kerl2013robust} and loop closure accuracy in VSLAM \cite{kummerle2009measuring,burgard2009comparison}. ATE evaluates global consistency by aligning estimated and ground truth trajectories \cite{horn1987closed,umeyama1991least} and measuring translational differences, for a more comprehensive assessment of long-term consistency. Both metrics have become standard in SLAM benchmarking \cite{engel2017direct,mur2015orb,forster2014svo}, enabling rigorous comparisons by relying on a highly precise, carefully calibrated, time-synchronized ground truth. Recently, \citet{lee2022s,lee2024alignment} have proposed robust variations of such metrics.

Zhang \etal \cite{zhang2018tutorial} presented a comprehensive tutorial on evaluating the quality of estimated trajectories based on specific sensing modalities (\eg, monocular, stereo, and visual-inertial). Their work analyzed the impact of various alignment methods and error metrics, primarily ATE and RPE, in relation to ground truth data. Building on this, Zhang \etal \cite{zhang2019rethinking} introduced a probabilistic, continuous-time framework for trajectory evaluation. By leveraging Gaussian processes as the underlying representation, they formulated estimation errors probabilistically, providing a theoretical link between relative and absolute error metrics and addressing temporal association in a principled way.

Geiger \etal \cite{geiger2012we} introduced the KITTI benchmark for visual odometry and SLAM, capturing data from a multi-sensor car platform driving through diverse environments such as city streets, rural areas, and highways. Ground truth poses were obtained from a localization system integrating GPS, IMU, and RTK correction signals, all precisely calibrated and synchronized with cameras and a laser scanner. They proposed separate metrics for \textbf{translational} \mbox{$t_r$ [\%]} and \textbf{rotational} $r_r$ [deg/m] \textbf{errors}, considering trajectory length and velocity. The benchmark’s large scale and novel metrics evaluated error statistics over all sub-sequences of a given trajectory length or driving speed, providing deeper insights into failure modes and setting a new standard for fairer comparisons across visual odometry and SLAM methods.

Engel \etal \cite{engel2016photometrically, engel2017direct} introduced the TUM monoVO dataset, featuring photometrically calibrated sequences recorded in various indoor and outdoor environments. The dataset emphasizes camera motion with a large loop-closure at the end of each sequence, enabling the evaluation of accumulated drift without requiring full ground truth poses. Visual odometry (VO) accuracy is assessed using the \textbf{alignment error} $e_{align}$, which measures the drift over the entire sequence. While Engel \etal demonstrated that pre-loop-closure drift is a strong indicator of system accuracy, loop-closure detection in full SLAM systems \cite{mur2017orb,engel2014lsd} must be disabled for valid evaluation. As a result, SLAM-specific challenges such as re-localization, map correction, and long-term map maintenance are not addressed, and failure modes during the sequence cannot be fully captured.

\subsection{Map Metrics}
Camera trajectory errors are mostly evaluated in SfM and VSLAM, due to the challenge of acquiring ground truth scene geometry. However, recent advancements in dense 3D reconstruction~\cite{sucar2021imap,Zhi:etal:arxiv2021,zhu2022nice,sandstrom2023point,anonymous2024implicit,deng2023nerfloam,li2023dense,gangopadhyay2024uncle,matsuki2024gaussian,hua2024hi,zhu2024nicer,li2024ddn} underscore the necessity of comprehensive map evaluation. 

Sucar \etal~\cite{sucar2021imap} evaluated their scene reconstruction by comparing ground-truth and reconstructed meshes using three metrics: \textbf{Accuracy} [cm], the average distance from reconstructed points to ground truth; \textbf{Completion} [cm], the average distance from ground-truth points to the reconstruction; and \textbf{Completion Ratio} [$<$5cm \%], the percentage of reconstructed points within 5 cm of the ground truth.

Matsuki \etal~\cite{matsuki2024gaussian} assessed the map quality of their monocular Gaussian Splatting SLAM using standard photometric rendering metrics: \textbf{Peak Signal-to-Noise Ratio (PSNR [dB])}, \textbf{Structural Similarity Index (SSIM)~\cite{wang2004image}}, and \textbf{Learned Perceptual Image Patch Similarity (LPIPS)}~\cite{zhang2018unreasonable}.


\section{Why is ground truth not necessary?}
\label{sec:formulation}

Similar to Kümmerle \etal~\cite{kummerle2009measuring} who argued that ``\textit{meaningful comparisons between different SLAM approaches require a common metric}", we propose that new metrics must support scalability to self-supervised or unsupervised training of SfM and VSLAM pipelines to foster generalization and robustness. Differently from all previous works mentioned above, and for the first time, we introduce a ground-truth-free metric, \textcolor{proxy}{GTF-ATE}, for evaluating the end-to-end performance of SfM/VSLAM systems, offering accuracy comparable to state-of-the-art ground-truth-based methods.

\subsection{The Jacobians model the sensitivity to noise}
\label{sec:jacobians}

Let us define the SfM/VSLAM state, containing the camera poses and 3D points' parameters, as $\boldsymbol{x} \in \mathcal{S}$, where $\mathcal{S}$ refers to a manifold due to camera rotations belonging to $SO(3)$. Its covariance matrix can be defined in the tangent space~\cite[Chapter~7.3]{barfoot2024state} as $\Sigma_{\boldsymbol{x}} \in \mathbb{S}^{n}_+$, where $n=6c+3d$, $c$ is the number of images and $d$ is the number of reconstructed 3D points. $\Sigma_{\boldsymbol{x}}$ can be approximated by a first-order propagation of the measurement covariance $\Sigma_{\boldsymbol{z}} \in \mathbb{S}^{m}_+$--$m$ standing for the total measurement vector size:
\begin{equation}
    \Lambda_{\boldsymbol{x}} \equiv \Sigma_{\boldsymbol{x}}^{-1} \simeq J^\top \Sigma_{\boldsymbol{z}}^{-1} J,
\label{eq:inf_matrix}
\end{equation}
where $J = \sfrac{\partial h(\boldsymbol{x})}{\partial \boldsymbol{x}} \in \mathbb{R}^{m \times n} $ is the Jacobian of the projection model $h(\boldsymbol{x})$, $\boldsymbol{z} = h(\boldsymbol{x}) + \boldsymbol{\epsilon} \in \mathbb{R}^m$ is the measurement vector for which we assume additive zero-mean Gaussian noise $\boldsymbol{\epsilon} \sim \mathcal{N}\left( \boldsymbol{0}, \Sigma_{\boldsymbol{z}} \right)$, and $\Lambda_{\boldsymbol{x}} \in \mathbb{R}^{n\times n}$ is the information matrix of the state. 

The expected variance improvement between two SfM/VSLAM setups with different hyperparameter sets, denoted as $p$ and $q$, can be quantified in terms of \textit{entropy reduction} as:
\begin{equation}
   E(p,q) = E(p) - E(q)= \frac{1}{2}\log_2\Big(\frac{|\Lambda^q_{\boldsymbol{x}}|}{|\Lambda^p_{\boldsymbol{x}}|}\Big),
\end{equation}

\noindent where $|{\cdot}|$ stands for the determinant of a matrix, and $E(p,q) \in \mathbb{R}$ represents how much information, in bits, is gained by using the setup $q$ instead of $p$, which in turn results in smaller expected errors. In simpler terms, the greater $|\Lambda^q_{\boldsymbol{x}}|$ is, compared to $|\Lambda^p_{\boldsymbol{x}}|$, the more accurate the setup $q$ is expected to be with respect to the setup $p$. 

A common assumption is that the measurement noise is isotropic, \ie, $\Sigma_{\boldsymbol{z}} = \sigma^2 I_m$, where $\sigma^2 \in \mathbb{R}_{>0}$ is the measurement noise variance and $I_m$ is the identity matrix of size $m$. This allows us to simplify the determinant of the information matrices as follows:
\begin{equation}
    |\Lambda_{\boldsymbol{x}}| \simeq \frac{1}{\sigma^{2n}}|J^\top J|.
    \label{eq:_simp_inf_matrix}
\end{equation}

From Eq.~\eqref{eq:_simp_inf_matrix}, and for the same variance $\sigma^2$ in setups $p$ and $q$, the one with higher expected accuracy is the one with the larger Jacobian’s Gram matrix determinant $|J^\top J|$:
\begin{equation}
    |J_q^\top J_q| > |J_p^\top J_p| \iff |\Lambda_{\boldsymbol{x}}^q| > |\Lambda_{\boldsymbol{x}}^p|  \iff E(p,q) > 0.
    \label{eq:accuracy_comp}
\end{equation}

The reader will have a more intuitive view of the above in a toy 1D linear example. Given two linear setups $z_p = px + \epsilon$ and $z_q = qx + \epsilon$ affected by same noise distribution $\epsilon \sim \mathcal{N}(0,\sigma^2)$, their respective information scalars $\Lambda^p_x = \left( \sfrac{p}{\sigma} \right)^2$ and $\Lambda^q_x = \left( \sfrac{q}{\sigma} \right)^2$ depend directly on their derivatives, and hence $q^2 > p^2 \iff \Lambda_q > \Lambda_p \iff E(p,q) > 0$. In words, for the same measurement noise distribution, the setup with the bigger derivative will lead to higher entropy reductions and then have smaller state variance.

\subsection{Sensitivity Sampling}
\label{sec:sampling}

From our derivations in the previous section, and in particular Eq.~\eqref{eq:accuracy_comp}, it follows that the relative accuracy of two SfM/SLAM pipelines could be assessed, in principle, by analytically computing $|J_q^\top J_q|$ and $|J_p^\top J_p|$. However, state errors in the estimation of $\boldsymbol{x}$ will be amplified by the derivatives, which will pose challenges in practice. Instead, we base our approach on sampling ground-truth-free versions of the metrics for the original and noisy augmentations of the data, from which we can estimate smoothed versions of the sensitivity.

Crucially for our purposes, note that the formulation in Section~\ref{sec:jacobians} still holds for functions $\phi( 
\cdot )$ of the above problems. For convenience, we define $\boldsymbol{x}_\boxminus = \phi_\boxminus(  \boldsymbol{x}, \boldsymbol{x}_\Delta ) = \boldsymbol{x}_\Delta \boxminus \boldsymbol{x} \in \mathbb{R}^n$, where $\boxminus$ is used as a generalization of the minus sign for a generic manifold, $\boldsymbol{x}$ is estimated from a set of measurements $\boldsymbol{z} \sim \mathcal{N}(\boldsymbol{0}, \sigma^2 I_m)$ and $\boldsymbol{x}_\Delta$ from a set of measurements with added variance $\boldsymbol{z}_\Delta \sim \mathcal{N}(\boldsymbol{0}, (\sigma^2 + \Delta \sigma^2) I_m)$. For small added variance $\Delta \sigma^2$, we can again approximate its covariance as

\begin{equation}
    |\Lambda_{ \boldsymbol{x}_\boxminus}| \simeq |\Lambda_{\boldsymbol{x}_\Delta} + \Lambda_{\boldsymbol{x}}| = \frac{1}{({2\sigma}^{2} + \Delta \sigma^2)^{n}}|J^\top J| 
\end{equation}


\noindent and, similarly to Eq.~\ref{eq:accuracy_comp}, for same measurement variances $\sigma^2$ and ${\Delta \sigma}^{2}$ in setups $p$ and $q$

\begin{equation}
    |J_q^\top J_q| > |J_p^\top J_p| \iff |\Lambda_{ \boldsymbol{x}_\boxminus}^q| > |\Lambda_{ \boldsymbol{x}_\boxminus}^p|  \iff E(p,q) > 0.
    \label{eq:final}
\end{equation}


As a summary, our derivations in this subsection leads to conclude that the relative performance of two SfM/VSLAM setups $p$ and $q$ (\ie, which one is better) can be assessed \emph{without ground truth} by comparing their relative degradation when data is perturbed for both cases with additional variance ${\Delta \sigma}^{2}$. In order to smooth noisy estimates of this degradation, we sample $k_\Delta$ noise instances with variance ${\Delta \sigma}^{2}$ and average them.

\subsection{The linearity assumption}
\label{sec:SfMlinearity}

\begin{figure}[t]
    \centering
    \begin{subfigure}[b]{0.012\textwidth}
        \centering
        \rotatebox{90}{\parbox[t]{2.3cm}{\raggedleft \textcolor{groundtruth}{\large \ate}}}
    \end{subfigure}
    \begin{subfigure}[b]{0.32\textwidth}
         \centering
         \includegraphics[width=1.0\textwidth]{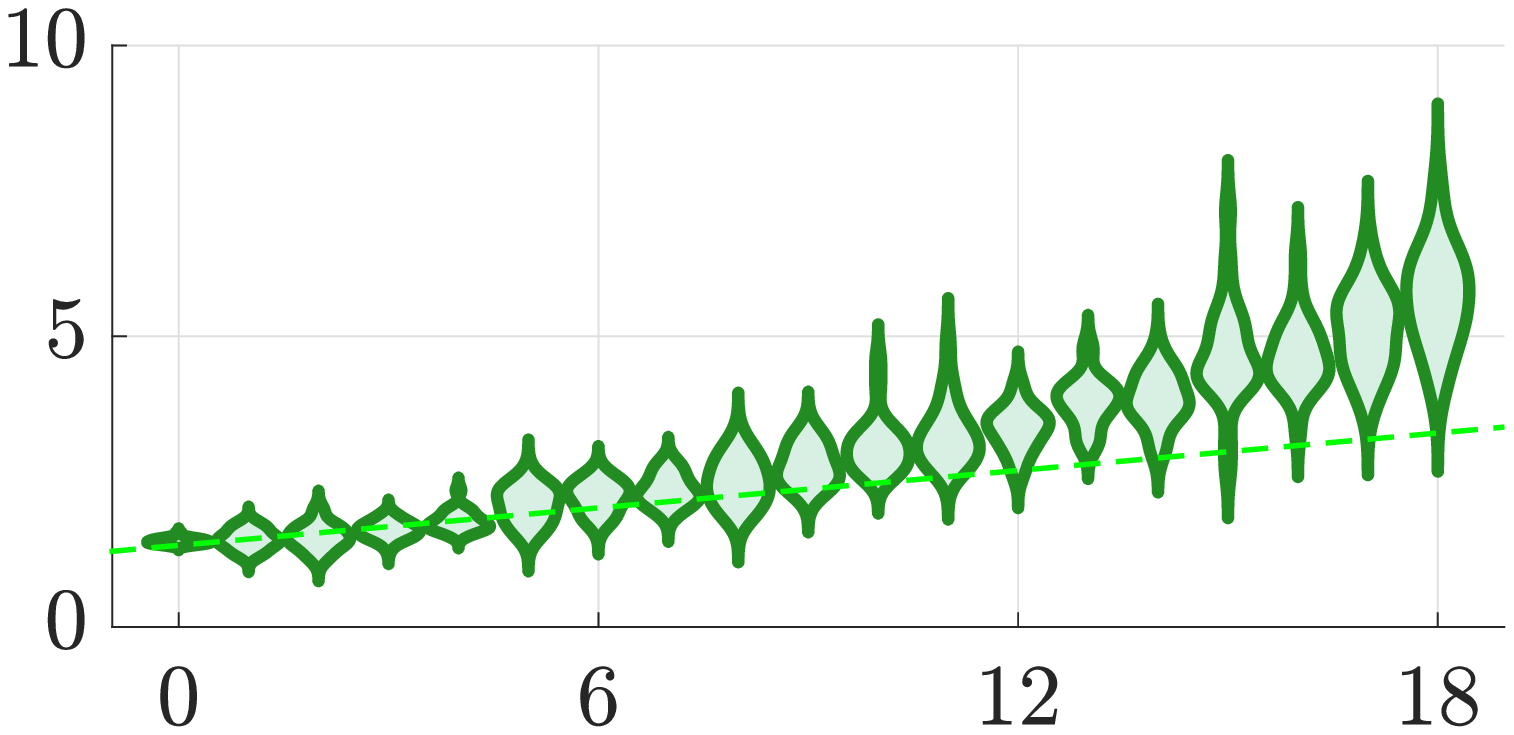} 
         \caption*{\large $\Delta\sigma$}
    \end{subfigure}
    \caption{\textbf{Experimental assessment of GLOMAP's linearity}. Our ground-truth-free tuning assumes a high degree of linearity in SfM/VSLAM pipelines. To assess this hypothesis, we run GLOMAP~\cite{pan2024global} $k_\Delta$ times for images perturbed with noises of different variances $\Delta \sigma$. Note, in the fit to the first values that we draw, how the ATE shows a high degree of linearity in its evolution.}
    \label{fig:linearity}
\end{figure}

The assumption of a high degree of linearity of SfM/VSLAM pipelines is at the core of our discussion here and the practical methodology of next section. We experimentally assessed the goodness of this assumption by running GLOMAP~\cite{pan2024global} over a set of images, each of them perturbed with Gaussian noise of variance $\Delta \sigma^2$, and those for different variance values. The results, aggregated in Figure~\ref{fig:linearity}, show a high degree of linearity. 

\subsection{\textcolor{proxy}{G}round-\textcolor{proxy}{T}ruth-\textcolor{proxy}{F}ree \textcolor{proxy}{A}bsolute \textcolor{proxy}{T}rajectory \textcolor{proxy}{E}rror}
\label{sec:methodology}

Algorithm~\ref{alg:gtf} outlines our methodology for computing the Ground-Truth-Free Absolute Trajectory Error (\textcolor{proxy}{GTF ATE}). Given a SfM or VSLAM pipeline \(H_p\), with a hyperparameter set \(p\), we first run the system \(k\) times on the raw input images \(\textit{I}\) to obtain \(k\) trajectory estimates $T = \{ \boldsymbol{t}_1, \hdots, \boldsymbol{t}_k \}$, each of the trajectories composed by the rotation and translation for the $c$ images $\boldsymbol{t}_{i \in \{ 1, \hdots, k \}} \in \mathbb{R}^{3c}$.

Next, we run the system \(k_\Delta\) additional times, each time augmenting the raw images with independent Gaussian noise, \ie the new images being \(I_{\Delta} \gets I + \mathcal{N}(\boldsymbol{0}, \Delta\sigma)\). This process produces \(k_\Delta\) noisy trajectory estimates $T_\Delta = \{ \boldsymbol{t}_{\Delta, 1}, \hdots, \boldsymbol{t}_{\Delta, k} \}$, with $\boldsymbol{t}_{\Delta, j \in \{ 1, \hdots, k_\Delta \}} \in \mathbb{R}^{3c}$.

For this work, we focus on monocular setups for both SfM and VSLAM. Consequently, we define \(\phi_\text{ATE}\) as the Absolute Trajectory Error (ATE)~\cite{sturm12iros,zhang2018tutorial} function. ATE measures the discrepancy between two trajectories (\eg, $\boldsymbol{t}_i$ and $\boldsymbol{t}_{\Delta, j}$) by first aligning them via a \(Sim(3)\) transformation and then computing the root mean squared error (RMSE) over all tuples of the translational component.

The \textcolor{proxy}{GTF ATE} precision metric is derived by averaging the ATE values across all trajectory comparisons:
\begin{equation}
    \text{\textcolor{proxy}{GTF ATE}} = \frac{1}{k \cdot k_{\Delta}}\sum_{i=1}^k \sum_{j=1}^{k_\Delta}\phi_{\text{ATE}}( \boldsymbol{t}_{i}, \boldsymbol{t}_{\Delta, j}).
    \label{eq:metric}
\end{equation}

\begin{algorithm}[t]
\caption{Compute \textcolor{proxy}{G}round-\textcolor{proxy}{T}ruth-\textcolor{proxy}{F}ree \textcolor{proxy}{ATE}} 
\begin{algorithmic}[1] 
\Function{\textcolor{proxy}{GTF ATE} }{$H_p$, \textit{I}}

    \Comment{$H_p$: SfM/VSLAM with hyperparameter conf.~$p$}

    \Comment{\textit{I}: Grayscale images}

    \For{$i = 1$ \textbf{to} $k$} \Comment{Execution step}
        \State $\boldsymbol{t}_i \gets H_p(\textit{I})$, \quad  $T \gets T \cup \boldsymbol{t}_i $
    \EndFor 

    \For{$j = 1$ \textbf{to} $k_\Delta$} \Comment{Perturbation step}
        \State $I_\Delta \gets I + \mathcal{N}(\boldsymbol{0}, \Delta\sigma)$
        \State $\boldsymbol{t}_{\Delta,j} \gets H_p(I_\Delta)$, \quad  $T_\Delta \gets T_\Delta \cup \boldsymbol{t}_{\Delta,j}$
    \EndFor 
    \For{$\boldsymbol{t}_i$ \textbf{in} $T$} \Comment{Evaluation step}
        \For{$\boldsymbol{t}_{\Delta,j}$ \textbf{in} $T_\Delta$} \Comment{$\phi_{\text{ATE}}$: ATE operator}
            \State $\text{ATE}_{i,j} = \phi_{\text{ATE}}(  \boldsymbol{t}_i, \boldsymbol{t}_{\Delta,j} )$
            \State $\text{ATE}_{\text{all}} \gets \text{ATE}_{\text{all}} \cup \text{ATE}_{i,j}$
        \EndFor 
    \EndFor 
    
    \State \Return \textcolor{proxy}{GTF ATE} $\gets \text{mean}(\text{ATE}_{\text{all}})$
\EndFunction
\end{algorithmic}
\label{alg:gtf}
\end{algorithm}
\section{Experiments}
\label{sec:experiments}

\begin{figure*}[ht]
\begin{subfigure}[b]{1.0\textwidth}
    \centering
    \captionsetup{position=above} 
    \caption*{\textbf{Dataset} : Replica -- \textbf{Sequence} : Office 0}
    \vspace{-0.2cm}
    \begin{subfigure}[b]{0.01\textwidth}
        \centering
        \rotatebox{90}{\parbox{1.7cm}{\centering  \footnotesize \ate}}
    \end{subfigure}
    \hfill
    \begin{subfigure}[b]{0.19\textwidth}
         \centering
         \includegraphics[width=1.00\textwidth]{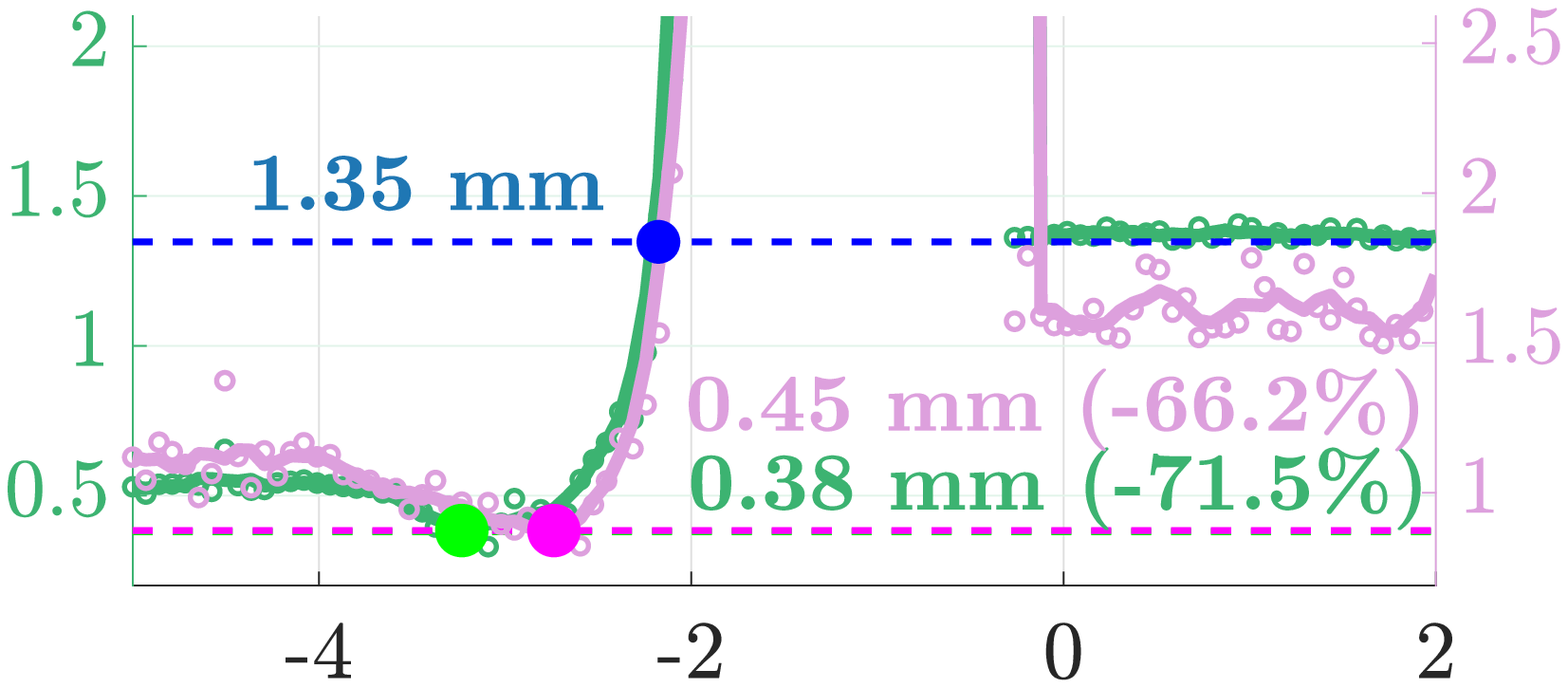}
    \end{subfigure}
    \hfill
    \begin{subfigure}[b]{0.19\textwidth}
         \centering
         \includegraphics[width=1.00\textwidth]{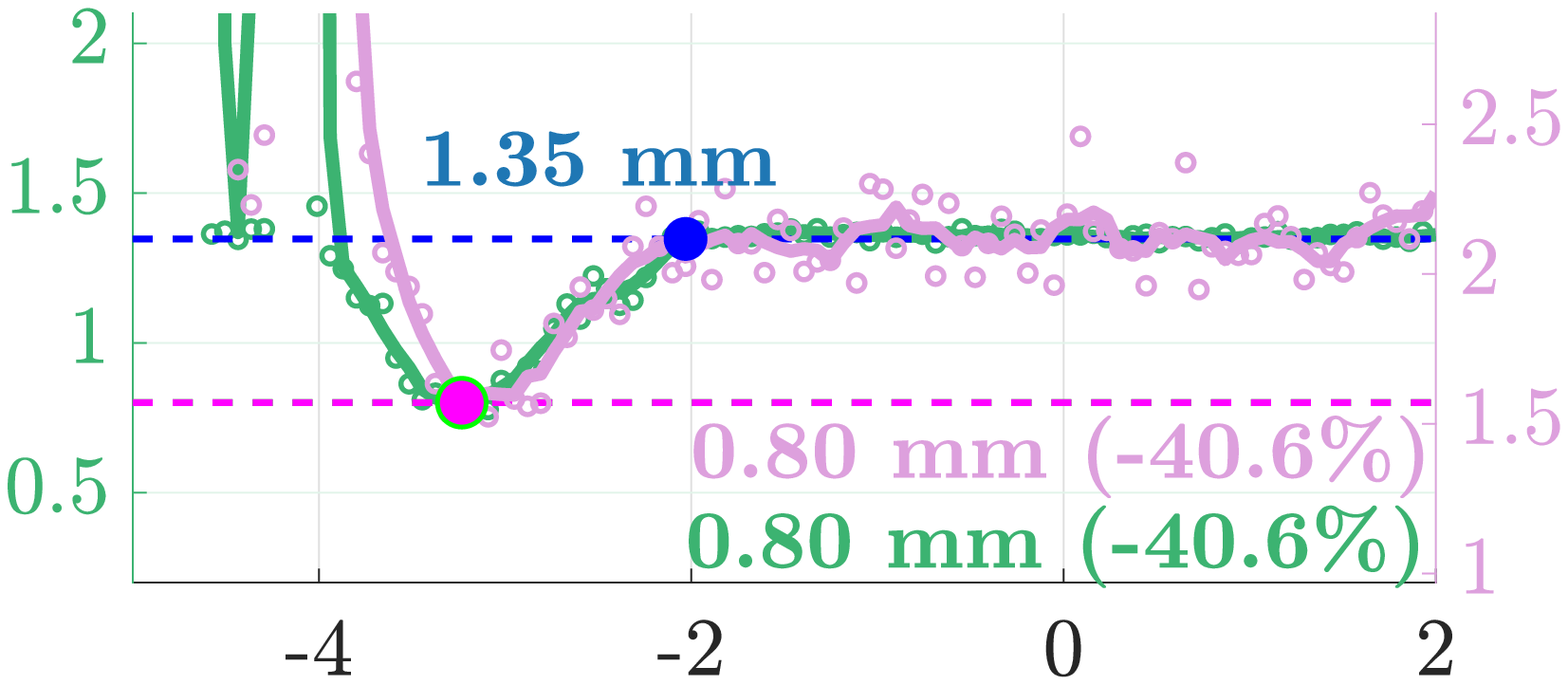}
    \end{subfigure}
    \hfill
    \begin{subfigure}[b]{0.19\textwidth}
         \centering
         \includegraphics[width=1.00\textwidth]{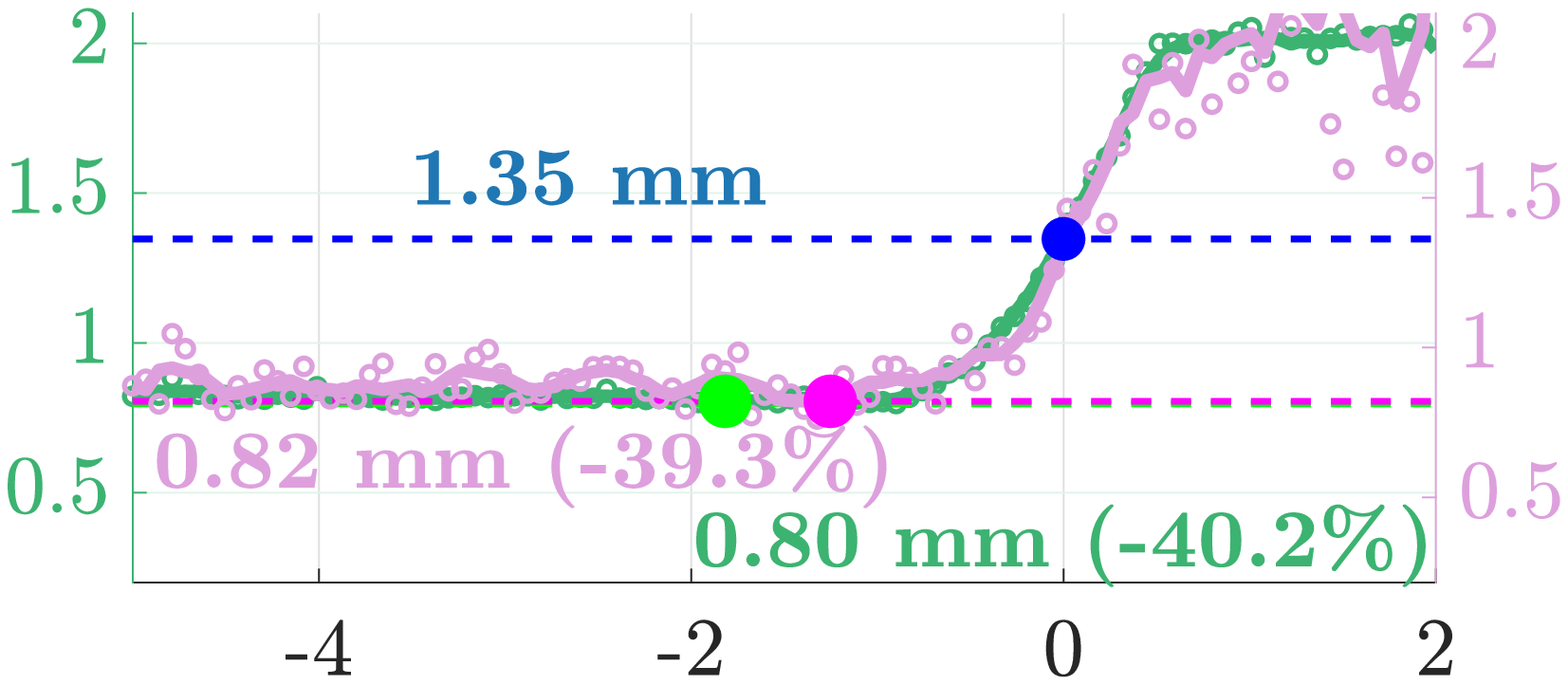}
    \end{subfigure}
    \hfill
    \begin{subfigure}[b]{0.19\textwidth}
         \centering
         \includegraphics[width=1.00\textwidth]{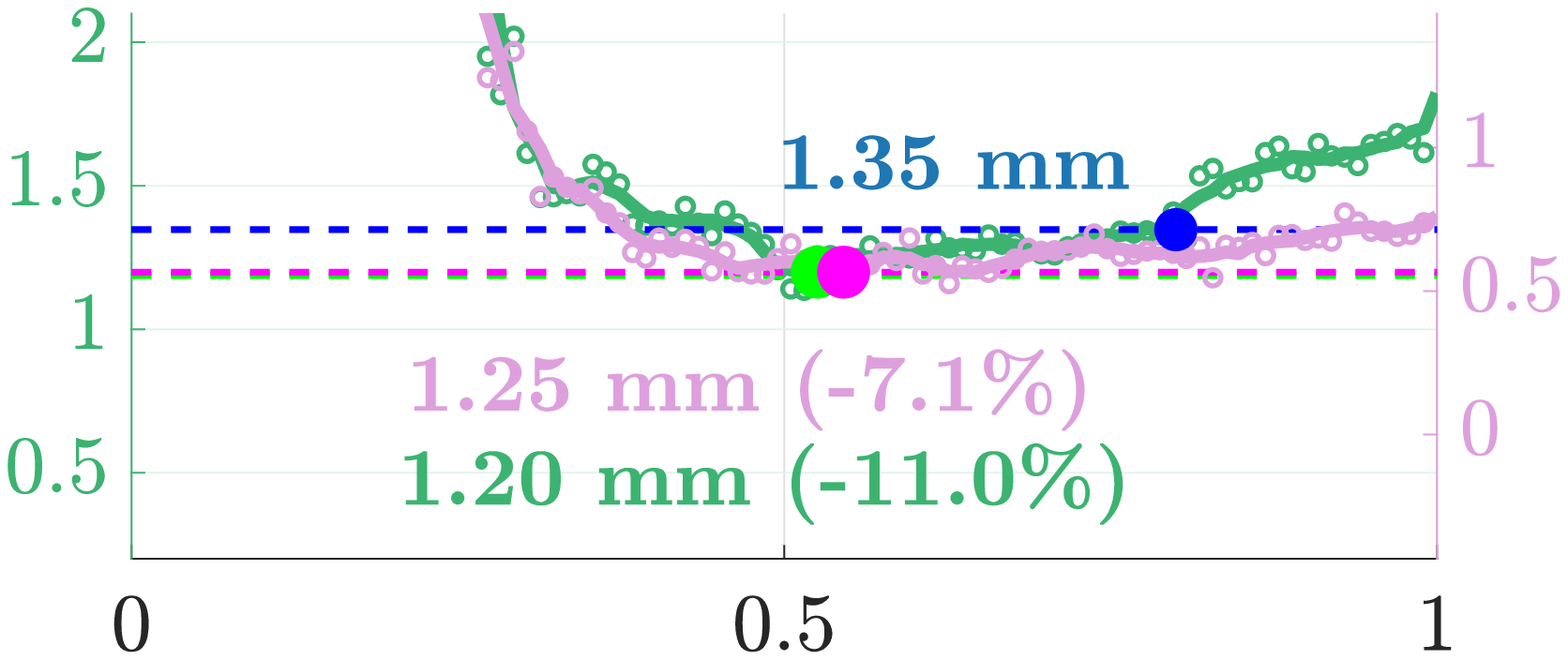}
    \end{subfigure}
    \hfill
    \begin{subfigure}[b]{0.19\textwidth}
         \centering
         \includegraphics[width=1.00\textwidth]{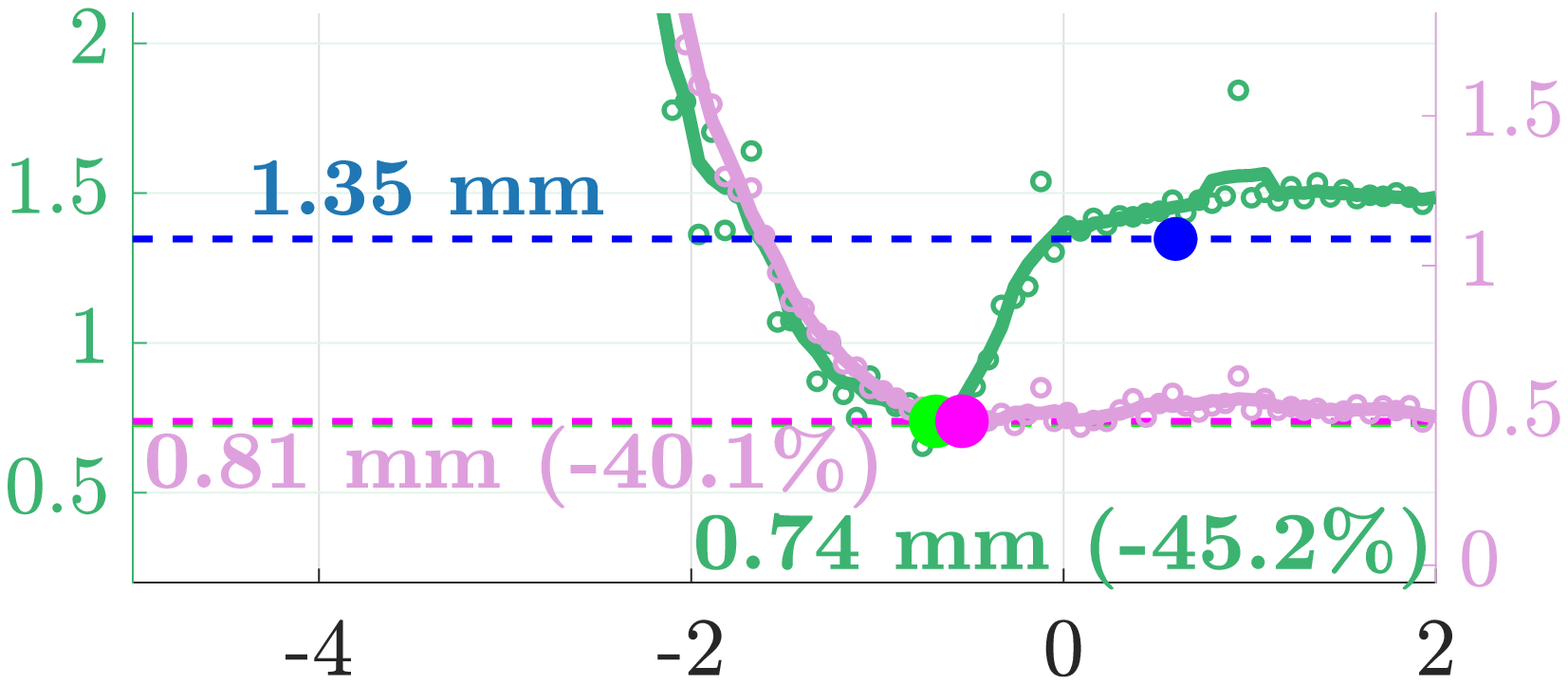}
    \end{subfigure}
    \hfill
    \begin{subfigure}[b]{0.01\textwidth}
        \centering
        \rotatebox{90}{\parbox{1.5cm}{\centering \footnotesize \gtf}}
    \end{subfigure}
\end{subfigure}  
\begin{subfigure}[b]{1.0\textwidth}
    \centering
    \captionsetup{position=above} 
    \caption*{\textbf{Dataset} : Tartanair -- \textbf{Sequence} : ME 001} 
    \vspace{-0.2cm}
    \begin{subfigure}[b]{0.01\textwidth}
        \centering
        \rotatebox{90}{\parbox{1.7cm}{\centering  \footnotesize \ate}}
    \end{subfigure}
    \hfill
    \begin{subfigure}[b]{0.19\textwidth}
         \centering
         \includegraphics[width=1.00\textwidth]{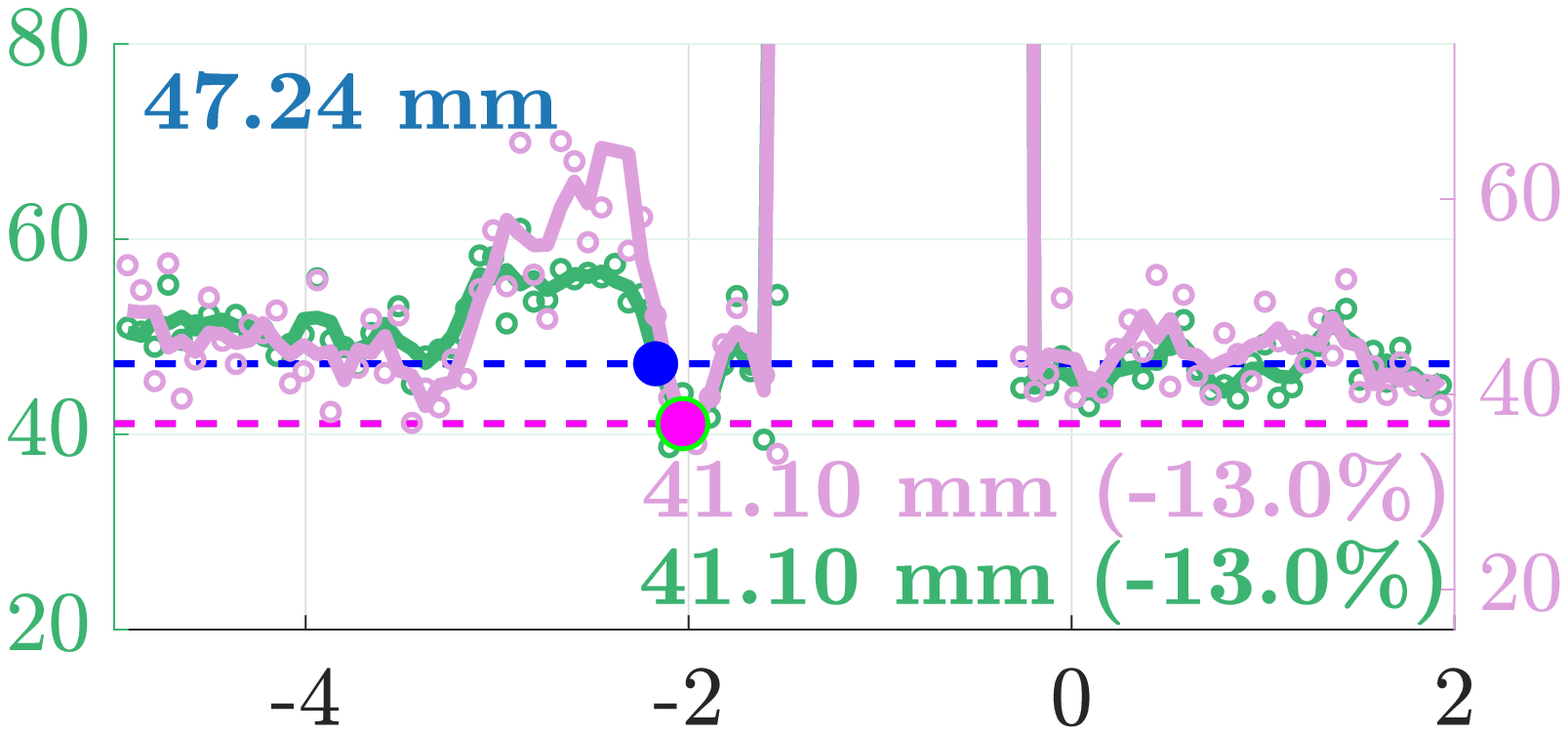}
    \end{subfigure}
    \hfill
    \begin{subfigure}[b]{0.19\textwidth}
         \centering
         \includegraphics[width=1.00\textwidth]{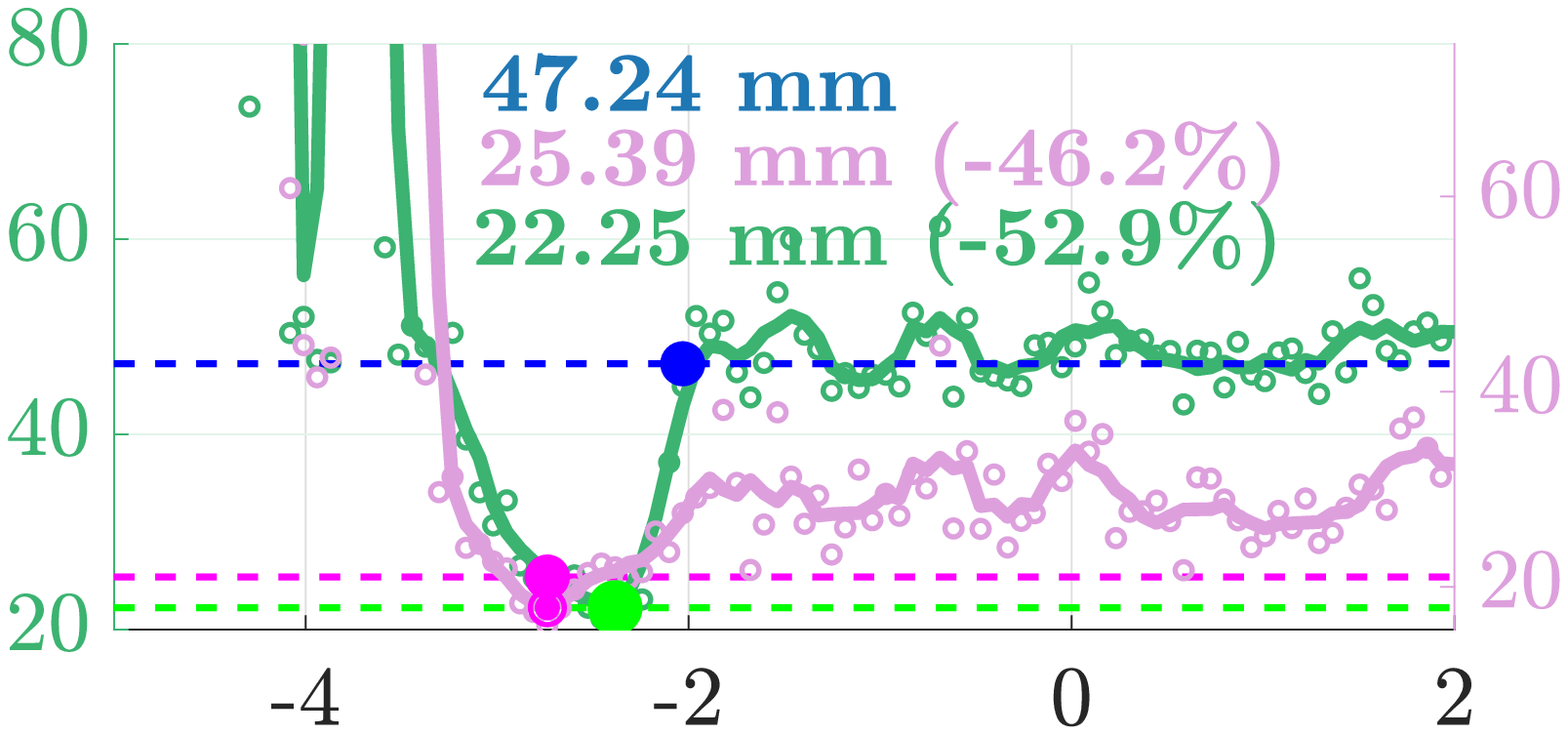}
    \end{subfigure}
    \hfill
    \begin{subfigure}[b]{0.19\textwidth}
         \centering
         \includegraphics[width=1.00\textwidth]{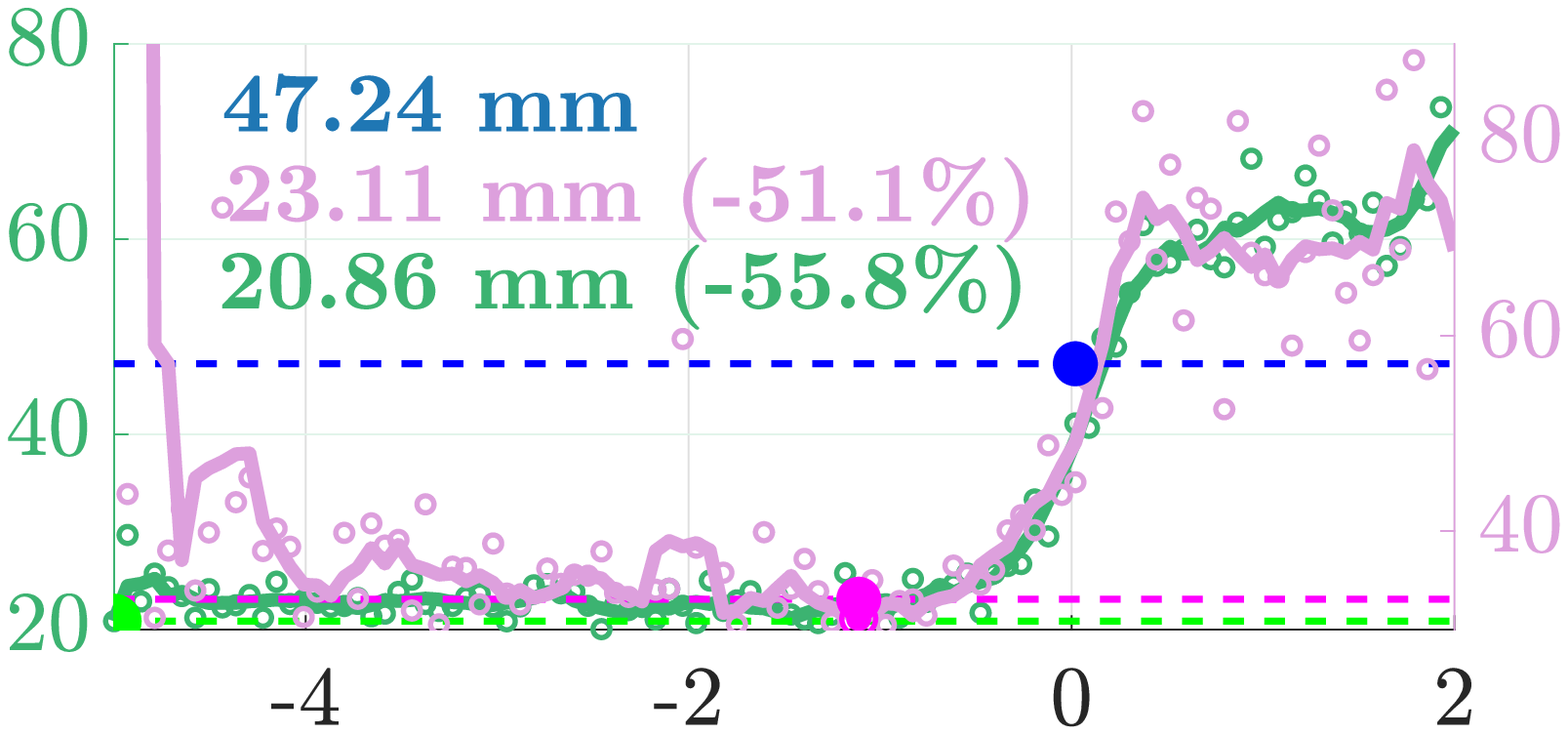}
    \end{subfigure}
    \hfill
    \begin{subfigure}[b]{0.19\textwidth}
         \centering
         \includegraphics[width=1.00\textwidth]{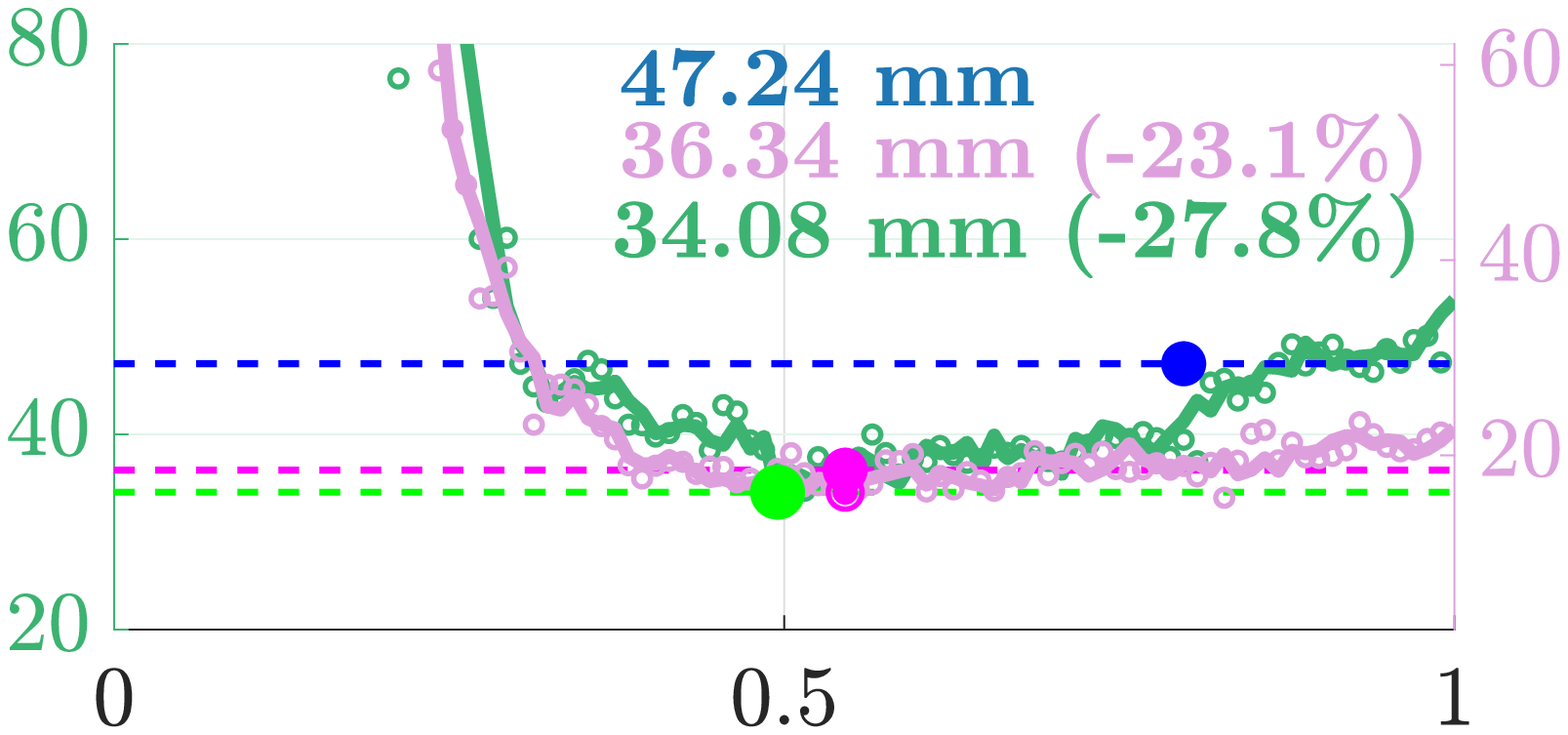}
    \end{subfigure}
    \hfill
    \begin{subfigure}[b]{0.19\textwidth}
         \centering
         \includegraphics[width=1.00\textwidth]{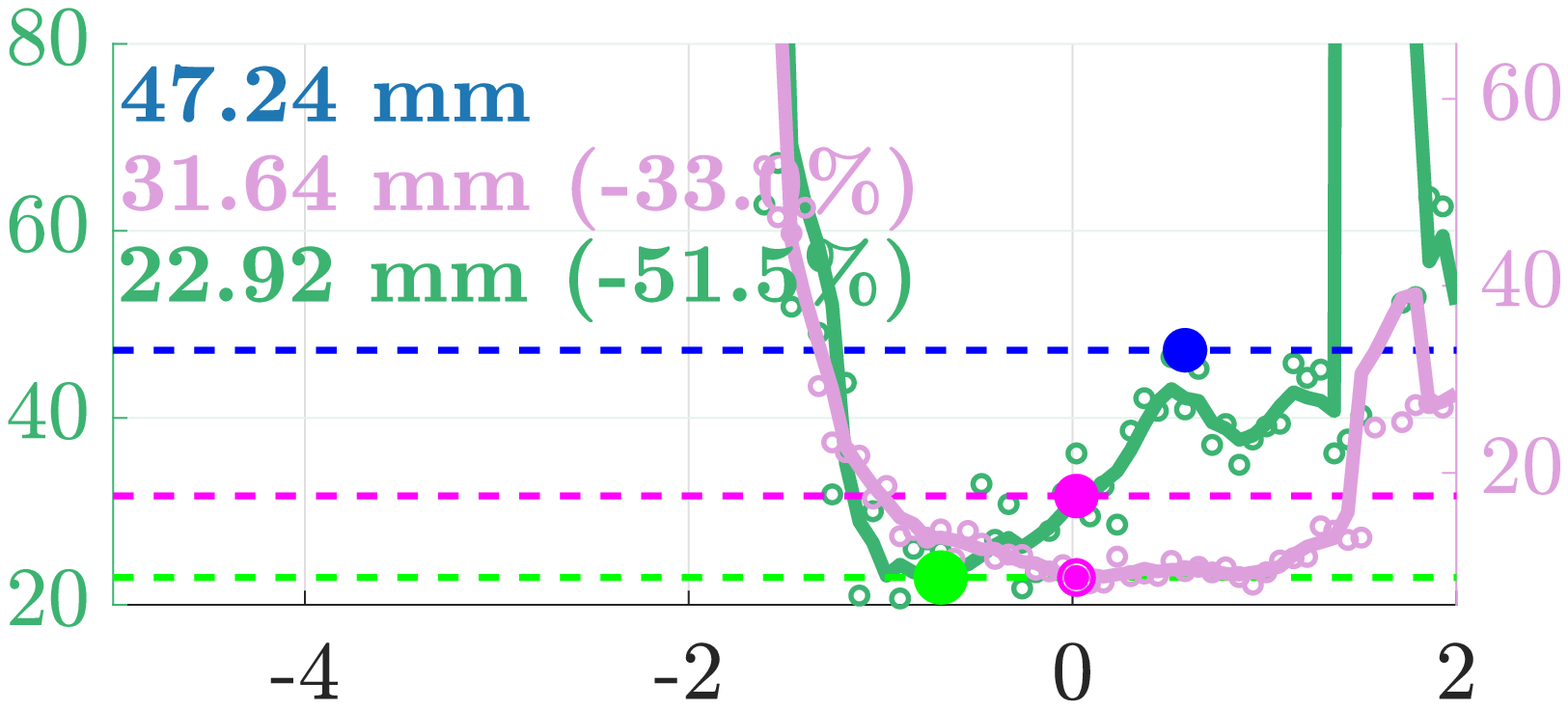}
    \end{subfigure}
    \hfill
    \begin{subfigure}[b]{0.01\textwidth}
        \centering
        \rotatebox{90}{\parbox{1.7cm}{\centering \footnotesize \gtf}}
    \end{subfigure}
\end{subfigure}
\begin{subfigure}[b]{1.0\textwidth}
    \centering
    \captionsetup{position=above} 
    \caption*{\textbf{Dataset} : ETH -- \textbf{Sequence} : Table 3}
    \vspace{-0.9cm}
    \begin{subfigure}[b]{0.01\textwidth}
        \centering
        \rotatebox{90}{\parbox{3cm}{\centering  \footnotesize \ate}}
    \end{subfigure}
    \hfill
    \begin{subfigure}[b]{0.19\textwidth}
         \centering
         \includegraphics[width=1.00\textwidth]{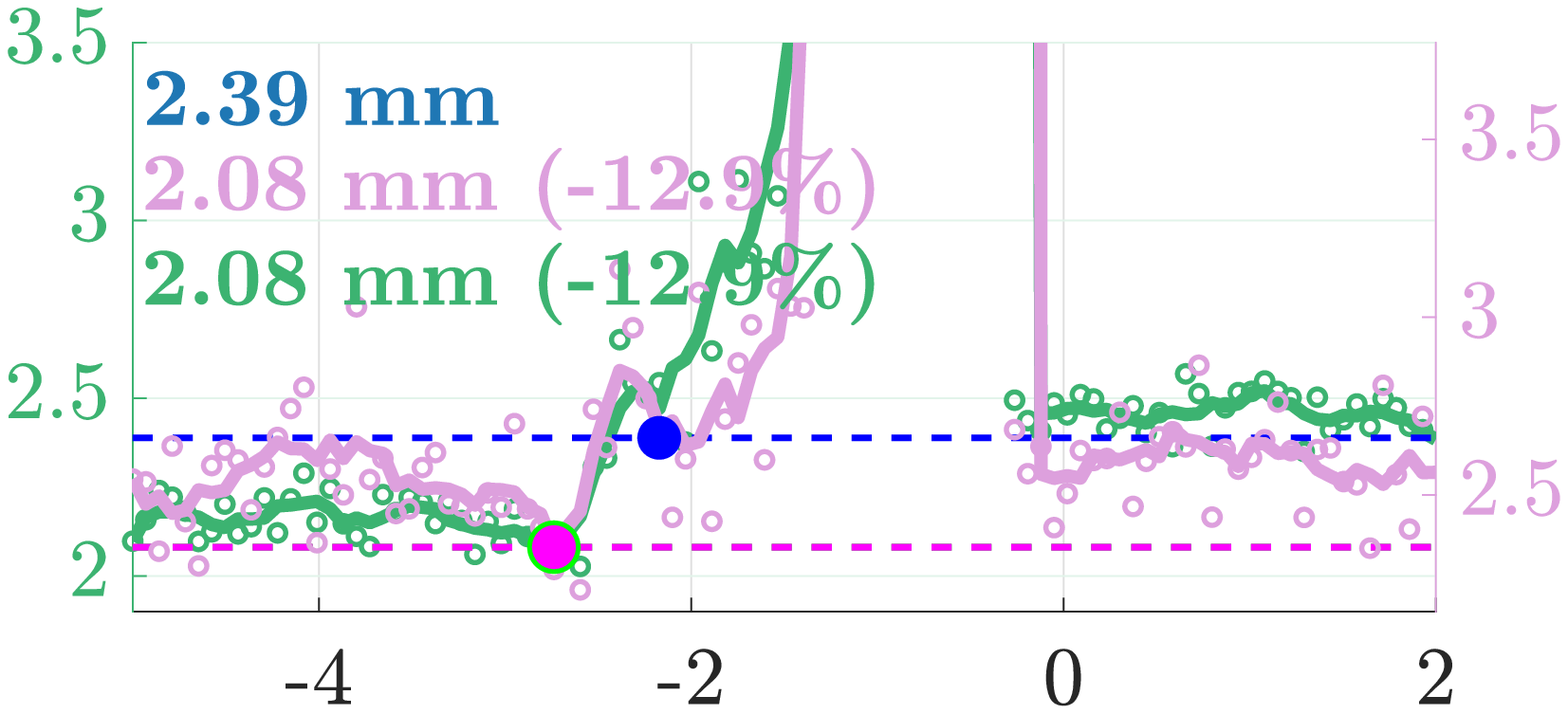}
         \caption*{\centering \textbf{$\log_{10}$ Sift Ext. Peak}}
    \end{subfigure}
    \hfill
    \begin{subfigure}[b]{0.19\textwidth}
         \centering
         \includegraphics[width=1.00\textwidth]{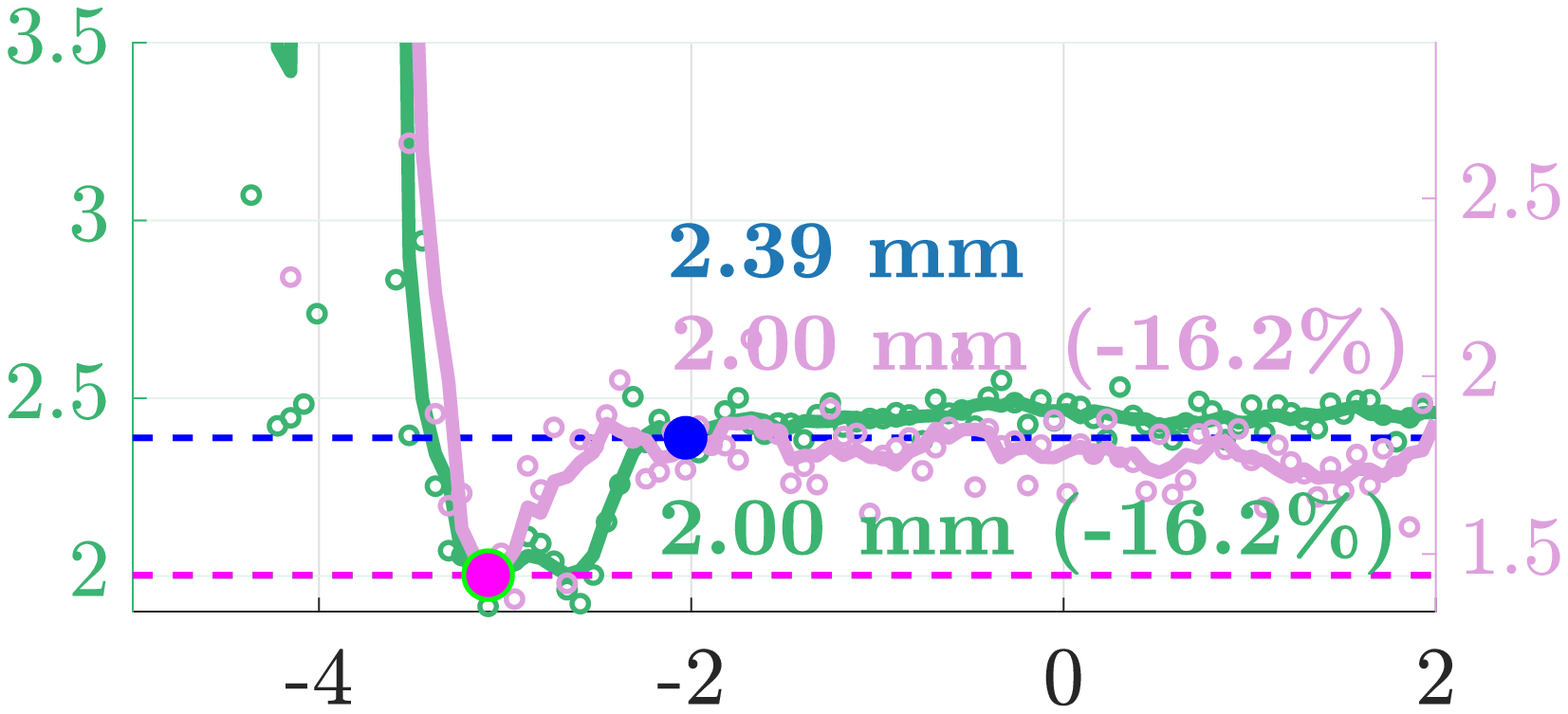}
         \caption*{\centering \textbf{$\log_{10}$ Max. BA $\boldsymbol{e_r}$}}
    \end{subfigure}
    \hfill
    \begin{subfigure}[b]{0.19\textwidth}
         \centering
         \includegraphics[width=1.00\textwidth]{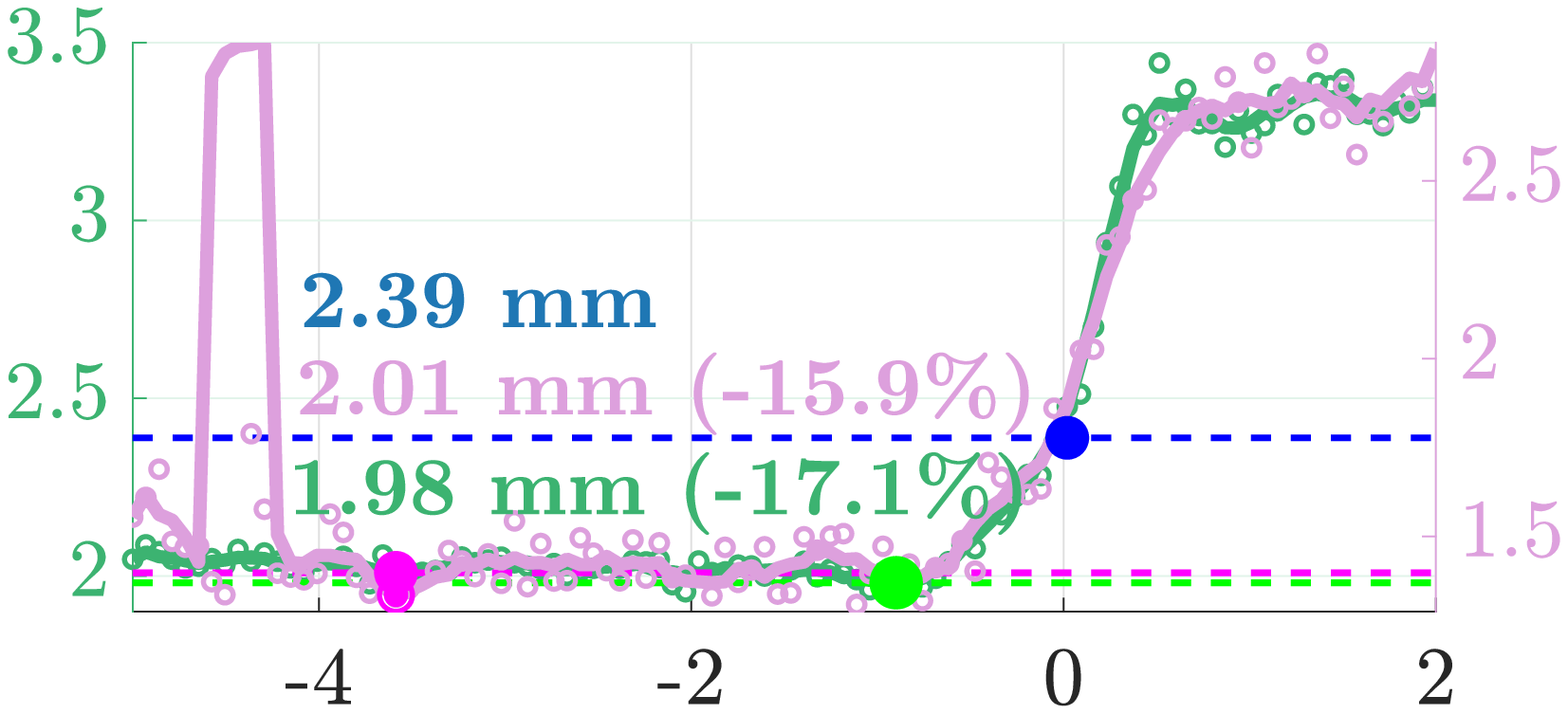}
         \caption*{\centering \textbf{$\log_{10}$ BA Huber Loss}}
    \end{subfigure}
    \hfill
    \begin{subfigure}[b]{0.19\textwidth}
         \centering
         \includegraphics[width=1.00\textwidth]{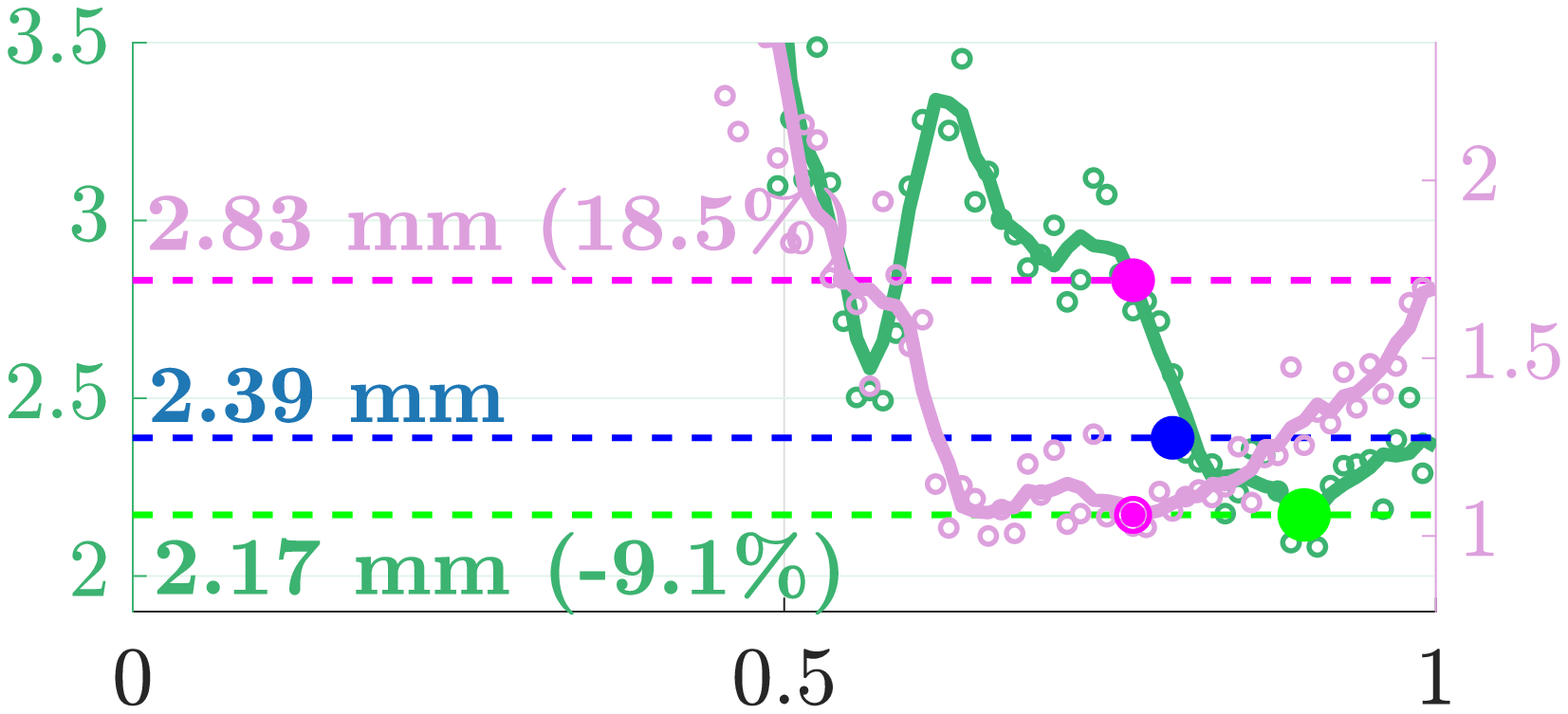}
        \caption*{\centering \textbf{Sift Match. Max. ratio}}
    \end{subfigure}
    \hfill
    \begin{subfigure}[b]{0.19\textwidth}
         \centering
         \includegraphics[width=1.00\textwidth]{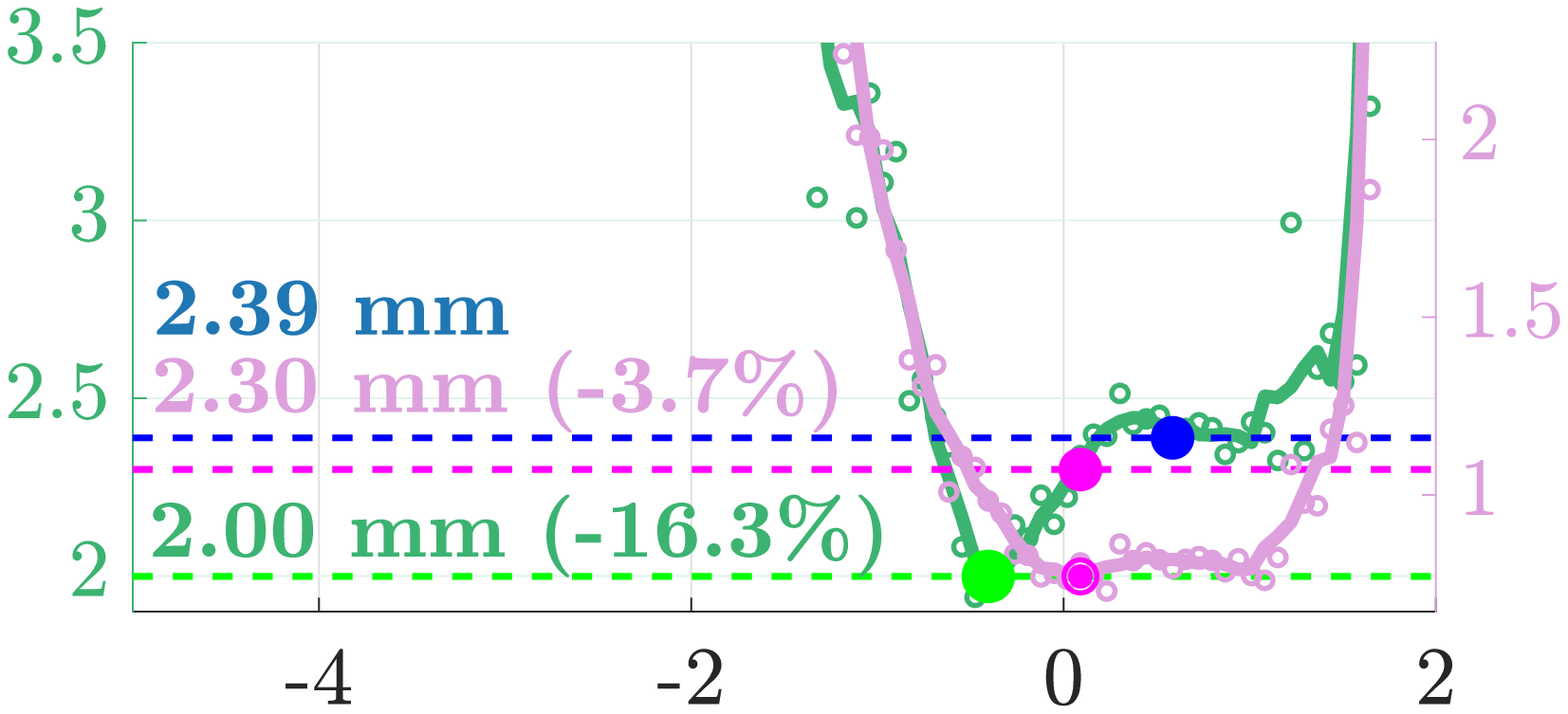}
         \caption*{\centering \textbf{$\log_{10}$ 2V Geo. Max. $e_r$}}
    \end{subfigure}
    \hfill
    \begin{subfigure}[b]{0.01\textwidth}
        \centering
        \rotatebox{90}{\parbox{2.7cm}{\centering \footnotesize \gtf}}
    \end{subfigure}
\end{subfigure}
\caption{\textcolor{groundtruth}{Green Left-Y-axis shows the ATE computed using ground truth.} \textcolor{proxy}{Pink Right-Y-axis shows our \gtf.} \textcolor{nominal}{$\bullet$Blue dots indicate the ATE of GLOMAP operating with nominal parameters.} \textcolor{groundtruth}{$\bullet$Minimum ATE achieved when fine-tuning with ground truth.} \textcolor{proxy}{$\bullet$Minimum ATE achieved using our \gtf, without requiring ground truth data.}}
\label{fig:fine_tuning_glomap}
\end{figure*}

\begin{table*}[ht]
\begin{minipage}{0.61\textwidth}
\centering
\resizebox{\columnwidth}{!}{ 
\begin{tabular}{c c c c c c c}
    \toprule
     & \rotatebox{60}{\makecell{\textcolor{nominal}{Nominal} \\ \textcolor{nominal}{ATE}}} 
     & \rotatebox{60}{\makecell{Sift Ext. \\ Peak}} 
     & \rotatebox{60}{\makecell{Max. BA \\ $e_r$}} 
     & \rotatebox{60}{\makecell{BA Huber \\ Loss}} 
     & \rotatebox{60}{\makecell{Sift Match. \\ Max. ratio}} 
     & \rotatebox{60}{\makecell{2V Geo. \\ Max. $e_r$}}\\
    \midrule
    \makecell{Replica~\cite{replica19arxiv} \\ Office 0} & \textcolor{nominal}{1.35 mm}&
    \makecell{\textcolor{proxy}{\underline{0.45 / 66.2\%}} \\ \textcolor{groundtruth}{0.38 / 71.5\%}} &
    \makecell{\textcolor{proxy}{\underline{0.80 / 40.6\%}} \\ \textcolor{groundtruth}{0.80 / 40.6\%}} &
    \makecell{\textcolor{proxy}{\underline{0.82 / 39.3\%}} \\ \textcolor{groundtruth}{0.80 / 40.2\%}} &
    \makecell{\textcolor{proxy}{\underline{1.25 / 7.1\%}} \\ \textcolor{groundtruth}{1.20 / 11.0\%}} &
    \makecell{\textcolor{proxy}{\underline{0.81 / 40.1\%}} \\ \textcolor{groundtruth}{0.74 / 45.2\%}} \\
    \midrule
    \makecell{TartanAir~\cite{wang2020tartanair} \\ ME 001} & \textcolor{nominal}{4.72 cm} &
    \makecell{\textcolor{proxy}{\underline{4.10 / 13.0\%}} \\ \textcolor{groundtruth}{4.10 / 13.0\%}} & 
    \makecell{\textcolor{proxy}{\underline{2.53 / 46.2\%}} \\ \textcolor{groundtruth}{2.22 / 52.9\%}} & 
    \makecell{\textcolor{proxy}{\underline{2.31 / 51.1\%}} \\ \textcolor{groundtruth}{2.08 / 55.8\%}} & 
    \makecell{\textcolor{proxy}{\underline{3.63 / 23.1\%}} \\ \textcolor{groundtruth}{3.41 / 27.8\%}} & 
    \makecell{\textcolor{proxy}{\underline{3.16 / 33.0\%}} \\ \textcolor{groundtruth}{2.29 / 51.5\%}} \\
    \midrule
    \makecell{ETH3D~\cite{schops2019bad} \\ Table 3} & \textcolor{nominal}{2.39 mm} &
    \makecell{\textcolor{proxy}{\underline{2.08 / 12.9\%}} \\ \textcolor{groundtruth}{2.08 / 12.9\%}} & 
    \makecell{\textcolor{proxy}{\underline{2.00 / 16.2\%}} \\ \textcolor{groundtruth}{2.00 / 16.2\%}} &
    \makecell{\textcolor{proxy}{\underline{2.01 / 15.9\%}} \\ \textcolor{groundtruth}{1.98 / 17.1\%}} &
    \makecell{\textcolor{proxy}{2.83 / -18.5\%} \\ \textcolor{groundtruth}{2.17 / 9.1\%}} &
    \makecell{\textcolor{proxy}{\underline{2.30 / 3.7\%}} \\ \textcolor{groundtruth}{2.00 / 16.3\%}} \\
    \bottomrule
\end{tabular}
}
\captionof{table}{GLOMAP~\cite{pan2024global}}
\label{tab:glomap_ablation}
\end{minipage} 
\hfill
\begin{minipage}{0.39\textwidth}
\resizebox{1\columnwidth}{!}{ 
\begin{tabular}{c c c c c}
    \toprule
     & \rotatebox{60}{\makecell{\textcolor{nominal}{Nominal} \\ \textcolor{nominal}{ATE}}} 
     & \rotatebox{60}{\makecell{KeyFrame \\ Threshold}} 
     & \rotatebox{60}{\makecell{Beta \\}}
     & \rotatebox{60}{\makecell{Frontend \\ Threshold }}\\
    \midrule
    \makecell{Replica~\cite{replica19arxiv} \\ Office 2} & \textcolor{nominal}{2.17 mm} &
    \makecell{\textcolor{proxy}{\underline{2.08 / 4.1\%}} \\ \textcolor{groundtruth}{2.01 / 7.4\%}} & 
    \makecell{\textcolor{proxy}{\underline{1.83 / 15.3\%}} \\ \textcolor{groundtruth}{1.83 / 15.3\%}} & 
    \makecell{\textcolor{proxy}{\underline{2.16 / 0.1\%}} \\ \textcolor{groundtruth}{2.12 / 2.3\%}} \\
    \midrule
    \makecell{NUIM~\cite{handa2010scalable} \\  lvr 0}  & \textcolor{nominal}{2.54 mm} &   
    \makecell{\textcolor{proxy}{\underline{2.31 / 8.9\%}} \\ \textcolor{groundtruth}{2.18 / 14.3\%}} & 
    \makecell{\textcolor{proxy}{\underline{2.26 / 11.2\%}} \\ \textcolor{groundtruth}{2.01 / 20.9\%}} & 
    \makecell{\textcolor{proxy}{\underline{2.37/ 6.9\%}} \\ \textcolor{groundtruth}{1.99 / 21.7\%}} \\
    \midrule
    \makecell{ETH3D~\cite{schops2019bad} \\  Cables 1}  & \textcolor{nominal}{4.86 mm} &  
    \makecell{\textcolor{proxy}{\underline{4.66 / 4.1\%}} \\ \textcolor{groundtruth}{4.53 / 7.0\%}} & 
    \makecell{\textcolor{proxy}{\underline{4.80 / 1.3\%}} \\ \textcolor{groundtruth}{4.72 / 2.9\%}} & 
    \makecell{\textcolor{proxy}{\underline{4.76 / 2.2\%}} \\ \textcolor{groundtruth}{4.58 / 5.9\%}} \\
    \bottomrule
\end{tabular}
}
\captionof{table}{DROID-SLAM~\cite{teed2021droid}}
\label{tab:droidslam_ablation}
\end{minipage}
\caption*{\textbf{Hyperparameter Fine-Tuning.} ATE for the system operating with {\textcolor{nominal}{nominal}} parameters, fine-tuned using our {\textcolor{proxy}{ground-truth-free}} metric, and fine-tuned using a {\textcolor{groundtruth}{ground-truth-based}} metric. Note how our approach consistently \textbf{improves precision in \numSuccesfulExperimentsGlomap\ out of \numExperimentsGlomap} experiments compared to GLOMAP with nominal parameters, achieving an \textbf{average improvement of \avgImprGtfGlomap}. Moreover, it delivers performance comparable to ground-truth-based tuning, which achieves an \textbf{average improvement of \avgImprGtGlomap}. Similarly, our approach \textbf{improves in \numSuccesfulExperimentsDroid\ out of \numExperimentsDroid} experiments compared to DROID-SLAM, achieving an \textbf{average improvement of \avgImprGtfDroid}, comparable to the ground-truth-based \textbf{average improvement of \avgImprGtDroid}.} 
\label{tab:fine_tuning_ablation}
\end{table*}
\begin{figure}[ht]
\begin{subfigure}[b]{0.5\textwidth}
    \centering
    \captionsetup{position=above} 
    \caption*{\textbf{Dataset}: Replica -- \textbf{Sequence}: Office 1}
    \vspace{-0.25cm}
    \begin{subfigure}[b]{0.01\textwidth}
        \centering
        \rotatebox{90}{\parbox{2.0cm}{\centering  \footnotesize \ate}}     
    \end{subfigure}
    \hfill
    \begin{subfigure}[b]{0.47\textwidth}
         \centering
         \includegraphics[width=1.0\textwidth]{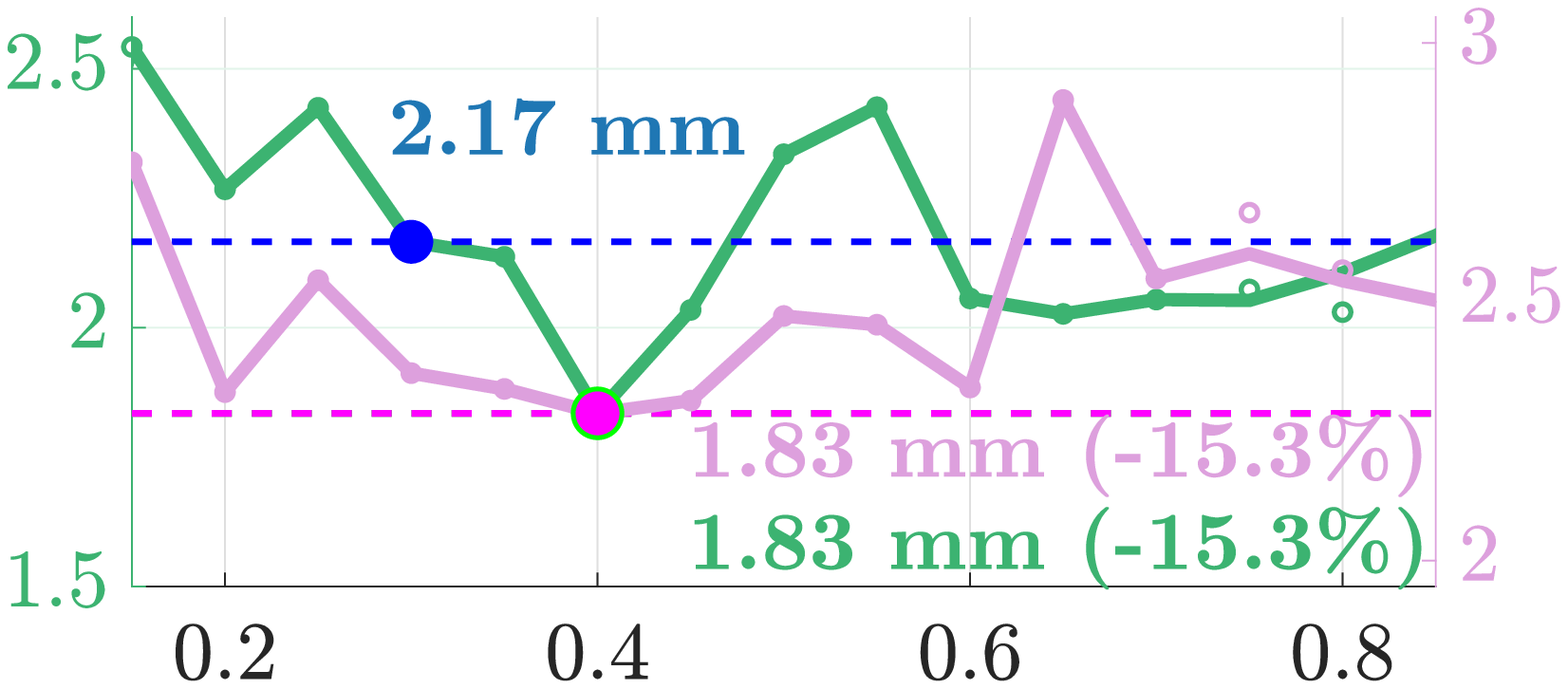}
    \end{subfigure}
    \hfill
    \begin{subfigure}[b]{0.47\textwidth}
         \centering
         \includegraphics[width=1.0\textwidth]{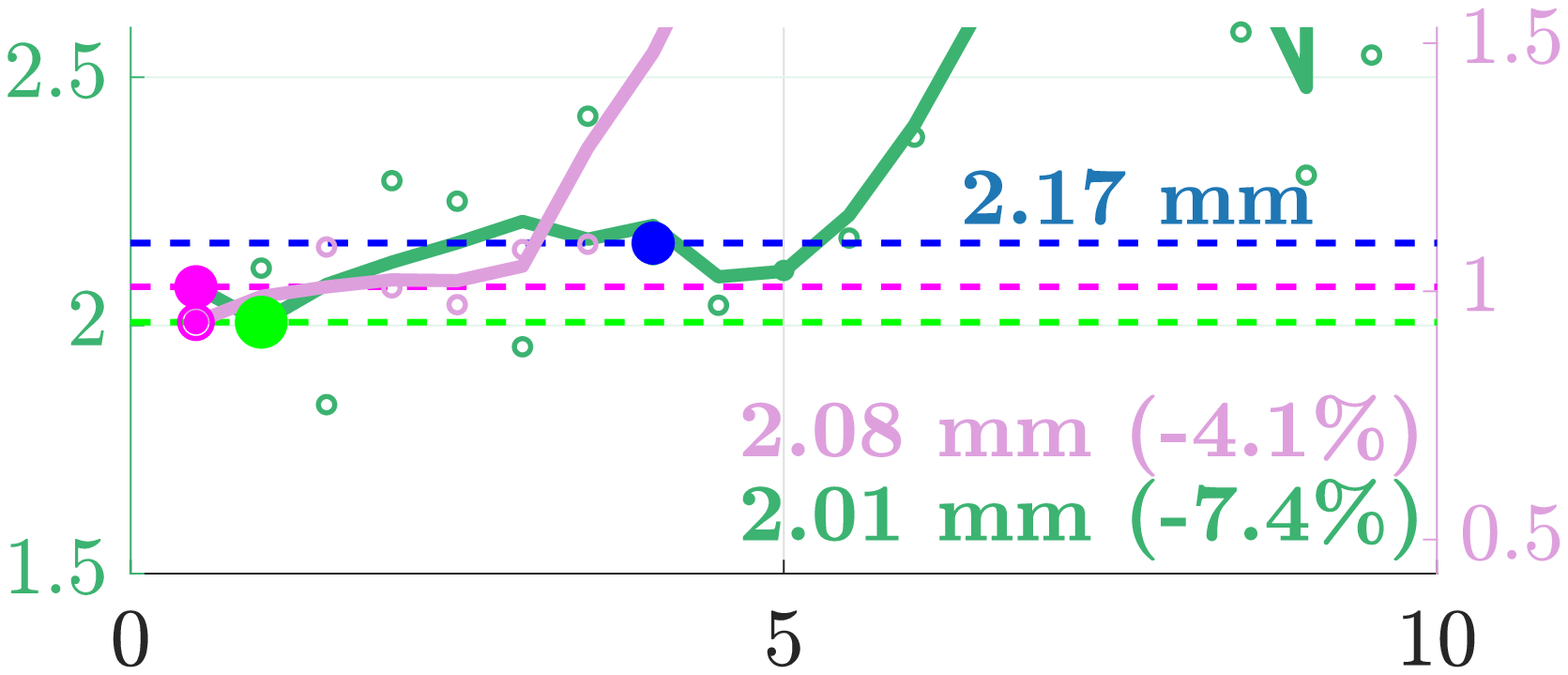}
    \end{subfigure}
    \hfill
    \begin{subfigure}[b]{0.01\textwidth}
        \centering
        \rotatebox{90}{\parbox{2cm}{\centering \footnotesize \gtf}}
    \end{subfigure}
\end{subfigure}
\vfill
\begin{subfigure}[b]{0.5\textwidth}
    \centering
    \captionsetup{position=above} 
    \caption*{\textbf{Dataset}: ETH -- \textbf{Sequence}: Cables 1}
    \vspace{-0.75cm}
    \begin{subfigure}[b]{0.01\textwidth}
        \centering
        \rotatebox{90}{\parbox{3.5cm}{\centering  \footnotesize \ate}}     
    \end{subfigure}
    \hfill
    \begin{subfigure}[b]{0.47\textwidth}
         \centering
         \includegraphics[width=1.0\textwidth]{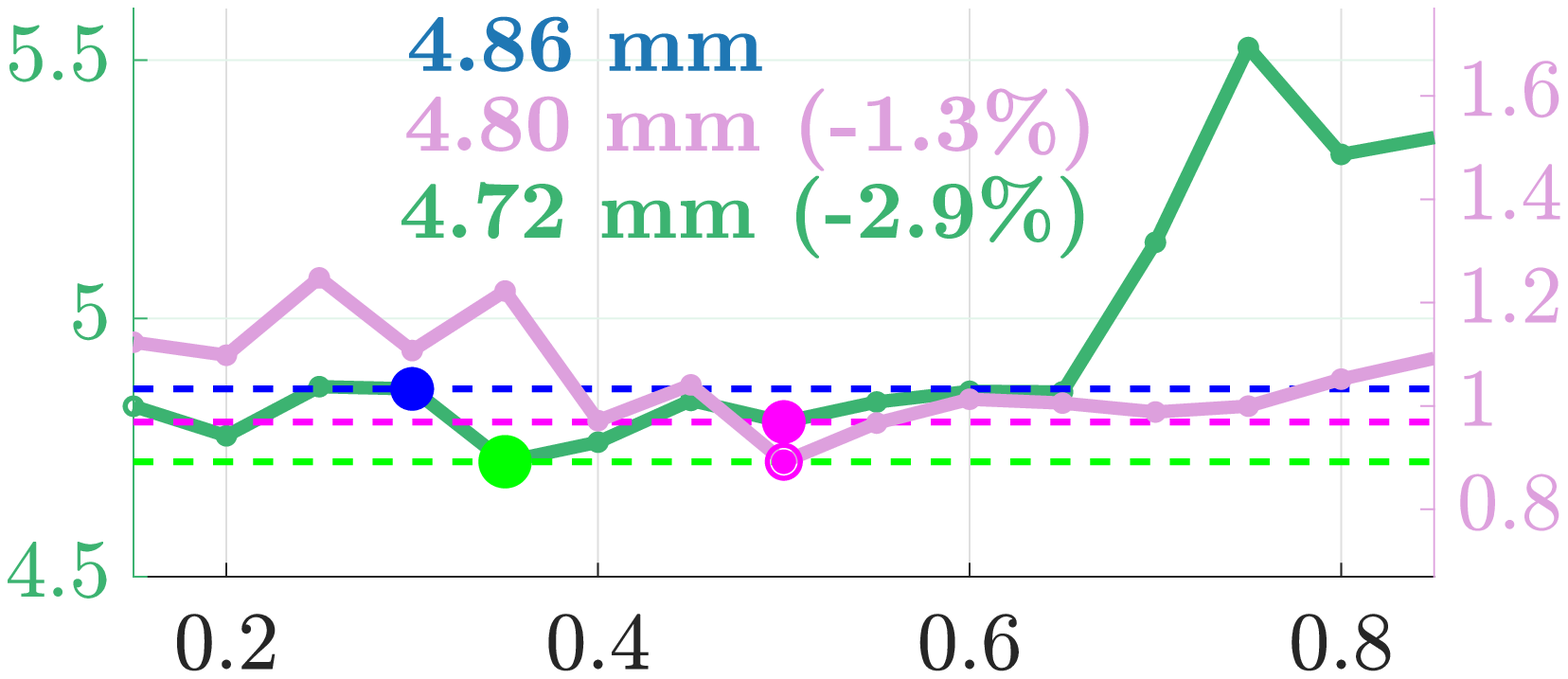}
         \caption*{\textbf{Beta}}
    \end{subfigure}
    \hfill
    \begin{subfigure}[b]{0.47\textwidth}
         \centering
         \includegraphics[width=1.0\textwidth]{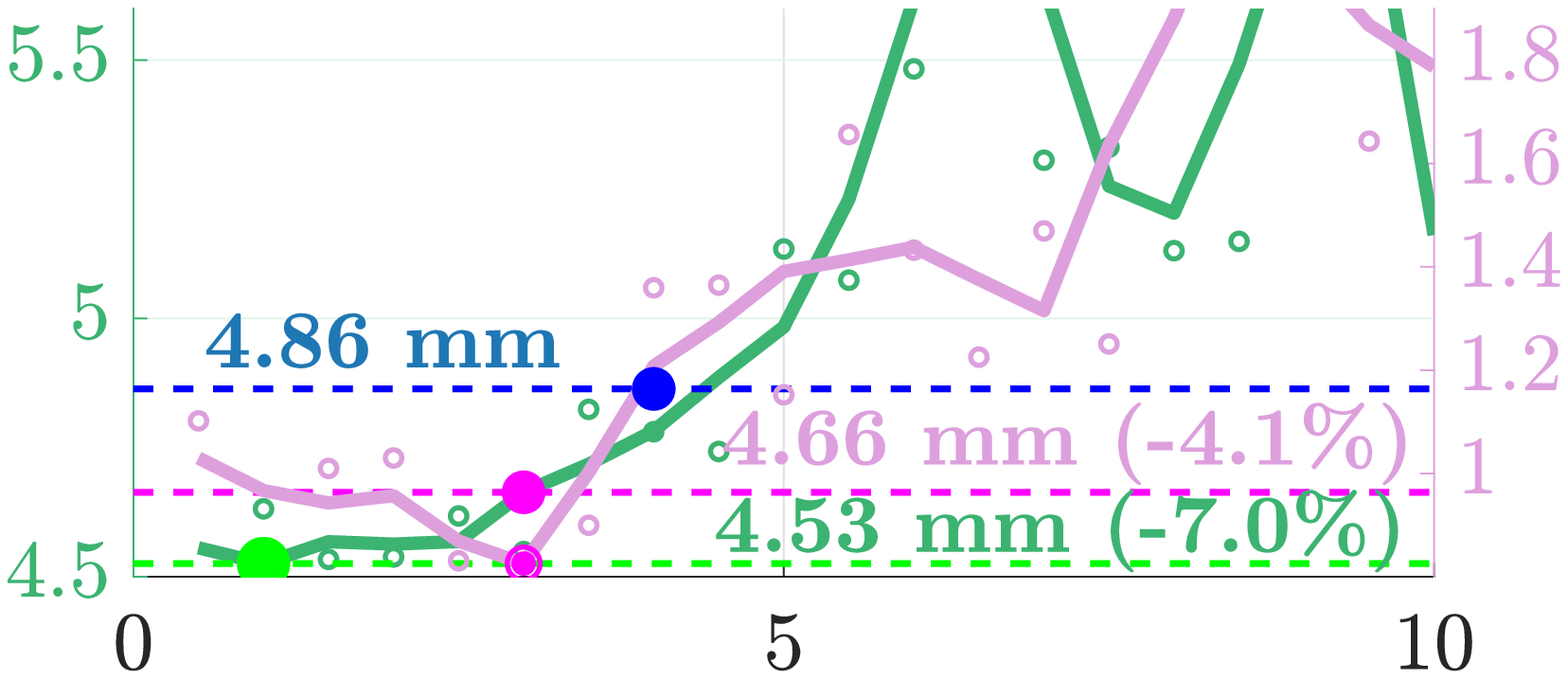}
         \caption*{\textbf{Keyframe. Th.}}
    \end{subfigure}
    \hfill
    \begin{subfigure}[b]{0.01\textwidth}
        \centering
        \rotatebox{90}{\parbox{3.5cm}{\centering \footnotesize \gtf}}
    \end{subfigure}
\end{subfigure}
\caption{\textcolor{groundtruth}{Green Left-Y-axis shows the ATE computed using ground truth.} \textcolor{proxy}{Pink Right-Y-axis shows our \gtf.} \textcolor{nominal}{$\bullet$Blue dots indicate the ATE of GLOMAP operating with nominal parameters.} \textcolor{groundtruth}{$\bullet$Minimum ATE achieved when fine-tuning with ground truth.} \textcolor{proxy}{$\bullet$Minimum ATE achieved using our \gtf, without requiring ground truth data.}}
\label{fig:fine_tuning_droid}
\end{figure}

The experiments in this section show that the \gtf\ described and motivated in Section~\ref{sec:formulation}, effectively and accurately correlates with standard \ate. Consequently, it can be used to tune SfM and VSLAM pipelines \emph{without ground truth}. Specifically, we evaluate the strength of this correlation by applying our approach to the downstream task of hyperparameter tuning.

\subsection{Experimental Setup}

\textbf{Datasets.} To rigorously evaluate the generality of our approach across a wide variety of conditions, and consistent with the evaluation methodologies of recent SfM/VSLAM studies~\cite{wang2021tartanvo,teed2021droid,ye2023pvo,pan2024global,matsuki2024gaussian,lipson2025deep}, we conduct a quantitative analysis on sequences from a representative selection of \numDatasets public datasets utilized by state-of-the-art baselines (see Table \ref{tab:state-of-the-art}). These include both synthetic datasets—Replica~\cite{replica19arxiv}, NUIM~\cite{handa2014benchmark} and TartanAir~\cite{wang2020tartanair}—as well as a real-world dataset—ETH3D~\cite{schops2019bad}.

\noindent \textbf{Baselines.} To solidly assess the generality of our approach, we selected two distinct pipelines: the recent feature-based SfM pipeline GLOMAP~\cite{pan2024global} and the deep learning-based VSLAM pipeline DROID-SLAM~\cite{teed2021droid}. Both represent the state-of-the-art in their respective fields, delivering notable accuracy improvements over prior work and exhibiting robust performance. Moreover, GLOMAP is substantially faster than other SfM pipelines and DROID-SLAM runs in real time, as expected from a SLAM code.

\noindent \textbf{Metrics.} We use the Absolute Trajectory Error (ATE), with a $Sim(3)$ alignment to account for scale differences between trajectories~\cite{sturm12iros,zhang2018tutorial}. As outlined in our formulation and methodology (Section~\ref{sec:formulation}), our approach is flexible and can be extended to other VSLAM modalities (\eg, \mbox{RGB-D}, stereo, or visual-inertial), employing metrics tailored to each specific setup, such as the Relative Pose Error (RPE) described in Section~\ref{sec:related}.

\noindent \textbf{Hardware Details.} We conducted the DROID-SLAM experiments on a desktop equipped with an Intel Core i7-12700K (3.60 GHz) processor and a single NVIDIA GeForce RTX 3090 GPU. For the GLOMAP experiments, we used desktops with varying CPU/GPU configurations, ensuring consistency within each dataset across all experiments.

\subsection{Hyperparameter Tuning in SfM}
\label{subsec:finetuning_glomap}

Hyperparameter tuning seeks to identify the set of hyperparameters that maximizes a model’s performance on a validation set~\cite{andonie2019hyperparameter}. In this paper, we adopt a straightforward \textbf{1-D brute-force parameter search}. This approach keeps the problem computationally constrained, identifies the optimal performance for each experiment, and demonstrates the correlation between our \gtf\ and the standard \ate.

Figure~\ref{fig:fine_tuning_glomap} illustrates the impact on trajectory accuracy (in the vertical axes) of the variation of five of the most influential hyperparameters of GLOMAP (in the horizontal axes). Specifically, in each graph, the \textcolor{groundtruth}{green left Y-axis} represents the standard \ate\ computed using ground truth, while the \textcolor{proxy}{pink right Y-axis} overlays our \gtf. \textcolor{nominal}{$\bullet$ Blue dots} indicate the ATE of GLOMAP with nominal parameters, \textcolor{groundtruth}{$\bullet$ green dots} represent the minimum ATE achieved by hyperparameter tuning using the ground truth, and \textcolor{proxy}{$\bullet$ pink dots} show the minimum ATE obtained using our \gtf, without requiring ground truth data.

First, note the strong correlation between the \ate\ computed with ground truth and our \gtf, as evidenced by the close alignment between the two curves. This highlights our approach’s ability to capture relative variations in trajectory accuracy across different sequences and parameters without relying on ground truth data. Second, observe how fine-tuning with our \gtf\ consistently improves accuracy compared to using the \textcolor{nominal}{nominal} parameters of GLOMAP, once again without requiring ground truth. Finally, our \gtf\ is capable of achieving optimal accuracy comparable to that obtained using ground truth in a substantial percentage of cases, demonstrating its effectiveness in approximating ground truth performance.

Table~\ref{tab:glomap_ablation} summarizes the ATE variations for nominal parameters, fine-tuning with ground truth, and fine-tuning without ground truth using our \gtf. Notably, we improve accuracy in \textbf{\numSuccesfulExperimentsGlomap/\numExperimentsGlomap} experiments, achieving an average improvement of \textbf{\avgImprGtfGlomap} compared to the nominal parameters of GLOMAP, approximating the optimal average improvement of \textbf{\avgImprGtGlomap} obtained when using ground truth.

\subsection{Hyperparameter Tuning in VSLAM}
\label{subsec:finetuning_droid}

Similar to the previous section, we perform 1-D brute-force parameter search for DROID-SLAM. Figure~\ref{fig:fine_tuning_droid} and Table~\ref{tab:droidslam_ablation} summarize the ATE variations for nominal parameters, fine-tuning with ground truth, and fine-tuning without ground truth using our \gtf. Notably, we improve accuracy in \textbf{\numSuccesfulExperimentsDroid/\numExperimentsDroid} experiments, achieving an average improvement of \textbf{\avgImprGtfDroid} compared to the nominal parameters of DROID-SLAM, which is close to the optimal average improvement of \textbf{\avgImprGtDroid} obtained with ground truth.



\section{Ablation Studies}
\label{sec:ablation}
Section \ref{sec:formulation} lays the foundation for our ground-truth-free precision metric. In this section, we perform a series of ablation studies to investigate some of the key aspects. First, we examine how the magnitude of input noise, $\Delta\sigma$, impacts performance (Section~\ref{subsec:noise}). Next, we compare the correlation between our \gtf\ and a reprojection error metric against actual ground truth data (Section~\ref{subsec:reproj_error}). Finally, we analyze the computational cost of our approach (Section~\ref{subsec:computational}).

\subsection{Input Noise Magnitude}
\label{subsec:noise}

\begin{figure}[t]
    \centering
    \begin{subfigure}[b]{0.03\textwidth}
        \centering
        \rotatebox{90}{\parbox{4cm}{\centering \large \ate}}
    \end{subfigure}
    \begin{subfigure}[b]{0.32\textwidth}
         \centering
         \includegraphics[width=1.0\textwidth]{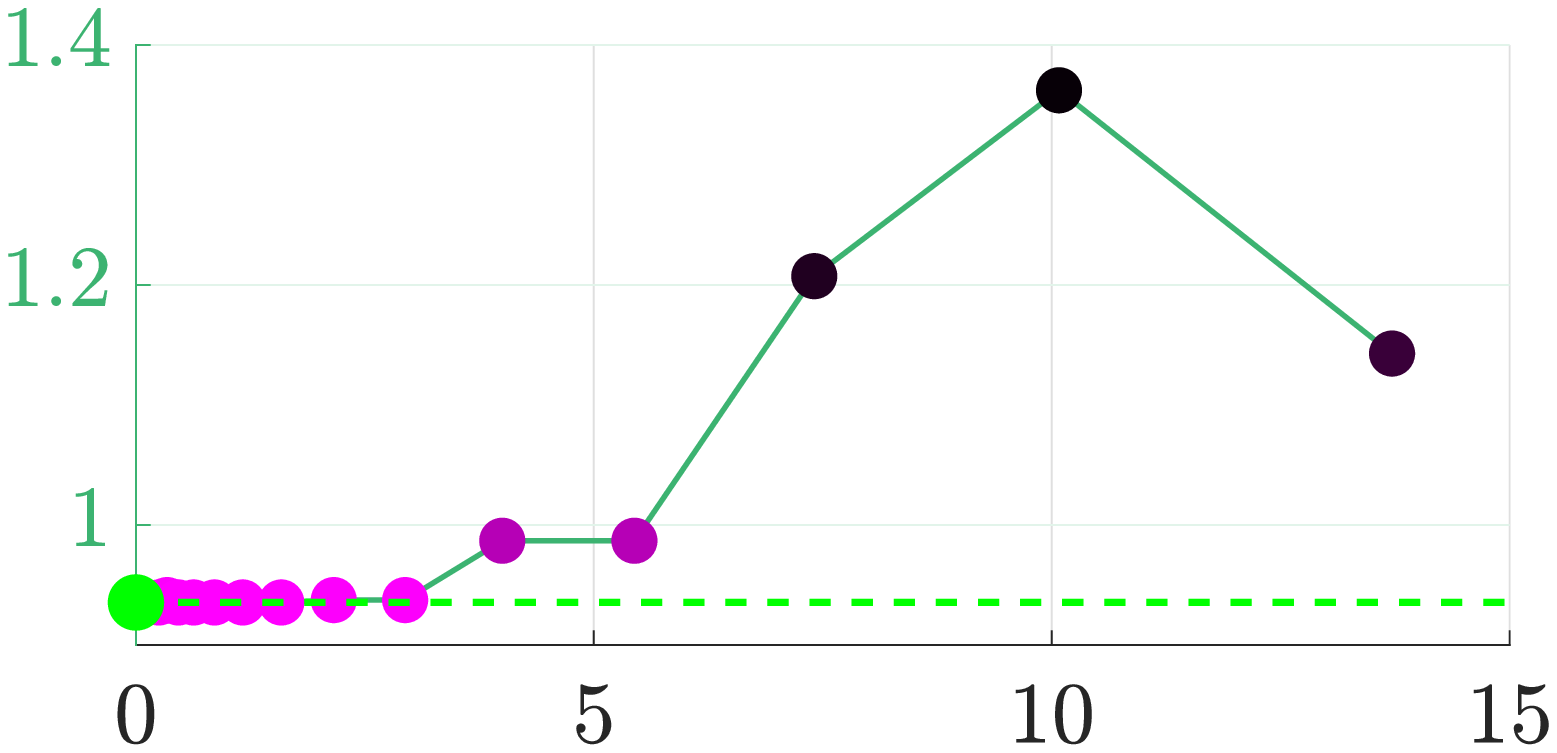} 
         \caption*{\centering \large $\Delta\sigma$}
    \end{subfigure}
    \caption{\textbf{Gaussian Noise Magnitude}. Minimum \ate\ achieved using our \gtf, estimated with varying noise levels $\Delta\sigma$, for an ablation study on GLOMAP’s hyperparameter controlling the maximum reprojection error for inliers in Bundle Adjustment (Max.~BA error) in radians. Our \gtf\ accurately identifies the optimal performance without requiring ground truth data for a specific range of noise magnitudes $\Delta\sigma$. $^\ddagger$\footnotesize Please refer to the supplementary material for extra plots and full details.}
    \label{fig:noise}
\end{figure}
Eq.~\eqref{eq:_simp_inf_matrix} assumes that the propagated input noise follows an isotropic Gaussian distribution. In line with this assumption, our methodology and experiments apply Gaussian noise directly to the grayscale intensities. This study examines different noise magnitudes to identify the configuration that achieves the strongest correlation with real ground truth. As illustrated in Figure \ref{fig:noise}, our \gtf\ effectively identifies the optimal \ate\ without relying on ground truth data within a specific range of noise magnitudes. Beyond this range, as the magnitude $\Delta\sigma$ increases, the noise starts to dominate the system response, impairing the detection of optimal accuracy.

\subsection{Comparison against Reprojection Error}
\label{subsec:reproj_error}

\begin{figure}[t]
    \centering
    \begin{subfigure}[b]{0.01\textwidth}
        \centering
         \rotatebox{90}{\parbox[t]{1.8cm}{\raggedleft \footnotesize \ate}}
    \end{subfigure}
    \hfill
    \begin{subfigure}[b]{0.21\textwidth}
        \centering
        \includegraphics[width=1.0\textwidth]{figures/fine_tuning/fineTune_glomap_50frames_Thresholds_max_reprojection_error_REPLICA_office0_fig_1.eps}  
        \caption*{\small \textbf{$\log_{10}$ Max. BA $\boldsymbol{e_r}$}}
    \end{subfigure}  
    \hfill
    \begin{subfigure}[b]{0.01\textwidth}
        \centering
         \rotatebox{90}{\parbox[t]{2cm}{\raggedleft \footnotesize \gtf}}
    \end{subfigure}
    \hfill
    \begin{subfigure}[b]{0.02\textwidth}
        \centering
        \rotatebox{90}{\parbox[t]{1.8cm}{\raggedleft  \small \textcolor{groundtruth}{\small \ate}}}
    \end{subfigure}
    \begin{subfigure}[b]{0.20\textwidth}
         \centering
         \includegraphics[width=1.0\textwidth]{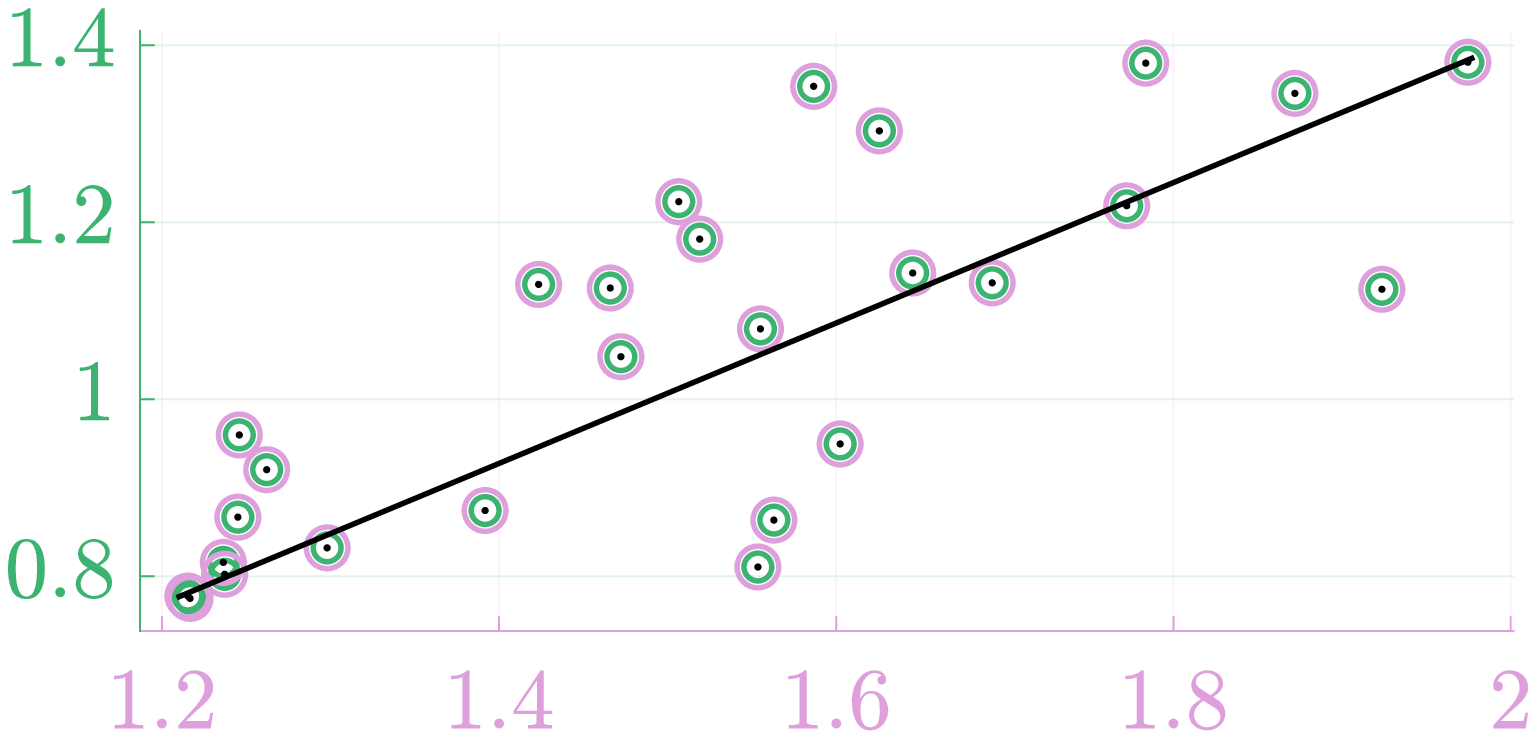}  
         \caption*{\small \textcolor{proxy}{\small \gtf}}
    \end{subfigure} 
    
    \begin{subfigure}[b]{0.01\textwidth}
        \centering
         \rotatebox{90}{\parbox[t]{1.8cm}{\raggedleft \footnotesize \ate}}
    \end{subfigure}
    \hfill
    \begin{subfigure}[b]{0.21\textwidth}
        \centering
        \includegraphics[width=1.0\textwidth]{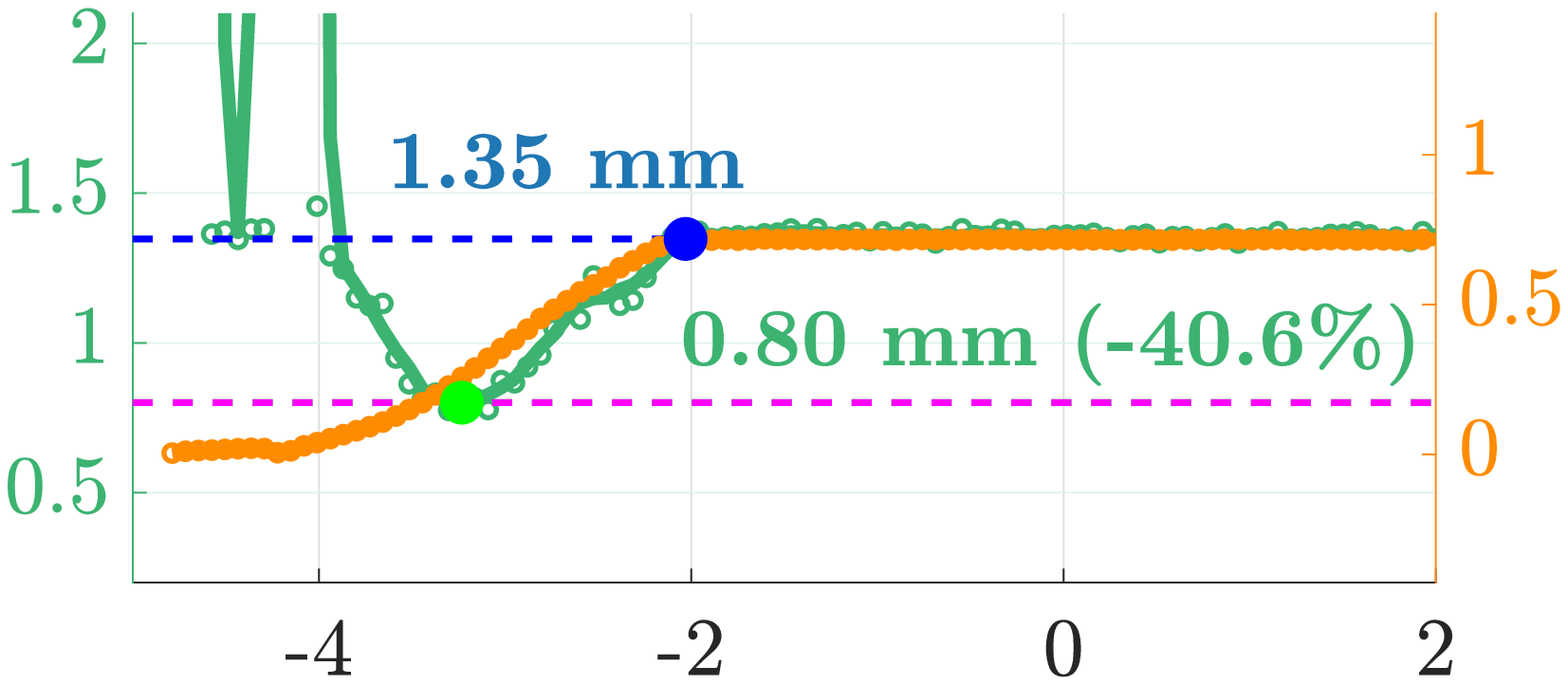}  
        \caption*{\small \textbf{$\log_{10}$ Max. BA $\boldsymbol{e_r}$}}
    \end{subfigure}  
    \hfill
    \begin{subfigure}[b]{0.01\textwidth}
        \centering
         \rotatebox{90}{\parbox[t]{2cm}{\raggedleft \footnotesize \textcolor{repr_error}{$e_{r} [px]$}}}
    \end{subfigure}
    \hfill
    \begin{subfigure}[b]{0.02\textwidth}
        \centering
        \rotatebox{90}{\parbox[t]{1.8cm}{\raggedleft  \small \textcolor{groundtruth}{\small \ate}}}
    \end{subfigure}
    \begin{subfigure}[b]{0.20\textwidth}
         \centering
         \includegraphics[width=1.0\textwidth]{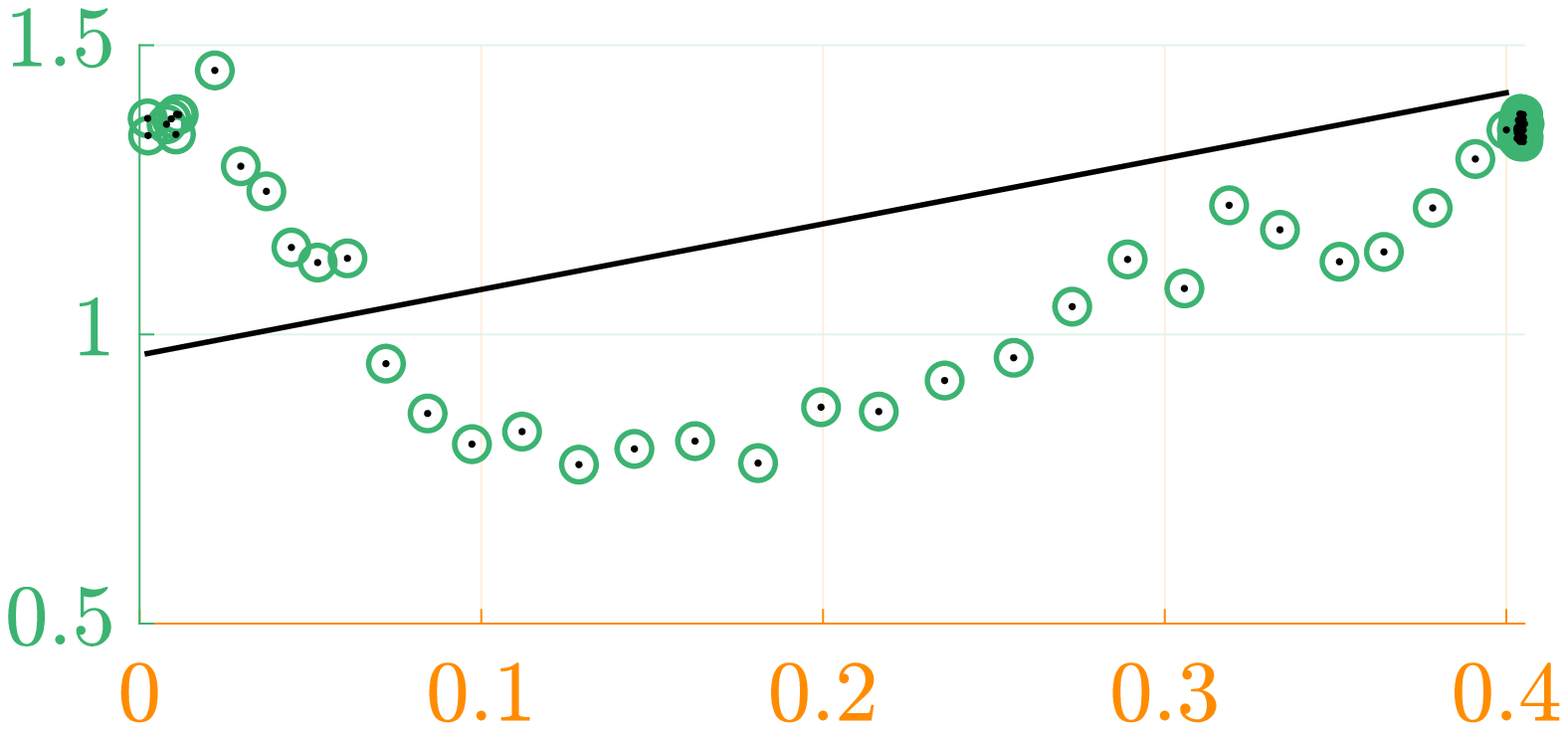}  
         \caption*{\small  \textcolor{repr_error}{$e_{r} [px]$}}
    \end{subfigure} 
    \caption{\gtf\ \textbf{vs} \textcolor{repr_error}{Reprojection Error $e_r [px]$}:  the \textbf{top} plots illustrate the correlation between our proposed ground-truth-free metric, \gtf, and the ground-truth-based \ate. The \textbf{bottom} plots show the correlation between the average reprojection error, \textcolor{repr_error}{$e_r [px]$}, and the ground-truth-based \ate. Note the strong alignment of our \gtf\ with the ground-truth-based \ate, contrasting with the weaker correlation observed with \textcolor{repr_error}{$e_r [px]$}.}
    \label{fig:rep_error}
\end{figure}

The reprojection error has commonly been used as a precision measure in SfM/VSLAM systems due to its ease of computation~\cite{schoenberger2016mvs,schoenberger2016sfm}. However, relying on optimized residuals for precision evaluation carries the risk of overfitting, leading to the trivial solution where a system with zero residuals would mistakenly be considered the most accurate.

Figure~\ref{fig:rep_error} illustrates the correlation between the averaged \textcolor{repr_error}{reprojection error $e_r$ [px]}, our \gtf, and the actual \ate\ obtained using ground truth as we vary the maximum reprojection error for inliers in the Bundle Adjustment of GLOMAP (\textbf{Max.~BA $e_r$}) in radians (see Section~\ref{subsec:finetuning_glomap}). Notably, \textcolor{repr_error}{reprojection error} correlates with \ate\ when it is large, particularly in the presence of outliers. In these cases, reducing the maximum reprojection error during Bundle Adjustment decreases both the average reprojection error and the trajectory error. However, when outliers are not a significant issue, further minimizing the reprojection error no longer aligns well with the actual \ate. By contrast, our \gtf\ exhibits a strong and consistent correlation with \ate\ across the entire ablation interval.

\subsection{Computational Cost Study}
\label{subsec:computational}

\begin{figure}[t]

    \centering

    \begin{subfigure}[b]{0.02\textwidth}
        \centering
        \rotatebox{90}{\parbox{3.0cm}{\centering $R^2$}}
    \end{subfigure}
    \begin{subfigure}[b]{0.22\textwidth}
         \centering
          \includegraphics[width=1.0\textwidth]{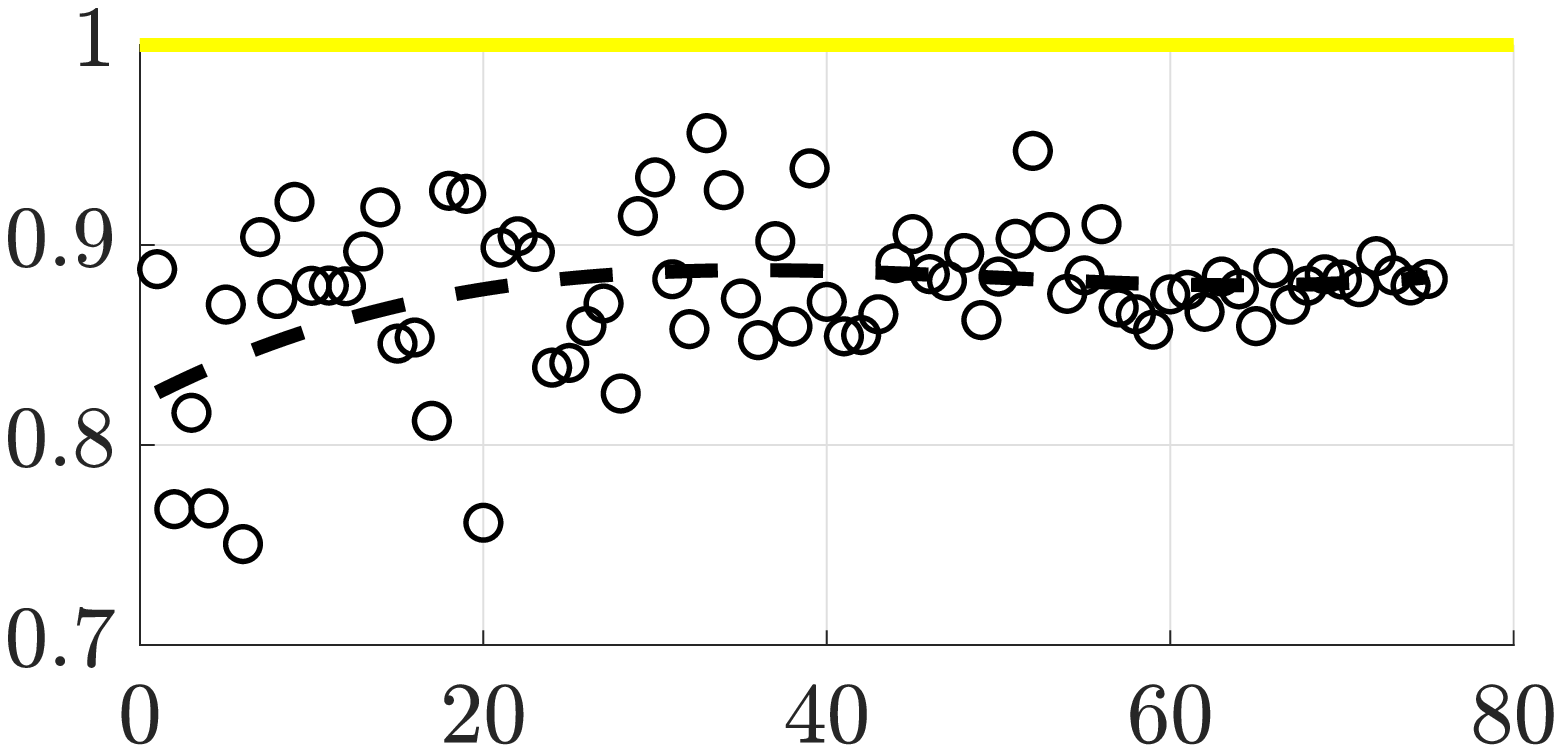} 
        \caption*{\centering \large $k_\Delta$}
    \end{subfigure} 
    \vspace{-0.7cm}
        
    \begin{subfigure}[b]{0.01\textwidth}
        \centering
        \rotatebox{90}{\parbox{2.75cm}{\centering \footnotesize \ate}}
    \end{subfigure}
    \hfill
    \begin{subfigure}[b]{0.22\textwidth}
         \centering
         \includegraphics[width=1.0\textwidth]{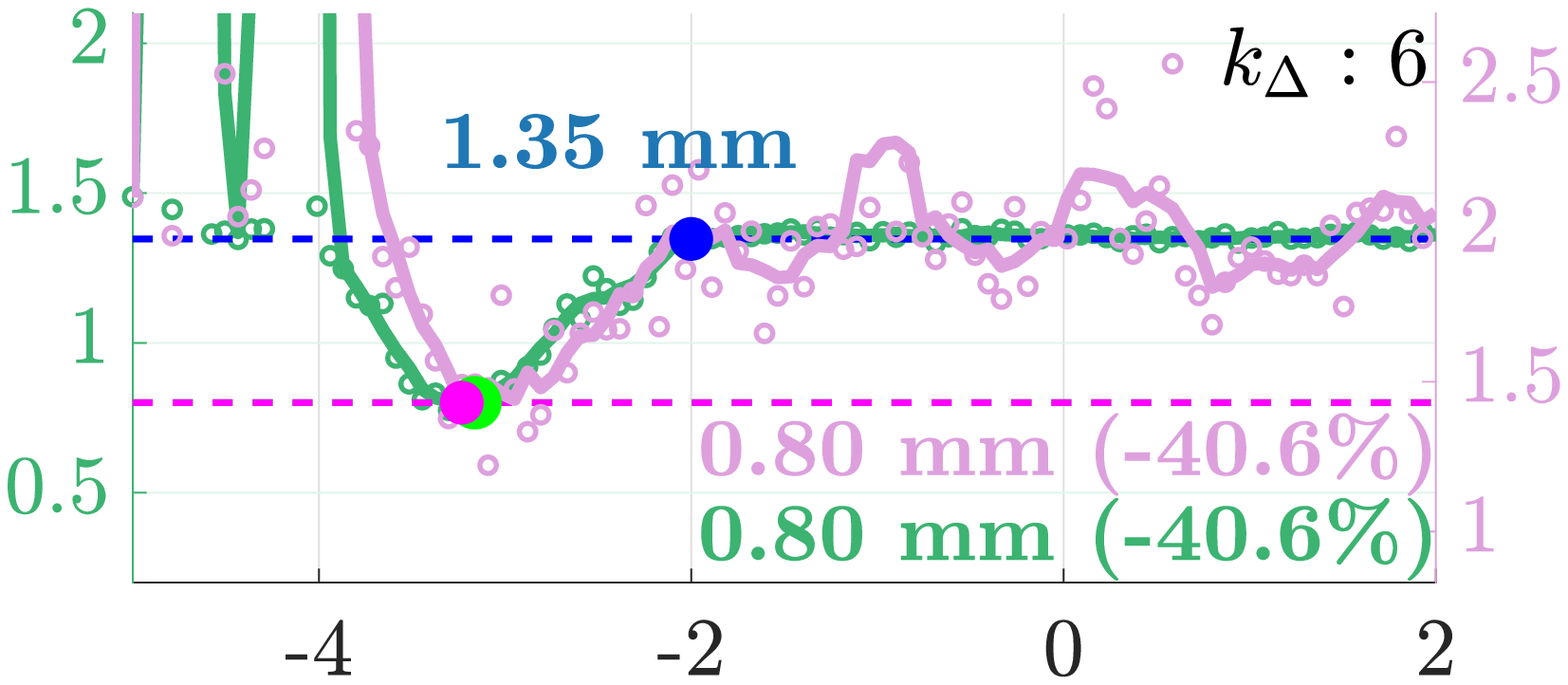} 
         \caption*{\small \textbf{$\log_{10}$ Max. BA $\boldsymbol{e_r}$}}
    \end{subfigure}
    \hfill
    \begin{subfigure}[b]{0.22\textwidth}
         \centering
         \includegraphics[width=1.0\textwidth]{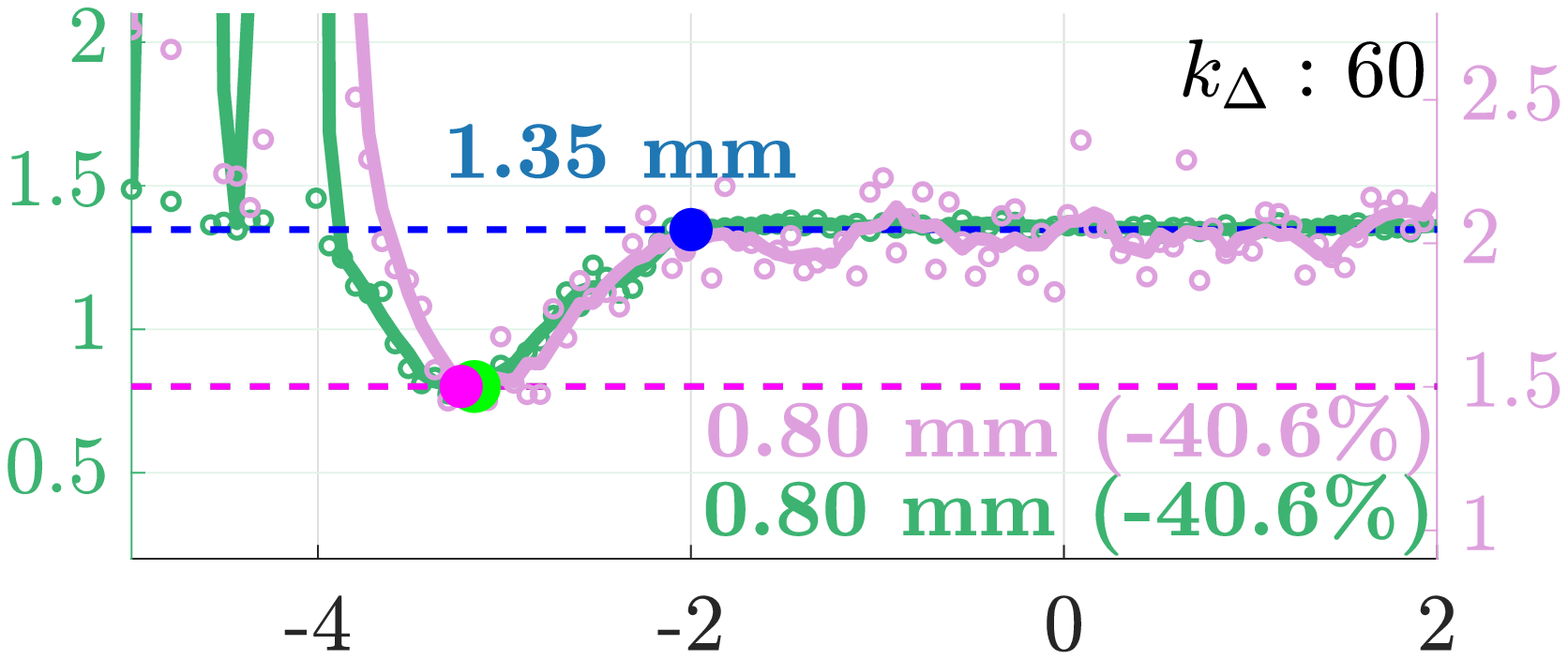} 
         \caption*{\small \textbf{$\log_{10}$ Max. BA $\boldsymbol{e_r}$}}
    \end{subfigure}
    \hfill
    \begin{subfigure}[b]{0.01\textwidth}
        \centering
        \rotatebox{90}{\parbox{2.75cm}{\centering \footnotesize \gtf}}
    \end{subfigure}
    \vspace{-1cm}
    
    \begin{subfigure}[b]{0.01\textwidth}
        \centering
        \rotatebox{90}{\parbox{3.0cm}{\centering \footnotesize \ate}}
    \end{subfigure}
    \hfill
    \begin{subfigure}[b]{0.21\textwidth}
         \centering
         \includegraphics[width=1.0\textwidth]{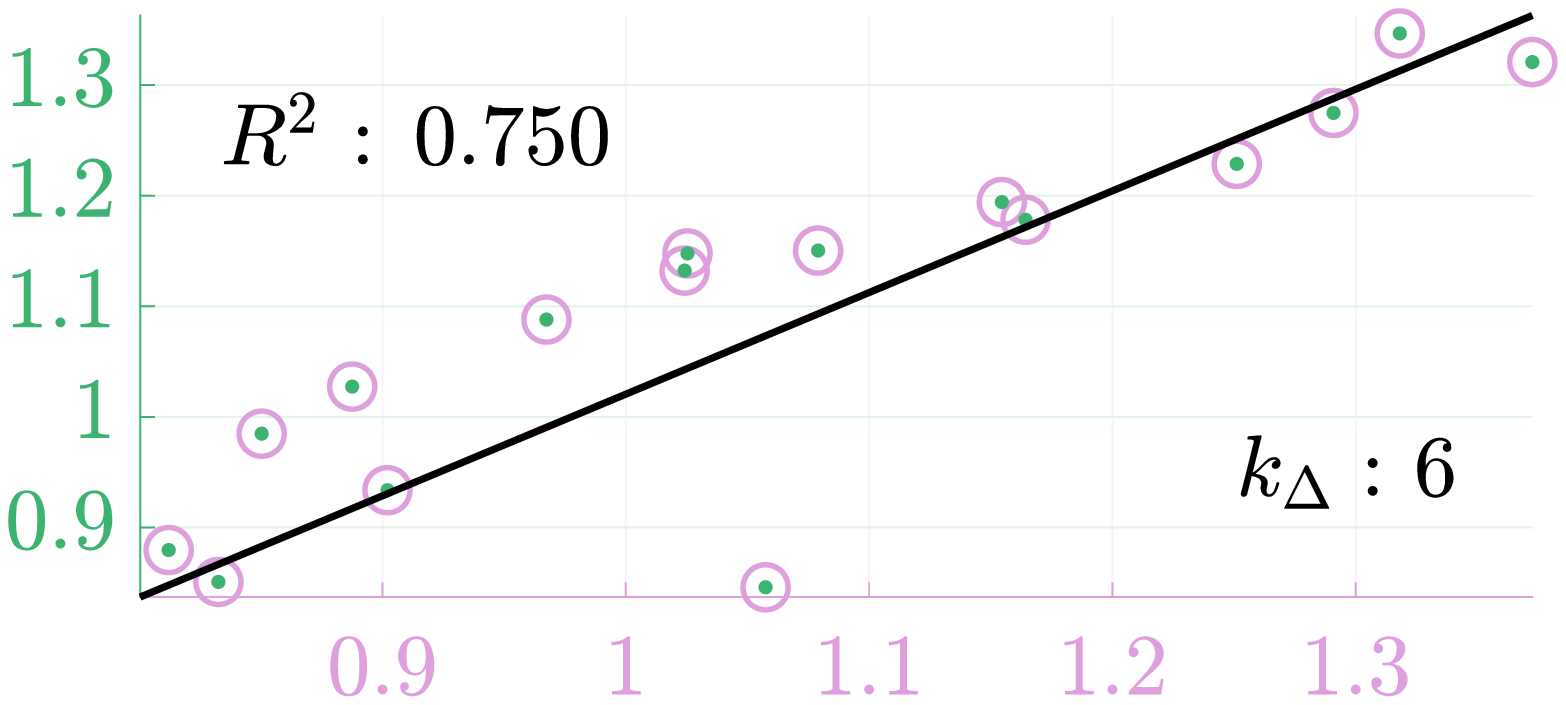} 
         \caption*{\centering \gtf}
    \end{subfigure}
    \hfill
    \begin{subfigure}[b]{0.22\textwidth}
         \centering
         \includegraphics[width=1.0\textwidth]{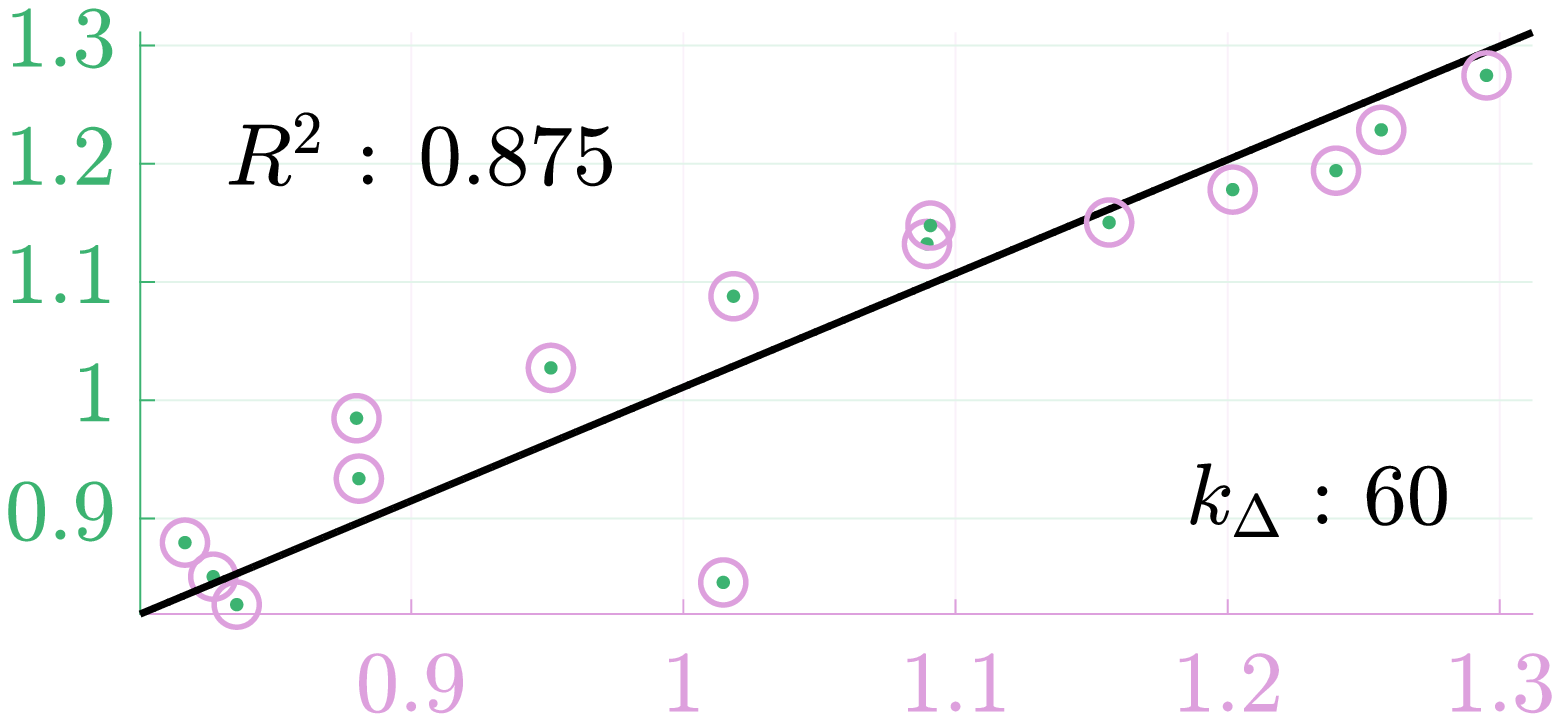} 
         \caption*{\centering \gtf}
    \end{subfigure}
    \hfill
    \begin{subfigure}[b]{0.01\textwidth}
        \centering
        \rotatebox{90}{\parbox{3.0cm}{\centering}}
    \end{subfigure}
    \caption{\textbf{Influence of the number of trajectories augmented with Gaussian noise ($k_\Delta$) on the correlation between \ate\ and \gtf}. \textbf{Top}: $R^2$ values of the regression between \ate\ and \gtf\ as the number of estimated trajectories ($k_\Delta$) with augmented Gaussian noise increases. \textbf{Middle}: Ablation of \ate\ and \gtf\ when tuning a GLOMAP hyperparameter, presented for $k_\Delta = 6$ and $k_\Delta = 60$. \textbf{Bottom}: Linear regression between \ate\ and \gtf\ for $k_\Delta = 6$ and $k_\Delta = 60$. Increasing $k_\Delta$ enhances the correlation between \ate\ and \gtf\ by reducing the impact of GLOMAP’s non-deterministic behavior.}
    
    \label{fig:computational}
\end{figure}

The computation of the \gtf, as outlined in Algorithm~\ref{alg:gtf}, involves generating $k$ trajectories to account for the non-deterministic behavior of SfM/VSLAM systems, and $k_\Delta$ trajectories to incorporate Gaussian noise augmentation applied to the images. Figure~\ref{fig:computational} illustrates that, as expected, the correlation between the actual \ate\ and our \gtf\ (computed without ground truth) improves as the number of evaluation samples increases.

The computational complexity of our approach is closely tied to the underlying SfM/VSLAM system and the specific task being performed, such as hyperparameter tuning. Generally, for a method that requires $k$ comparisons against ground truth, our approach operates with a linear complexity \( O(k \cdot k_{\Delta}) \), corresponding to the number of noisy experiments needed. Our \gtf\ eliminates the need for ground-truth data while maintaining computational feasibility.

\section{Conclusions and Future Work}
\label{sec:conclusions}
This paper is the first one demonstrating the feasibility of \textcolor{proxy}{G}round-\textcolor{proxy}{T}ruth-\textcolor{proxy}{F}ree benchmarking of SfM and VSLAM pipelines, addressing key challenges in scalability and applicability to real-world datasets. This achievement is based on the novel ideas of characterizing the sensitivity of the pipelines with respect to the noise in a particular image set by sampling several instances of such pipelines in the original and perturbed data and averaging them to smooth the noise. 

Although our methodology could be extended, in principle, to any metric, we demonstrate it here using \ate, the arguably standard metric in SfM and VSLAM. Our experimental results show a strong correlation between our \gtf\ and the standard ground-truth-based \ate, making it suitable for tasks like hyperparameter tuning and performance benchmarking.

Our ground-truth-free methodology opens new possibilities for scalable, data-driven localization and mapping, potentially enabling significant advancements in real-world applications. Future work will focus on leveraging our new metric with state-of-the-art efficient fine-tuning approaches and researching ways to build, train, and enhance VSLAM pipelines in a self-supervised and online manner. This exploration will contribute to developing scalable, adaptable VSLAM systems that continuously improve in diverse and challenging environments.

{
    \small
    \bibliographystyle{ieeenat_fullname}
    \bibliography{fontan}

\begin{thebibliography}{114}
\providecommand{\natexlab}[1]{#1}
\providecommand{\url}[1]{\texttt{#1}}
\expandafter\ifx\csname urlstyle\endcsname\relax
  \providecommand{\doi}[1]{doi: #1}\else
  \providecommand{\doi}{doi: \begingroup \urlstyle{rm}\Url}\fi

\bibitem[Agarwal et~al.(2020)Agarwal, Vora, Pandey, Williams, Kourous, and McBride]{agarwal2020ford}
Siddharth Agarwal, Ankit Vora, Gaurav Pandey, Wayne Williams, Helen Kourous, and James McBride.
\newblock Ford multi-av seasonal dataset.
\newblock \emph{The International Journal of Robotics Research}, 39\penalty0 (12):\penalty0 1367--1376, 2020.

\bibitem[Ahmadyan et~al.(2021)Ahmadyan, Zhang, Ablavatski, Wei, and Grundmann]{ahmadyan2021objectron}
Adel Ahmadyan, Liangkai Zhang, Artsiom Ablavatski, Jianing Wei, and Matthias Grundmann.
\newblock Objectron: A large scale dataset of object-centric videos in the wild with pose annotations.
\newblock In \emph{Proceedings of the IEEE/CVF conference on computer vision and pattern recognition}, pages 7822--7831, 2021.

\bibitem[Amigoni et~al.(2007)Amigoni, Gasparini, and Gini]{amigoni2007good}
Francesco Amigoni, Simone Gasparini, and Maria Gini.
\newblock Good experimental methodologies for robotic mapping: A proposal.
\newblock In \emph{Proceedings of the IEEE International Conference on Robotics and Automation}, pages 4176--4181, 2007.

\bibitem[Andonie(2019)]{andonie2019hyperparameter}
R{\u{a}}zvan Andonie.
\newblock Hyperparameter optimization in learning systems.
\newblock \emph{Journal of Membrane Computing}, 1\penalty0 (4):\penalty0 279--291, 2019.

\bibitem[Balaguer et~al.(2007)Balaguer, Carpin, and Balakirsky]{balaguer2007towards}
Benjamin Balaguer, Stefano Carpin, and Stephen Balakirsky.
\newblock Towards quantitative comparisons of robot algorithms: Experiences with {SLAM} in simulation and real world systems.
\newblock In \emph{IEEE/RSJ International Conference on Intelligent Robots and Systems Workshops}, 2007.

\bibitem[Bao and Savarese(2011)]{bao2011semantic}
Sid~Yingze Bao and Silvio Savarese.
\newblock Semantic structure from motion.
\newblock In \emph{IEEE/CVF Conference on Computer Vision and Pattern Recognition}, pages 2025--2032, 2011.

\bibitem[Barfoot(2024)]{barfoot2024state}
Timothy~D Barfoot.
\newblock \emph{State estimation for robotics}.
\newblock Cambridge University Press, 2024.

\bibitem[Barron et~al.(2022)Barron, Mildenhall, Verbin, Srinivasan, and Hedman]{barron2022mip}
Jonathan~T Barron, Ben Mildenhall, Dor Verbin, Pratul~P Srinivasan, and Peter Hedman.
\newblock Mip-nerf 360: Unbounded anti-aliased neural radiance fields.
\newblock In \emph{Proceedings of the IEEE/CVF Conference on Computer Vision and Pattern Recognition}, pages 5470--5479, 2022.

\bibitem[Bonarini et~al.(2006)Bonarini, Burgard, Fontana, Matteucci, Sorrenti, Tardos, et~al.]{bonarini2006rawseeds}
Andrea Bonarini, Wolfram Burgard, Giulio Fontana, Matteo Matteucci, Domenico~Giorgio Sorrenti, Juan~Domingo Tardos, et~al.
\newblock Rawseeds: Robotics advancement through web-publishing of sensorial and elaborated extensive data sets.
\newblock In \emph{IEEE/RSJ International Conference on Intelligent Robots and Systems}, page~93, 2006.

\bibitem[Burgard et~al.(2009)Burgard, Stachniss, Grisetti, Steder, K{\"u}mmerle, Dornhege, Ruhnke, Kleiner, and Tard{\"o}s]{burgard2009comparison}
Wolfram Burgard, Cyrill Stachniss, Giorgio Grisetti, Bastian Steder, Rainer K{\"u}mmerle, Christian Dornhege, Michael Ruhnke, Alexander Kleiner, and Juan~D Tard{\"o}s.
\newblock A comparison of slam algorithms based on a graph of relations.
\newblock In \emph{2009 IEEE/RSJ International Conference on Intelligent Robots and Systems}, pages 2089--2095, 2009.

\bibitem[Burri et~al.(2016)Burri, Nikolic, Gohl, Schneider, Rehder, Omari, Achtelik, and Siegwart]{burri2016euroc}
Michael Burri, Janosch Nikolic, Pascal Gohl, Thomas Schneider, Joern Rehder, Sammy Omari, Markus~W Achtelik, and Roland Siegwart.
\newblock The euroc micro aerial vehicle datasets.
\newblock \emph{The International Journal of Robotics Research}, 35\penalty0 (10):\penalty0 1157--1163, 2016.

\bibitem[Cabon et~al.(2020)Cabon, Murray, and Humenberger]{cabon2020vkitti2}
Yohann Cabon, Naila Murray, and Martin Humenberger.
\newblock Virtual kitti 2, 2020.

\bibitem[Cadena et~al.(2016)Cadena, Carlone, Carrillo, Latif, Scaramuzza, Neira, Reid, and Leonard]{cadena2016past}
Cesar Cadena, Luca Carlone, Henry Carrillo, Yasir Latif, Davide Scaramuzza, Jos{\'e} Neira, Ian Reid, and John~J Leonard.
\newblock Past, present, and future of simultaneous localization and mapping: Toward the robust-perception age.
\newblock \emph{IEEE Transactions on Robotics}, 32\penalty0 (6):\penalty0 1309--1332, 2016.

\bibitem[Campos et~al.(2021)Campos, Elvira, Rodr{\'\i}guez, Montiel, and Tard{\'o}s]{campos2021orb}
Carlos Campos, Richard Elvira, Juan J~G{\'o}mez Rodr{\'\i}guez, Jos{\'e}~MM Montiel, and Juan~D Tard{\'o}s.
\newblock Orb-slam3: An accurate open-source library for visual, visual--inertial, and multimap slam.
\newblock \emph{IEEE Transactions on Robotics}, 37\penalty0 (6):\penalty0 1874--1890, 2021.

\bibitem[Choi et~al.(2015)Choi, Zhou, and Koltun]{Choi_2015_CVPR}
Sungjoon Choi, Qian-Yi Zhou, and Vladlen Koltun.
\newblock Robust reconstruction of indoor scenes.
\newblock In \emph{IEEE Conference on Computer Vision and Pattern Recognition}, 2015.

\bibitem[Cordts et~al.(2016)Cordts, Omran, Ramos, Rehfeld, Enzweiler, Benenson, Franke, Roth, and Schiele]{cordts2016cityscapes}
Marius Cordts, Mohamed Omran, Sebastian Ramos, Timo Rehfeld, Markus Enzweiler, Rodrigo Benenson, Uwe Franke, Stefan Roth, and Bernt Schiele.
\newblock The cityscapes dataset for semantic urban scene understanding.
\newblock In \emph{Proceedings of the IEEE Conference on Computer Vision and Pattern Recognition}, pages 3213--3223, 2016.

\bibitem[Dai et~al.(2017)Dai, Chang, Savva, Halber, Funkhouser, and Nie{\ss}ner]{dai2017scannet}
Angela Dai, Angel~X Chang, Manolis Savva, Maciej Halber, Thomas Funkhouser, and Matthias Nie{\ss}ner.
\newblock Scannet: Richly-annotated 3d reconstructions of indoor scenes.
\newblock In \emph{Proceedings of the IEEE Conference on Computer Vision and Pattern Recognition}, pages 5828--5839, 2017.

\bibitem[Davison(2018)]{davison2018futuremapping}
Andrew~J Davison.
\newblock {FutureMapping: The computational structure of spatial AI systems}.
\newblock \emph{arXiv preprint arXiv:1803.11288}, 2018.

\bibitem[Davison et~al.(2007)Davison, Reid, Molton, and Stasse]{davison2007monoslam}
Andrew~J Davison, Ian~D Reid, Nicholas~D Molton, and Olivier Stasse.
\newblock {MonoSLAM: Real-time single camera SLAM}.
\newblock \emph{IEEE Transactions on Pattern Analysis and Machine Intelligence}, 29\penalty0 (6):\penalty0 1052--1067, 2007.

\bibitem[Deng et~al.(2009)Deng, Dong, Socher, Li, Li, and Fei-Fei]{deng2009imagenet}
Jia Deng, Wei Dong, Richard Socher, Li-Jia Li, Kai Li, and Li Fei-Fei.
\newblock Imagenet: A large-scale hierarchical image database.
\newblock In \emph{Proceedings of the IEEE Conference on Computer Vision and Pattern Recognition}, pages 248--255, 2009.

\bibitem[Deng et~al.(2023)Deng, Wu, Chen, Xia, Sun, Liu, Yu, and Pei]{deng2023nerfloam}
Junyuan Deng, Qi Wu, Xieyuanli Chen, Songpengcheng Xia, Zhen Sun, Guoqing Liu, Wenxian Yu, and Ling Pei.
\newblock Nerf-loam: Neural implicit representation for large-scale incremental lidar odometry and mapping.
\newblock In \emph{Proceedings of the IEEE/CVF International Conference on Computer Vision}, 2023.

\bibitem[Diba et~al.(2020)Diba, Fayyaz, Sharma, Paluri, Gall, Stiefelhagen, and Van~Gool]{diba2020large}
Ali Diba, Mohsen Fayyaz, Vivek Sharma, Manohar Paluri, J{\"u}rgen Gall, Rainer Stiefelhagen, and Luc Van~Gool.
\newblock Large scale holistic video understanding.
\newblock In \emph{European Conference on Computer Vision}, pages 593--610. Springer, 2020.

\bibitem[Duisterhof et~al.(2024)Duisterhof, Zust, Weinzaepfel, Leroy, Cabon, and Revaud]{duisterhof2024mast3r}
Bardienus Duisterhof, Lojze Zust, Philippe Weinzaepfel, Vincent Leroy, Yohann Cabon, and Jerome Revaud.
\newblock Mast3r-sfm: a fully-integrated solution for unconstrained structure-from-motion.
\newblock \emph{arXiv preprint arXiv:2409.19152}, 2024.

\bibitem[Engel et~al.(2014)Engel, Sch{\"o}ps, and Cremers]{engel2014lsd}
Jakob Engel, Thomas Sch{\"o}ps, and Daniel Cremers.
\newblock {LSD-SLAM: Large-scale direct monocular SLAM}.
\newblock In \emph{European Conference on Computer Vision}, pages 834--849, 2014.

\bibitem[Engel et~al.(2016)Engel, Usenko, and Cremers]{engel2016photometrically}
Jakob Engel, Vladyslav Usenko, and Daniel Cremers.
\newblock A photometrically calibrated benchmark for monocular visual odometry.
\newblock \emph{arXiv preprint arXiv:1607.02555}, 2016.

\bibitem[Engel et~al.(2017)Engel, Koltun, and Cremers]{engel2017direct}
Jakob Engel, Vladlen Koltun, and Daniel Cremers.
\newblock Direct sparse odometry.
\newblock \emph{IEEE Transactions on Pattern Analysis and Machine Intelligence}, 40\penalty0 (3):\penalty0 611--625, 2017.

\bibitem[Fontan et~al.(2023{\natexlab{a}})Fontan, Civera, and Milford]{fontan2023motion}
Alejandro Fontan, Javier Civera, and Michael Milford.
\newblock Motion-bias-free feature-based slam.
\newblock In \emph{British Machine Vision Conference}, 2023{\natexlab{a}}.

\bibitem[Fontan et~al.(2023{\natexlab{b}})Fontan, Giubilato, Oliva, Civera, and Triebel]{fontan2023sid}
Alejandro Fontan, Riccardo Giubilato, Laura Oliva, Javier Civera, and Rudolph Triebel.
\newblock Sid-slam: Semi-direct information-driven rgb-d slam.
\newblock \emph{IEEE Robotics and Automation Letters}, 2023{\natexlab{b}}.

\bibitem[Fontan et~al.(2024)Fontan, Civera, and Milford]{fontanRSS2024}
Alejandro Fontan, Javier Civera, and Michael Milford.
\newblock {AnyFeature-VSLAM}: Automating the usage of any chosen feature into visual slam.
\newblock In \emph{Robotics: Science and Systems}, 2024.

\bibitem[Forster et~al.(2014)Forster, Pizzoli, and Scaramuzza]{forster2014svo}
Christian Forster, Matia Pizzoli, and Davide Scaramuzza.
\newblock Svo: Fast semi-direct monocular visual odometry.
\newblock In \emph{2014 IEEE international conference on robotics and automation (ICRA)}, pages 15--22, 2014.

\bibitem[Forster et~al.(2016)Forster, Zhang, Gassner, Werlberger, and Scaramuzza]{forster2016svo}
Christian Forster, Zichao Zhang, Michael Gassner, Manuel Werlberger, and Davide Scaramuzza.
\newblock Svo: Semidirect visual odometry for monocular and multicamera systems.
\newblock \emph{IEEE Transactions on Robotics}, 33\penalty0 (2):\penalty0 249--265, 2016.

\bibitem[Gaidon et~al.(2016)Gaidon, Wang, Cabon, and Vig]{Gaidon:Virtual:CVPR2016}
A Gaidon, Q Wang, Y Cabon, and E Vig.
\newblock Virtual worlds as proxy for multi-object tracking analysis.
\newblock In \emph{IEEE/CVF Conference on Computer Vision and Pattern Recognition}, 2016.

\bibitem[Gangopadhyay et~al.(2024)Gangopadhyay, Chen, Chu, Rim, Park, and Wong]{gangopadhyay2024uncle}
Suchisrit Gangopadhyay, Xien Chen, Michael Chu, Patrick Rim, Hyoungseob Park, and Alex Wong.
\newblock Uncle: Unsupervised continual learning of depth completion.
\newblock \emph{arXiv preprint arXiv:2410.18074}, 2024.

\bibitem[Geiger et~al.(2012)Geiger, Lenz, and Urtasun]{geiger2012we}
Andreas Geiger, Philip Lenz, and Raquel Urtasun.
\newblock Are we ready for autonomous driving? the kitti vision benchmark suite.
\newblock In \emph{IEEE Conference on Computer Vision and Pattern Recognition}, pages 3354--3361, 2012.

\bibitem[Geiger et~al.(2013)Geiger, Lenz, Stiller, and Urtasun]{geiger2013vision}
Andreas Geiger, Philip Lenz, Christoph Stiller, and Raquel Urtasun.
\newblock Vision meets robotics: The kitti dataset.
\newblock \emph{The International Journal of Robotics Research}, 32\penalty0 (11):\penalty0 1231--1237, 2013.

\bibitem[Giubilato et~al.(2022)Giubilato, St{\"u}rzl, Wedler, and Triebel]{giubilato2022challenges}
Riccardo Giubilato, Wolfgang St{\"u}rzl, Armin Wedler, and Rudolph Triebel.
\newblock Challenges of slam in extremely unstructured environments: The dlr planetary stereo, solid-state lidar, inertial dataset.
\newblock \emph{IEEE Robotics and Automation Letters}, 7\penalty0 (4):\penalty0 8721--8728, 2022.

\bibitem[Glocker et~al.(2013)Glocker, Izadi, Shotton, and Criminisi]{glocker2013real}
Ben Glocker, Shahram Izadi, Jamie Shotton, and Antonio Criminisi.
\newblock Real-time rgb-d camera relocalization.
\newblock In \emph{IEEE International Symposium on Mixed and Augmented Reality}, pages 173--179, 2013.

\bibitem[Grauman et~al.(2022)Grauman, Westbury, Byrne, Chavis, Furnari, Girdhar, Hamburger, Jiang, Liu, Liu, et~al.]{grauman2022ego4d}
Kristen Grauman, Andrew Westbury, Eugene Byrne, Zachary Chavis, Antonino Furnari, Rohit Girdhar, Jackson Hamburger, Hao Jiang, Miao Liu, Xingyu Liu, et~al.
\newblock Ego4d: Around the world in 3,000 hours of egocentric video.
\newblock In \emph{Proceedings of the IEEE/CVF Conference on Computer Vision and Pattern Recognition}, pages 18995--19012, 2022.

\bibitem[Handa et~al.(2010)Handa, Chli, Strasdat, and Davison]{handa2010scalable}
Ankur Handa, Margarita Chli, Hauke Strasdat, and Andrew~J Davison.
\newblock Scalable active matching.
\newblock In \emph{IEEE Computer Society Conference on Computer Vision and Pattern Recognition}, pages 1546--1553, 2010.

\bibitem[Handa et~al.(2014)Handa, Whelan, McDonald, and Davison]{handa2014benchmark}
Ankur Handa, Thomas Whelan, John McDonald, and Andrew~J Davison.
\newblock {A benchmark for RGB-D visual odometry, 3D reconstruction and SLAM}.
\newblock In \emph{IEEE International Conference on Robotics and Automation}, 2014.

\bibitem[Helmberger et~al.(2022)Helmberger, Morin, Berner, Kumar, Cioffi, and Scaramuzza]{helmberger2022hilti}
Michael Helmberger, Kristian Morin, Beda Berner, Nitish Kumar, Giovanni Cioffi, and Davide Scaramuzza.
\newblock The hilti slam challenge dataset.
\newblock \emph{IEEE Robotics and Automation Letters}, 7\penalty0 (3):\penalty0 7518--7525, 2022.

\bibitem[Horn(1987)]{horn1987closed}
Berthold~KP Horn.
\newblock Closed-form solution of absolute orientation using unit quaternions.
\newblock \emph{Journal of the Optical Society of America A}, 4\penalty0 (4):\penalty0 629--642, 1987.

\bibitem[Hua et~al.(2024)Hua, Bai, Cao, Liu, Tao, and Wang]{hua2024hi}
Tongyan Hua, Haotian Bai, Zidong Cao, Ming Liu, Dacheng Tao, and Lin Wang.
\newblock Hi-map: Hierarchical factorized radiance field for high-fidelity monocular dense mapping.
\newblock \emph{arXiv preprint arXiv:2401.03203}, 2024.

\bibitem[Humphries et~al.(2023)Humphries, Horne, Olsen, Dunbabin, and Tranter]{humphries2023uncrewed}
Fran Humphries, Rachel Horne, Melanie Olsen, Matthew Dunbabin, and Kieran Tranter.
\newblock Uncrewed autonomous marine vessels test the limits of maritime safety frameworks.
\newblock \emph{WMU Journal of Maritime Affairs}, 22\penalty0 (3):\penalty0 317--344, 2023.

\bibitem[Judd and Gammell(2019)]{judd2019oxford}
Kevin~Michael Judd and Jonathan~D Gammell.
\newblock {The Oxford multimotion dataset: Multiple SE(3) motions with ground truth}.
\newblock \emph{IEEE Robotics and Automation Letters}, 4\penalty0 (2):\penalty0 800--807, 2019.

\bibitem[Judd et~al.(2018)Judd, Gammell, and Newman]{judd2018multimotion}
Kevin~M Judd, Jonathan~D Gammell, and Paul Newman.
\newblock Multimotion visual odometry (mvo): Simultaneous estimation of camera and third-party motions.
\newblock In \emph{IEEE/RSJ International Conference on Intelligent Robots and Systems}, pages 3949--3956, 2018.

\bibitem[Kerl et~al.(2013)Kerl, Sturm, and Cremers]{kerl2013robust}
Christian Kerl, J{\"u}rgen Sturm, and Daniel Cremers.
\newblock Robust odometry estimation for {RGB-D} cameras.
\newblock In \emph{IEEE international conference on robotics and automation}, pages 3748--3754, 2013.

\bibitem[K{\"u}mmerle et~al.(2009)K{\"u}mmerle, Steder, Dornhege, Ruhnke, Grisetti, Stachniss, and Kleiner]{kummerle2009measuring}
Rainer K{\"u}mmerle, Bastian Steder, Christian Dornhege, Michael Ruhnke, Giorgio Grisetti, Cyrill Stachniss, and Alexander Kleiner.
\newblock On measuring the accuracy of slam algorithms.
\newblock \emph{Autonomous Robots}, 27:\penalty0 387--407, 2009.

\bibitem[K{\"u}mmerle et~al.(2011)K{\"u}mmerle, Grisetti, Strasdat, Konolige, and Burgard]{kummerle2011g}
Rainer K{\"u}mmerle, Giorgio Grisetti, Hauke Strasdat, Kurt Konolige, and Wolfram Burgard.
\newblock g2o: A general framework for graph optimization.
\newblock In \emph{IEEE International Conference on Robotics and Automation}, pages 3607--3613, 2011.

\bibitem[Lamarca et~al.(2020)Lamarca, Parashar, Bartoli, and Montiel]{lamarca2020defslam}
Jose Lamarca, Shaifali Parashar, Adrien Bartoli, and JMM Montiel.
\newblock Defslam: Tracking and mapping of deforming scenes from monocular sequences.
\newblock \emph{IEEE Transactions on Robotics}, 37\penalty0 (1):\penalty0 291--303, 2020.

\bibitem[Lee and Civera(2024{\natexlab{a}})]{lee2022s}
Seong~Hun Lee and Javier Civera.
\newblock What's wrong with the absolute trajectory error?
\newblock In \emph{Proceedings of the European Conference on Computer Vision Workshops}, 2024{\natexlab{a}}.

\bibitem[Lee and Civera(2024{\natexlab{b}})]{lee2024alignment}
Seong~Hun Lee and Javier Civera.
\newblock Alignment scores: Robust metrics for multiview pose accuracy evaluation.
\newblock \emph{arXiv preprint arXiv:2407.20391}, 2024{\natexlab{b}}.

\bibitem[Li et~al.(2023)Li, Gu, Yuan, Yang, Dong, and Tan]{li2023dense}
Heng Li, Xiaodong Gu, Weihao Yuan, Luwei Yang, Zilong Dong, and Ping Tan.
\newblock Dense {RGB SLAM} with neural implicit maps.
\newblock In \emph{Proceedings of the International Conference on Learning Representations}, 2023.

\bibitem[Li et~al.(2024)Li, He, Jiang, and Wang]{li2024ddn}
Mingrui Li, Jiaming He, Guangan Jiang, and Hongyu Wang.
\newblock Ddn-slam: Real-time dense dynamic neural implicit slam with joint semantic encoding.
\newblock \emph{arXiv preprint arXiv:2401.01545}, 2024.

\bibitem[Lin et~al.(2014)Lin, Maire, Belongie, Hays, Perona, Ramanan, Doll{\'a}r, and Zitnick]{lin2014microsoft}
Tsung-Yi Lin, Michael Maire, Serge Belongie, James Hays, Pietro Perona, Deva Ramanan, Piotr Doll{\'a}r, and C~Lawrence Zitnick.
\newblock Microsoft coco: Common objects in context.
\newblock In \emph{European Conference on Computer Vision}, pages 740--755. Springer, 2014.

\bibitem[Lipson et~al.(2025)Lipson, Teed, and Deng]{lipson2025deep}
Lahav Lipson, Zachary Teed, and Jia Deng.
\newblock Deep patch visual slam.
\newblock In \emph{European Conference on Computer Vision}, pages 424--440. Springer, 2025.

\bibitem[Maddern et~al.(2017)Maddern, Pascoe, Linegar, and Newman]{maddern20171}
Will Maddern, Geoffrey Pascoe, Chris Linegar, and Paul Newman.
\newblock 1 year, 1000 km: The oxford robotcar dataset.
\newblock \emph{The International Journal of Robotics Research}, 36\penalty0 (1):\penalty0 3--15, 2017.

\bibitem[Mallios et~al.(2017)Mallios, Vidal, Campos, and Carreras]{mallios2017underwater}
Angelos Mallios, Eduard Vidal, Ricard Campos, and Marc Carreras.
\newblock Underwater caves sonar data set.
\newblock \emph{The International Journal of Robotics Research}, 36\penalty0 (12):\penalty0 1247--1251, 2017.

\bibitem[Matsuki et~al.(2024)Matsuki, Murai, Kelly, and Davison]{matsuki2024gaussian}
Hidenobu Matsuki, Riku Murai, Paul~HJ Kelly, and Andrew~J Davison.
\newblock Gaussian splatting slam.
\newblock In \emph{Proceedings of the IEEE/CVF Conference on Computer Vision and Pattern Recognition}, pages 18039--18048, 2024.

\bibitem[Meyer et~al.(2021)Meyer, Sm{\'\i}{\v{s}}ek, Fontan~Villacampa, Oliva~Maza, Medina, Schuster, Steidle, Vayugundla, M{\"u}ller, Rebele, et~al.]{meyer2021madmax}
Lukas Meyer, Michal Sm{\'\i}{\v{s}}ek, Alejandro Fontan~Villacampa, Laura Oliva~Maza, Daniel Medina, Martin~J Schuster, Florian Steidle, Mallikarjuna Vayugundla, Marcus~G M{\"u}ller, Bernhard Rebele, et~al.
\newblock The {MADMAX} data set for visual-inertial rover navigation on mars.
\newblock \emph{Journal of Field Robotics}, 38\penalty0 (6):\penalty0 833--853, 2021.

\bibitem[Miller et~al.(2018)Miller, Chung, and Hutchinson]{canoeDataset}
Martin Miller, Soon-Jo Chung, and Seth Hutchinson.
\newblock The visual–inertial canoe dataset.
\newblock \emph{The International Journal of Robotics Research}, 37\penalty0 (1):\penalty0 13--20, 2018.

\bibitem[Morlana et~al.(2024)Morlana, Tard{\'o}s, and Montiel]{morlana2024topological}
Javier Morlana, Juan~D Tard{\'o}s, and Jos{\'e}~MM Montiel.
\newblock Topological slam in colonoscopies leveraging deep features and topological priors.
\newblock In \emph{International Conference on Medical Image Computing and Computer-Assisted Intervention}, pages 733--743. Springer, 2024.

\bibitem[Moulon et~al.(2016)Moulon, Monasse, Perrot, and Marlet]{moulon2016openmvg}
Pierre Moulon, Pascal Monasse, Romuald Perrot, and Renaud Marlet.
\newblock Open{MVG}: Open multiple view geometry.
\newblock In \emph{International Workshop on Reproducible Research in Pattern Recognition}, pages 60--74. Springer, 2016.

\bibitem[Mountney et~al.(2010)Mountney, Stoyanov, and Yang]{mountney2010three}
Peter Mountney, Danail Stoyanov, and Guang-Zhong Yang.
\newblock Three-dimensional tissue deformation recovery and tracking.
\newblock \emph{IEEE Signal Processing Magazine}, 27\penalty0 (4):\penalty0 14--24, 2010.

\bibitem[Mur-Artal and Tard{\'o}s(2017)]{mur2017orb}
Raul Mur-Artal and Juan~D Tard{\'o}s.
\newblock Orb-slam2: An open-source slam system for monocular, stereo, and rgb-d cameras.
\newblock \emph{IEEE Transactions on Robotics}, 33\penalty0 (5):\penalty0 1255--1262, 2017.

\bibitem[Mur-Artal et~al.(2015)Mur-Artal, Montiel, and Tardos]{mur2015orb}
Raul Mur-Artal, Jose Maria~Martinez Montiel, and Juan~D Tardos.
\newblock Orb-slam: a versatile and accurate monocular slam system.
\newblock \emph{IEEE Transactions on Robotics}, 31\penalty0 (5):\penalty0 1147--1163, 2015.

\bibitem[Olson and Kaess(2009)]{olson2009evaluating}
Edwin Olson and Michael Kaess.
\newblock Evaluating the performance of map optimization algorithms.
\newblock In \emph{RSS Workshop on Good Experimental Methodology in Robotics}, page~35, 2009.

\bibitem[Pan et~al.(2024)Pan, Bar{\'a}th, Pollefeys, and Sch{\"o}nberger]{pan2024global}
Linfei Pan, D{\'a}niel Bar{\'a}th, Marc Pollefeys, and Johannes~L Sch{\"o}nberger.
\newblock Global structure-from-motion revisited.
\newblock In \emph{European Conference on Computer Vision}, 2024.

\bibitem[Pire et~al.(2019)Pire, Mujica, Civera, and Kofman]{pire2019rosario}
Taih{\'u} Pire, Mart{\'i}n Mujica, Javier Civera, and Ernesto Kofman.
\newblock The rosario dataset: Multisensor data for localization and mapping in agricultural environments.
\newblock \emph{The International Journal of Robotics Research}, 38\penalty0 (6):\penalty0 633--641, 2019.

\bibitem[Placed et~al.(2023)Placed, Strader, Carrillo, Atanasov, Indelman, Carlone, and Castellanos]{placed2023survey}
Julio~A Placed, Jared Strader, Henry Carrillo, Nikolay Atanasov, Vadim Indelman, Luca Carlone, and Jos{\'e}~A Castellanos.
\newblock A survey on active simultaneous localization and mapping: State of the art and new frontiers.
\newblock \emph{IEEE Transactions on Robotics}, 39\penalty0 (3):\penalty0 1686--1705, 2023.

\bibitem[Pomerleau et~al.(2011)Pomerleau, Magnenat, Colas, Liu, and Siegwart]{pomerleau2011tracking}
Fran{\c{c}}ois Pomerleau, St{\'e}phane Magnenat, Francis Colas, Ming Liu, and Roland Siegwart.
\newblock Tracking a depth camera: Parameter exploration for fast icp.
\newblock In \emph{IEEE/RSJ International Conference on Intelligent Robots and Systems}, pages 3824--3829, 2011.

\bibitem[Qu et~al.(2024)Qu, Yan, Wang, Yin, Chen, Xu, Zhang, Zhao, and Li]{anonymous2024implicit}
Delin Qu, Chi Yan, Dong Wang, Jie Yin, Qizhi Chen, Dan Xu, Yiting Zhang, Bin Zhao, and Xuelong Li.
\newblock Implicit event-{RGBD} neural {SLAM}.
\newblock In \emph{Proceedings of the IEEE/CVF Conference on Computer Vision and Pattern Recognition}, pages 19584--19594, 2024.

\bibitem[Recasens et~al.(2021)Recasens, Lamarca, F{\'a}cil, Montiel, and Civera]{recasens2021endo}
David Recasens, Jos{\'e} Lamarca, Jos{\'e}~M F{\'a}cil, JMM Montiel, and Javier Civera.
\newblock Endo-depth-and-motion: Reconstruction and tracking in endoscopic videos using depth networks and photometric constraints.
\newblock \emph{IEEE Robotics and Automation Letters}, 6\penalty0 (4):\penalty0 7225--7232, 2021.

\bibitem[Recasens et~al.(2023)Recasens, Oswald, Pollefeys, and Civera]{recasens2023drunkard}
David Recasens, Martin~R Oswald, Marc Pollefeys, and Javier Civera.
\newblock The drunkard's odometry: Estimating camera motion in deforming scenes.
\newblock In \emph{International Conference on Neural Information Processing Systems}, pages 48877--48889, 2023.

\bibitem[Rombach et~al.(2022)Rombach, Blattmann, Lorenz, Esser, and Ommer]{rombach2022high}
Robin Rombach, Andreas Blattmann, Dominik Lorenz, Patrick Esser, and Bj{\"o}rn Ommer.
\newblock High-resolution image synthesis with latent diffusion models.
\newblock In \emph{Proceedings of the IEEE/CVF Conference on Computer Vision and Pattern Recognition}, pages 10684--10695, 2022.

\bibitem[Sandstr{\"o}m et~al.(2023)Sandstr{\"o}m, Li, Van~Gool, and Oswald]{sandstrom2023point}
Erik Sandstr{\"o}m, Yue Li, Luc Van~Gool, and Martin~R Oswald.
\newblock Point-slam: Dense neural point cloud-based slam.
\newblock In \emph{Proceedings of the IEEE/CVF International Conference on Computer Vision}, pages 18433--18444, 2023.

\bibitem[Sarlin et~al.(2022)Sarlin, Dusmanu, Sch{\"o}nberger, Speciale, Gruber, Larsson, Miksik, and Pollefeys]{sarlin2022lamar}
Paul-Edouard Sarlin, Mihai Dusmanu, Johannes~L Sch{\"o}nberger, Pablo Speciale, Lukas Gruber, Viktor Larsson, Ondrej Miksik, and Marc Pollefeys.
\newblock Lamar: Benchmarking localization and mapping for augmented reality.
\newblock In \emph{European Conference on Computer Vision}, pages 686--704. Springer, 2022.

\bibitem[Sauder and Tuia(2024)]{sauder2024self}
Jonathan Sauder and Devis Tuia.
\newblock Self-supervised underwater caustics removal and descattering via deep monocular slam.
\newblock In \emph{European Conference on Computer Vision}. Springer, 2024.

\bibitem[Sch\"{o}nberger and Frahm(2016)]{schoenberger2016sfm}
Johannes~Lutz Sch\"{o}nberger and Jan-Michael Frahm.
\newblock Structure-from-motion revisited.
\newblock In \emph{Conference on Computer Vision and Pattern Recognition}, 2016.

\bibitem[Sch\"{o}nberger et~al.(2016)Sch\"{o}nberger, Zheng, Pollefeys, and Frahm]{schoenberger2016mvs}
Johannes~Lutz Sch\"{o}nberger, Enliang Zheng, Marc Pollefeys, and Jan-Michael Frahm.
\newblock Pixelwise view selection for unstructured multi-view stereo.
\newblock In \emph{European Conference on Computer Vision}, 2016.

\bibitem[Schops et~al.(2017)Schops, Schonberger, Galliani, Sattler, Schindler, Pollefeys, and Geiger]{schops2017multi}
Thomas Schops, Johannes~L Schonberger, Silvano Galliani, Torsten Sattler, Konrad Schindler, Marc Pollefeys, and Andreas Geiger.
\newblock A multi-view stereo benchmark with high-resolution images and multi-camera videos.
\newblock In \emph{Proceedings of the IEEE Conference on Computer Vision and Pattern Recognition}, pages 3260--3269, 2017.

\bibitem[Schops et~al.(2019)Schops, Sattler, and Pollefeys]{schops2019bad}
Thomas Schops, Torsten Sattler, and Marc Pollefeys.
\newblock {BAD SLAM: Bundle Adjusted Direct RGB-D SLAM}.
\newblock In \emph{IEEE/CVF Conference on Computer Vision and Pattern Recognition}, 2019.

\bibitem[Sener et~al.(2022)Sener, Chatterjee, Shelepov, He, Singhania, Wang, and Yao]{sener2022assembly101}
Fadime Sener, Dibyadip Chatterjee, Daniel Shelepov, Kun He, Dipika Singhania, Robert Wang, and Angela Yao.
\newblock Assembly101: A large-scale multi-view video dataset for understanding procedural activities.
\newblock In \emph{Proceedings of the IEEE/CVF Conference on Computer Vision and Pattern Recognition}, pages 21096--21106, 2022.

\bibitem[Shotton et~al.(2013)Shotton, Glocker, Zach, Izadi, Criminisi, and Fitzgibbon]{shotton2013scene}
Jamie Shotton, Ben Glocker, Christopher Zach, Shahram Izadi, Antonio Criminisi, and Andrew Fitzgibbon.
\newblock Scene coordinate regression forests for camera relocalization in rgb-d images.
\newblock In \emph{Proceedings of the IEEE Conference on Computer Vision and Pattern Recognition}, pages 2930--2937, 2013.

\bibitem[Singh et~al.(2024)Singh, Dharmadhikari, and Alexis]{singh2023online}
Mohit Singh, Mihir Dharmadhikari, and Kostas Alexis.
\newblock An online self-calibrating refractive camera model with application to underwater odometry, 2024.

\bibitem[Straub et~al.(2019)Straub, Whelan, Ma, Chen, Wijmans, Green, Engel, Mur-Artal, Ren, Verma, Clarkson, Yan, Budge, Yan, Pan, Yon, Zou, Leon, Carter, Briales, Gillingham, Mueggler, Pesqueira, Savva, Batra, Strasdat, Nardi, Goesele, Lovegrove, and Newcombe]{replica19arxiv}
Julian Straub, Thomas Whelan, Lingni Ma, Yufan Chen, Erik Wijmans, Simon Green, Jakob~J. Engel, Raul Mur-Artal, Carl Ren, Shobhit Verma, Anton Clarkson, Mingfei Yan, Brian Budge, Yajie Yan, Xiaqing Pan, June Yon, Yuyang Zou, Kimberly Leon, Nigel Carter, Jesus Briales, Tyler Gillingham, Elias Mueggler, Luis Pesqueira, Manolis Savva, Dhruv Batra, Hauke~M. Strasdat, Renzo~De Nardi, Michael Goesele, Steven Lovegrove, and Richard Newcombe.
\newblock The {R}eplica dataset: A digital replica of indoor spaces.
\newblock \emph{arXiv preprint arXiv:1906.05797}, 2019.

\bibitem[Sturm et~al.(2012)Sturm, Engelhard, Endres, Burgard, and Cremers]{sturm12iros}
J. Sturm, N. Engelhard, F. Endres, W. Burgard, and D. Cremers.
\newblock {A Benchmark for the Evaluation of RGB-D SLAM Systems}.
\newblock In \emph{IEEE/RSJ International Conference on Intelligent Robots and Systems}, 2012.

\bibitem[Sucar et~al.(2021)Sucar, Liu, Ortiz, and Davison]{sucar2021imap}
Edgar Sucar, Shikun Liu, Joseph Ortiz, and Andrew~J Davison.
\newblock imap: Implicit mapping and positioning in real-time.
\newblock In \emph{Proceedings of the IEEE/CVF International Conference on Computer Vision}, pages 6229--6238, 2021.

\bibitem[Sweeney(2016)]{theia-manual}
Chris Sweeney.
\newblock Theia multiview geometry library: Tutorial \& reference.
\newblock \url{http://theia-sfm.org}, 2016.

\bibitem[Teed and Deng(2021)]{teed2021droid}
Zachary Teed and Jia Deng.
\newblock Droid-slam: Deep visual slam for monocular, stereo, and rgb-d cameras.
\newblock \emph{Advances in Neural Information Processing Systems}, 34:\penalty0 16558--16569, 2021.

\bibitem[Teed et~al.(2024)Teed, Lipson, and Deng]{teed2024deep}
Zachary Teed, Lahav Lipson, and Jia Deng.
\newblock Deep patch visual odometry.
\newblock \emph{Advances in Neural Information Processing Systems}, 36, 2024.

\bibitem[Tosi et~al.(2024)Tosi, Zhang, Gong, Sandstr{\"o}m, Mattoccia, Oswald, and Poggi]{tosi2024nerfs}
Fabio Tosi, Youmin Zhang, Ziren Gong, Erik Sandstr{\"o}m, Stefano Mattoccia, Martin~R Oswald, and Matteo Poggi.
\newblock {How NeRFs and 3D Gaussian Splatting are Reshaping SLAM: a Survey}.
\newblock \emph{arXiv preprint arXiv:2402.13255}, 2024.

\bibitem[Touvron et~al.(2023)Touvron, Lavril, Izacard, Martinet, Lachaux, Lacroix, Rozi{\`e}re, Goyal, Hambro, Azhar, et~al.]{touvron2023llama}
Hugo Touvron, Thibaut Lavril, Gautier Izacard, Xavier Martinet, Marie-Anne Lachaux, Timoth{\'e}e Lacroix, Baptiste Rozi{\`e}re, Naman Goyal, Eric Hambro, Faisal Azhar, et~al.
\newblock Llama: Open and efficient foundation language models.
\newblock \emph{arXiv preprint arXiv:2302.13971}, 2023.

\bibitem[Umeyama(1991)]{umeyama1991least}
Shinji Umeyama.
\newblock Least-squares estimation of transformation parameters between two point patterns.
\newblock \emph{IEEE Transactions on Pattern Analysis \& Machine Intelligence}, 13\penalty0 (04):\penalty0 376--380, 1991.

\bibitem[Wang and Agapito(2024)]{wang20243d}
Hengyi Wang and Lourdes Agapito.
\newblock 3d reconstruction with spatial memory.
\newblock \emph{arXiv preprint arXiv:2408.16061}, 2024.

\bibitem[Wang et~al.(2024)Wang, Leroy, Cabon, Chidlovskii, and Revaud]{wang2024dust3r}
Shuzhe Wang, Vincent Leroy, Yohann Cabon, Boris Chidlovskii, and Jerome Revaud.
\newblock Dust3r: Geometric 3d vision made easy.
\newblock In \emph{Proceedings of the IEEE/CVF Conference on Computer Vision and Pattern Recognition}, pages 20697--20709, 2024.

\bibitem[Wang et~al.(2020)Wang, Zhu, Wang, Hu, Qiu, Wang, Hu, Kapoor, and Scherer]{wang2020tartanair}
Wenshan Wang, Delong Zhu, Xiangwei Wang, Yaoyu Hu, Yuheng Qiu, Chen Wang, Yafei Hu, Ashish Kapoor, and Sebastian Scherer.
\newblock Tartanair: A dataset to push the limits of visual slam.
\newblock In \emph{IEEE/RSJ International Conference on Intelligent Robots and Systems}, pages 4909--4916, 2020.

\bibitem[Wang et~al.(2021)Wang, Hu, and Scherer]{wang2021tartanvo}
Wenshan Wang, Yaoyu Hu, and Sebastian Scherer.
\newblock Tartanvo: A generalizable learning-based vo.
\newblock In \emph{Conference on Robot Learning}, pages 1761--1772. PMLR, 2021.

\bibitem[Wang et~al.(2004)Wang, Bovik, Sheikh, and Simoncelli]{wang2004image}
Zhou Wang, Alan~C Bovik, Hamid~R Sheikh, and Eero~P Simoncelli.
\newblock Image quality assessment: from error visibility to structural similarity.
\newblock \emph{IEEE Transactions on Image Processing}, 13\penalty0 (4):\penalty0 600--612, 2004.

\bibitem[Wenzel et~al.(2020)Wenzel, Wang, Yang, Cheng, Khan, von Stumberg, Zeller, and Cremers]{wenzel2020fourseasons}
P. Wenzel, R. Wang, N. Yang, Q. Cheng, Q. Khan, L. von Stumberg, N. Zeller, and D. Cremers.
\newblock {4Seasons}: A cross-season dataset for multi-weather {SLAM} in autonomous driving.
\newblock In \emph{Proceedings of the German Conference on Pattern Recognition}, 2020.

\bibitem[Wenzel et~al.(2024)Wenzel, Yang, Wang, Zeller, and Cremers]{wenzel20244seasons}
Patrick Wenzel, Nan Yang, Rui Wang, Niclas Zeller, and Daniel Cremers.
\newblock 4seasons: Benchmarking visual slam and long-term localization for autonomous driving in challenging conditions.
\newblock \emph{International Journal of Computer Vision}, pages 1--23, 2024.

\bibitem[Wulf et~al.(2007)Wulf, Nuchter, Hertzberg, and Wagner]{wulf2007ground}
Oliver Wulf, Andreas Nuchter, Joachim Hertzberg, and Bernardo Wagner.
\newblock Ground truth evaluation of large urban 6d slam.
\newblock In \emph{IEEE/RSJ International Conference on Intelligent Robots and Systems}, pages 650--657, 2007.

\bibitem[Yang et~al.(2018)Yang, Wang, Gao, and Cremers]{yang2018challenges}
Nan Yang, Rui Wang, Xiang Gao, and Daniel Cremers.
\newblock Challenges in monocular visual odometry: Photometric calibration, motion bias, and rolling shutter effect.
\newblock \emph{IEEE Robotics and Automation Letters}, 3\penalty0 (4):\penalty0 2878--2885, 2018.

\bibitem[Yang et~al.(2020)Yang, Stumberg, Wang, and Cremers]{yang2020d3vo}
Nan Yang, Lukas~von Stumberg, Rui Wang, and Daniel Cremers.
\newblock D3vo: Deep depth, deep pose and deep uncertainty for monocular visual odometry.
\newblock In \emph{Proceedings of the IEEE/CVF Conference on Computer Vision and Pattern Recognition}, pages 1281--1292, 2020.

\bibitem[Ye et~al.(2023)Ye, Lan, Chen, Ming, Yu, Bao, Cui, and Zhang]{ye2023pvo}
Weicai Ye, Xinyue Lan, Shuo Chen, Yuhang Ming, Xingyuan Yu, Hujun Bao, Zhaopeng Cui, and Guofeng Zhang.
\newblock Pvo: Panoptic visual odometry.
\newblock In \emph{Proceedings of the IEEE/CVF Conference on Computer Vision and Pattern Recognition}, pages 9579--9589, 2023.

\bibitem[Yeshwanth et~al.(2023)Yeshwanth, Liu, Nie{\ss}ner, and Dai]{yeshwanth2023scannet++}
Chandan Yeshwanth, Yueh-Cheng Liu, Matthias Nie{\ss}ner, and Angela Dai.
\newblock Scannet++: A high-fidelity dataset of 3d indoor scenes.
\newblock In \emph{Proceedings of the IEEE/CVF International Conference on Computer Vision}, pages 12--22, 2023.

\bibitem[Zhang et~al.(2024)Zhang, Herrmann, Hur, Jampani, Darrell, Cole, Sun, and Yang]{zhang2024monst3r}
Junyi Zhang, Charles Herrmann, Junhwa Hur, Varun Jampani, Trevor Darrell, Forrester Cole, Deqing Sun, and Ming-Hsuan Yang.
\newblock Monst3r: A simple approach for estimating geometry in the presence of motion.
\newblock \emph{arXiv preprint arXiv:2410.03825}, 2024.

\bibitem[Zhang et~al.(2022)Zhang, Helmberger, Fu, Wisth, Camurri, Scaramuzza, and Fallon]{zhang2022hilti}
Lintong Zhang, Michael Helmberger, Lanke Frank~Tarimo Fu, David Wisth, Marco Camurri, Davide Scaramuzza, and Maurice Fallon.
\newblock Hilti-oxford dataset: A millimeter-accurate benchmark for simultaneous localization and mapping.
\newblock \emph{IEEE Robotics and Automation Letters}, 8\penalty0 (1):\penalty0 408--415, 2022.

\bibitem[Zhang et~al.(2018)Zhang, Isola, Efros, Shechtman, and Wang]{zhang2018unreasonable}
Richard Zhang, Phillip Isola, Alexei~A Efros, Eli Shechtman, and Oliver Wang.
\newblock The unreasonable effectiveness of deep features as a perceptual metric.
\newblock In \emph{Proceedings of the IEEE Conference on Computer Vision and Pattern Recognition}, pages 586--595, 2018.

\bibitem[Zhang and Scaramuzza(2018)]{zhang2018tutorial}
Zichao Zhang and Davide Scaramuzza.
\newblock A tutorial on quantitative trajectory evaluation for visual (-inertial) odometry.
\newblock In \emph{IEEE/RSJ International Conference on Intelligent Robots and Systems}, pages 7244--7251, 2018.

\bibitem[Zhang and Scaramuzza(2019)]{zhang2019rethinking}
Zichao Zhang and Davide Scaramuzza.
\newblock Rethinking trajectory evaluation for slam: A probabilistic, continuous-time approach.
\newblock \emph{arXiv preprint arXiv:1906.03996}, 2019.

\bibitem[Zhi et~al.(2021)Zhi, Sucar, Mouton, Haughton, Laidlow, and Davison]{Zhi:etal:arxiv2021}
Shuaifeng Zhi, Edgar Sucar, Andre Mouton, Iain Haughton, Tristan Laidlow, and Andrew~J. Davison.
\newblock {iLabel}: Interactive neural scene labelling.
\newblock \emph{arXiv}, 2021.

\bibitem[Zhu et~al.(2022)Zhu, Peng, Larsson, Xu, Bao, Cui, Oswald, and Pollefeys]{zhu2022nice}
Zihan Zhu, Songyou Peng, Viktor Larsson, Weiwei Xu, Hujun Bao, Zhaopeng Cui, Martin~R Oswald, and Marc Pollefeys.
\newblock Nice-slam: Neural implicit scalable encoding for slam.
\newblock In \emph{Proceedings of the IEEE/CVF Conference on Computer Vision and Pattern Recognition}, pages 12786--12796, 2022.

\bibitem[Zhu et~al.(2024)Zhu, Peng, Larsson, Cui, Oswald, Geiger, and Pollefeys]{zhu2024nicer}
Zihan Zhu, Songyou Peng, Viktor Larsson, Zhaopeng Cui, Martin~R Oswald, Andreas Geiger, and Marc Pollefeys.
\newblock Nicer-slam: Neural implicit scene encoding for rgb slam.
\newblock In \emph{International Conference on 3D Vision}, pages 42--52, 2024.

\end{thebibliography}
}

\clearpage
\setcounter{page}{1}
\maketitlesupplementary

\section{Hyperparameter Selection}
\subsection{Hyperparameter Tuning SfM}
\begin{figure*}[t]
    \centering
    \highlightedsubfigure{0.220\textwidth}{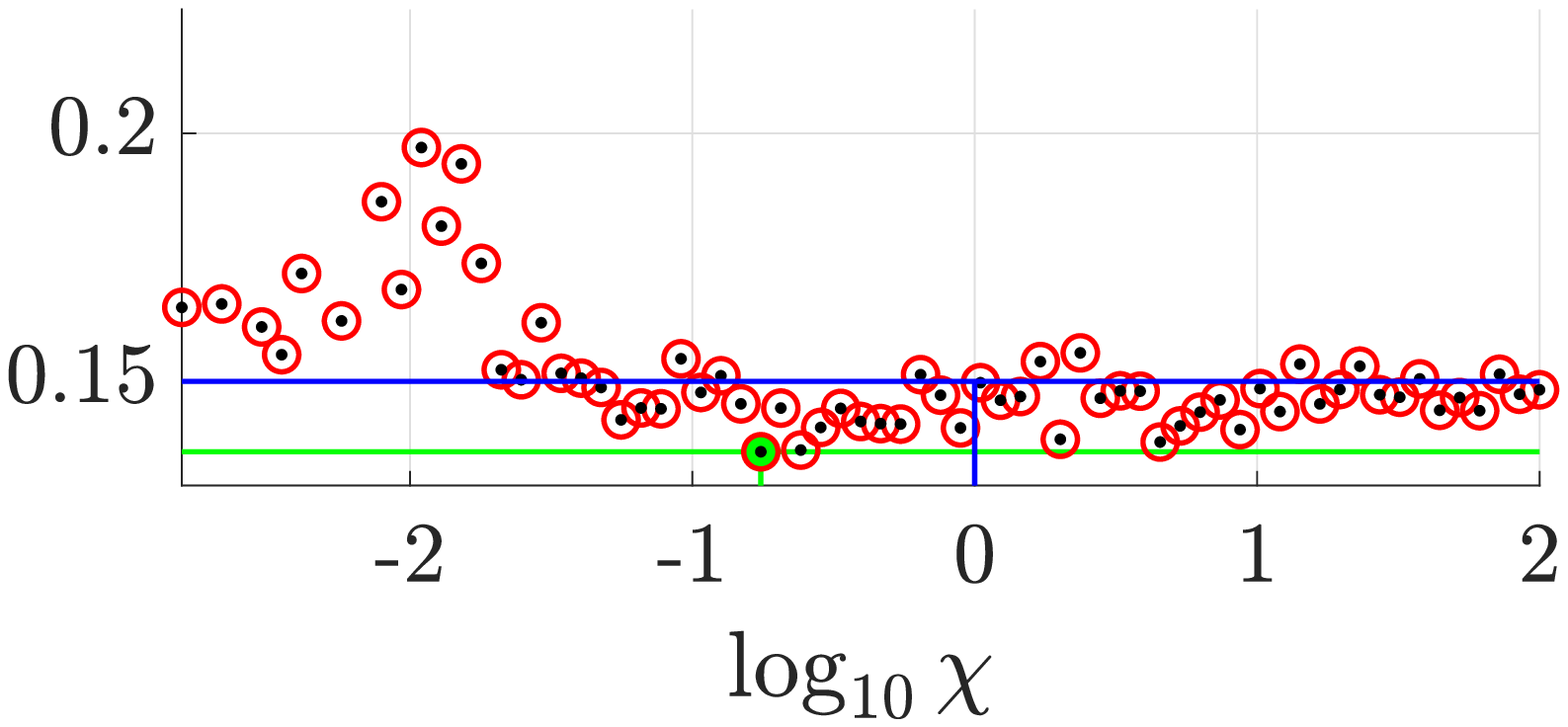}{Th. max angle error}{green}
    \highlightedsubfigure{0.220\textwidth}{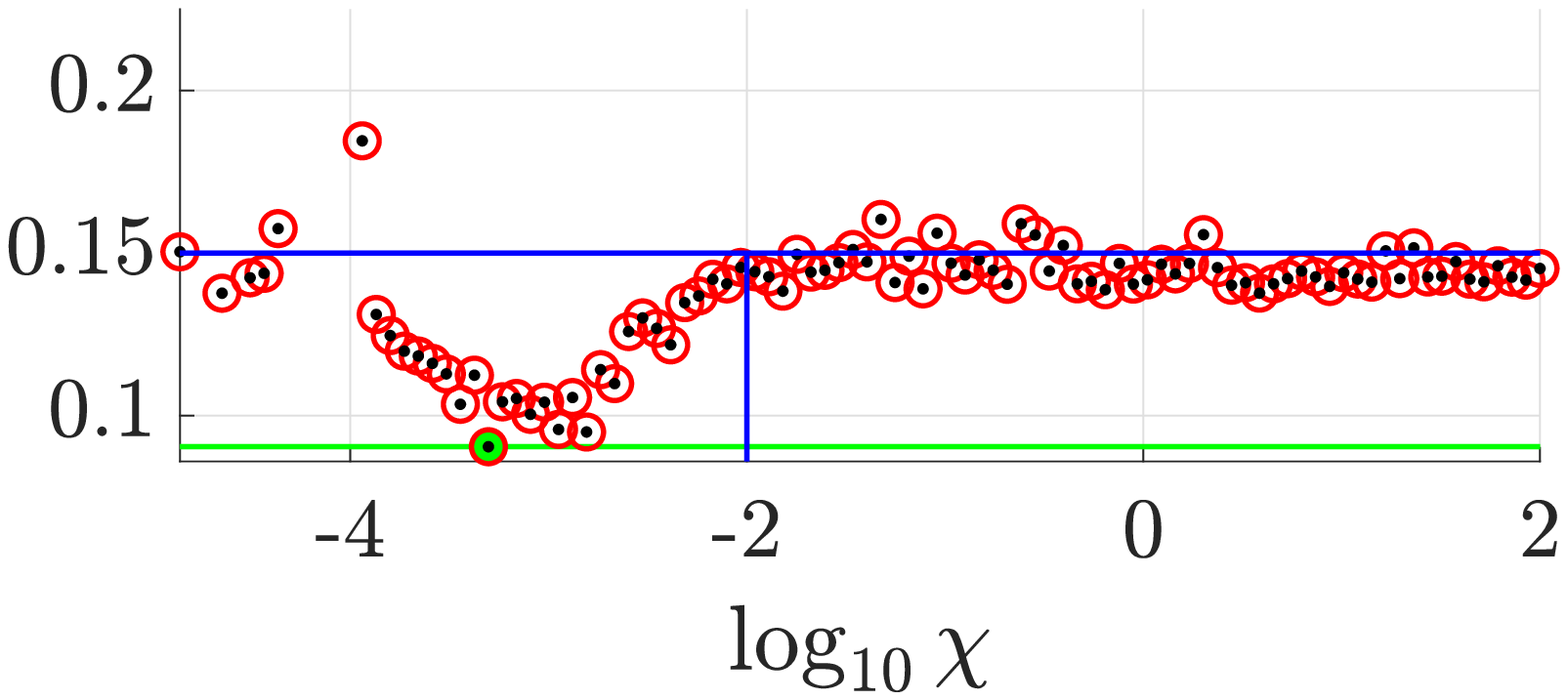}{Th. max reprojection error}{green}
    \highlightedsubfigure{0.220\textwidth}{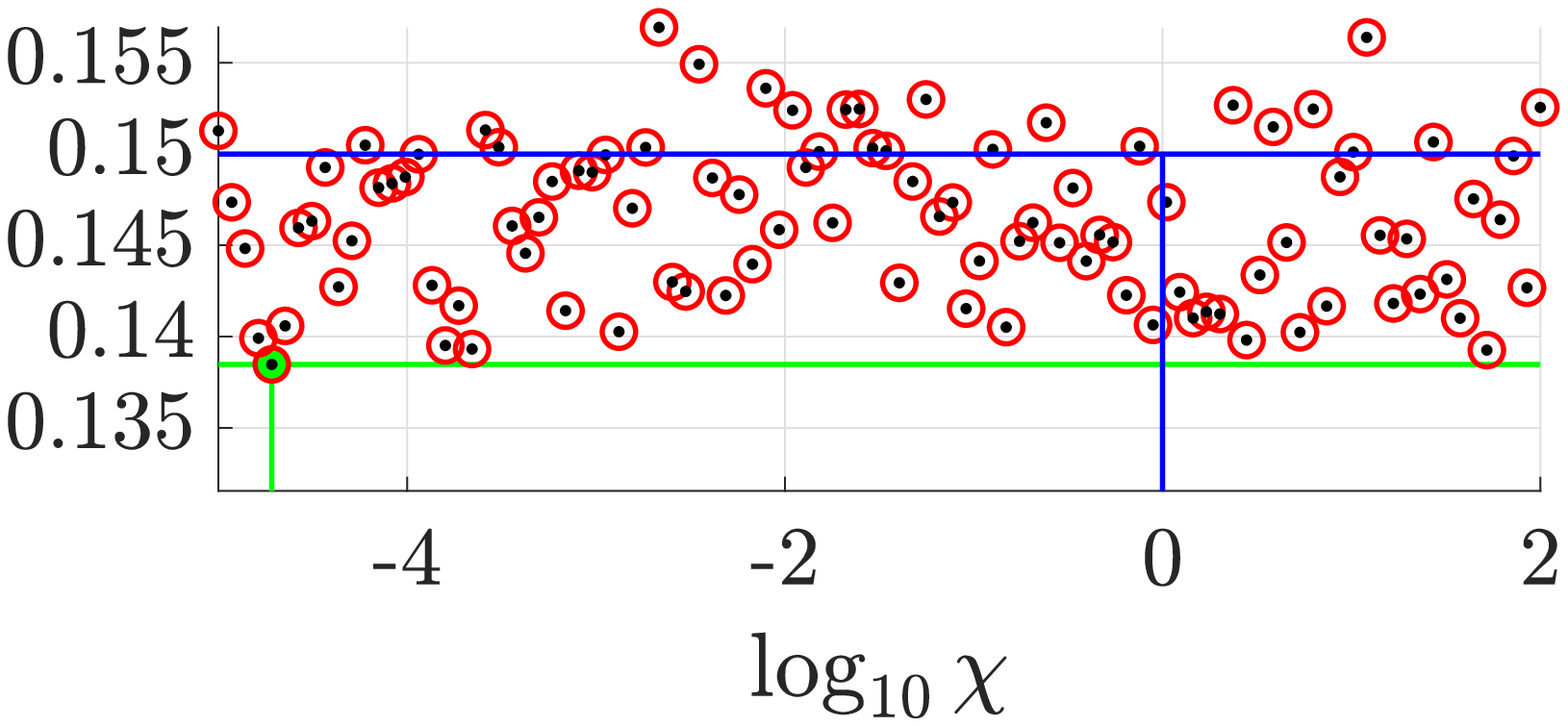}{Th. min triangulation angle}{white}
    \highlightedsubfigure{0.220\textwidth}{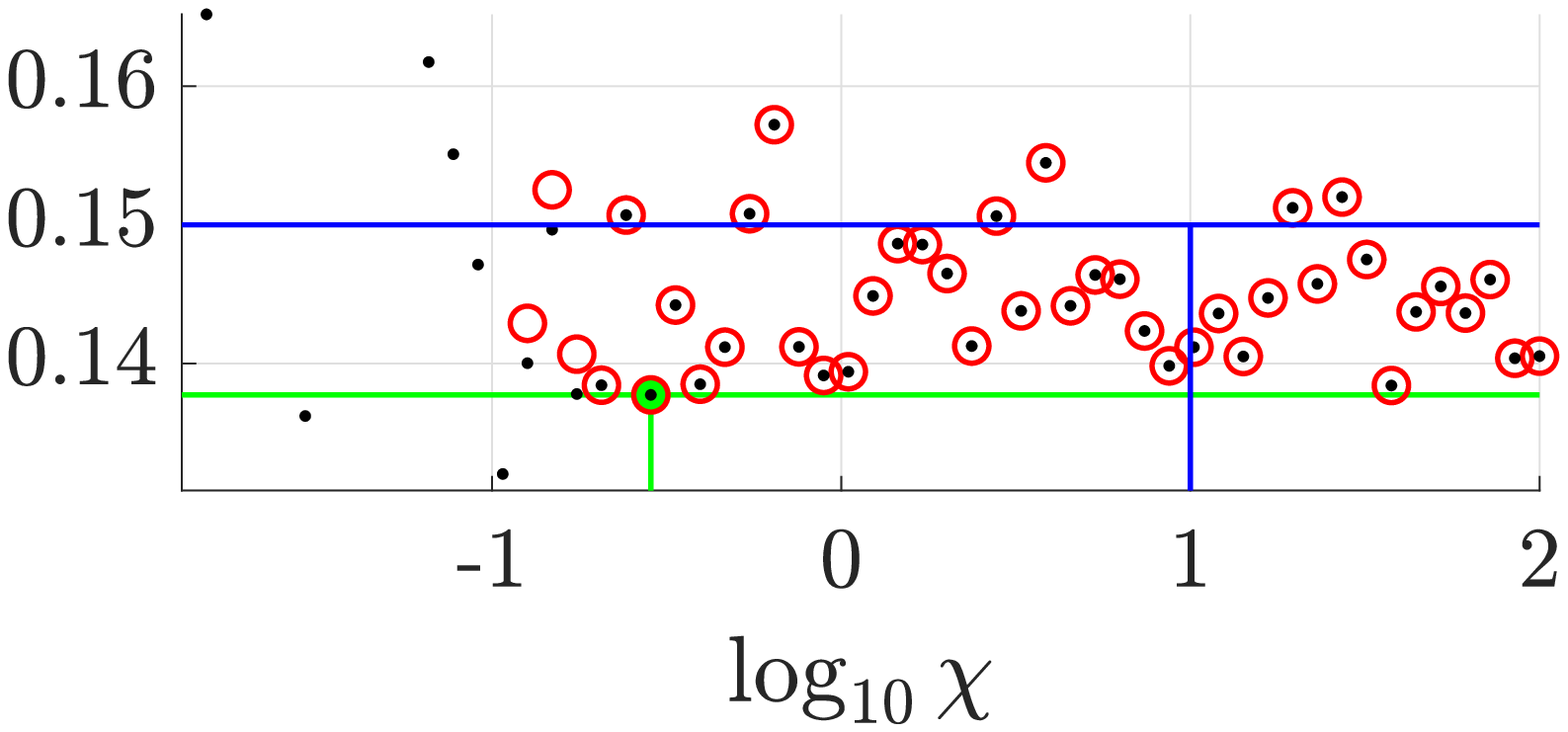}{Th. max rotation error}{white}
    \highlightedsubfigure{0.220\textwidth}{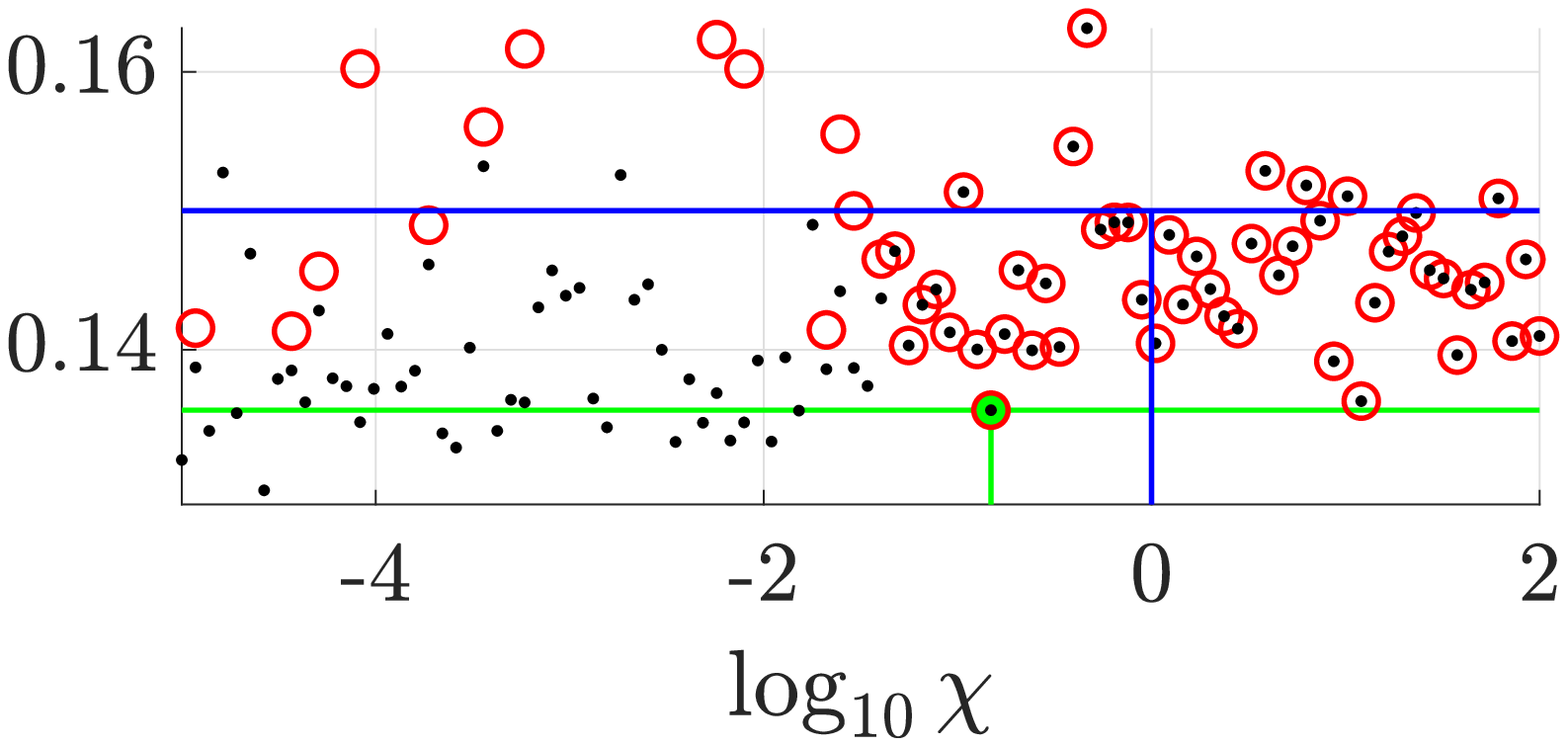}{Th. max epipolar error E}{white}
    \highlightedsubfigure{0.220\textwidth}{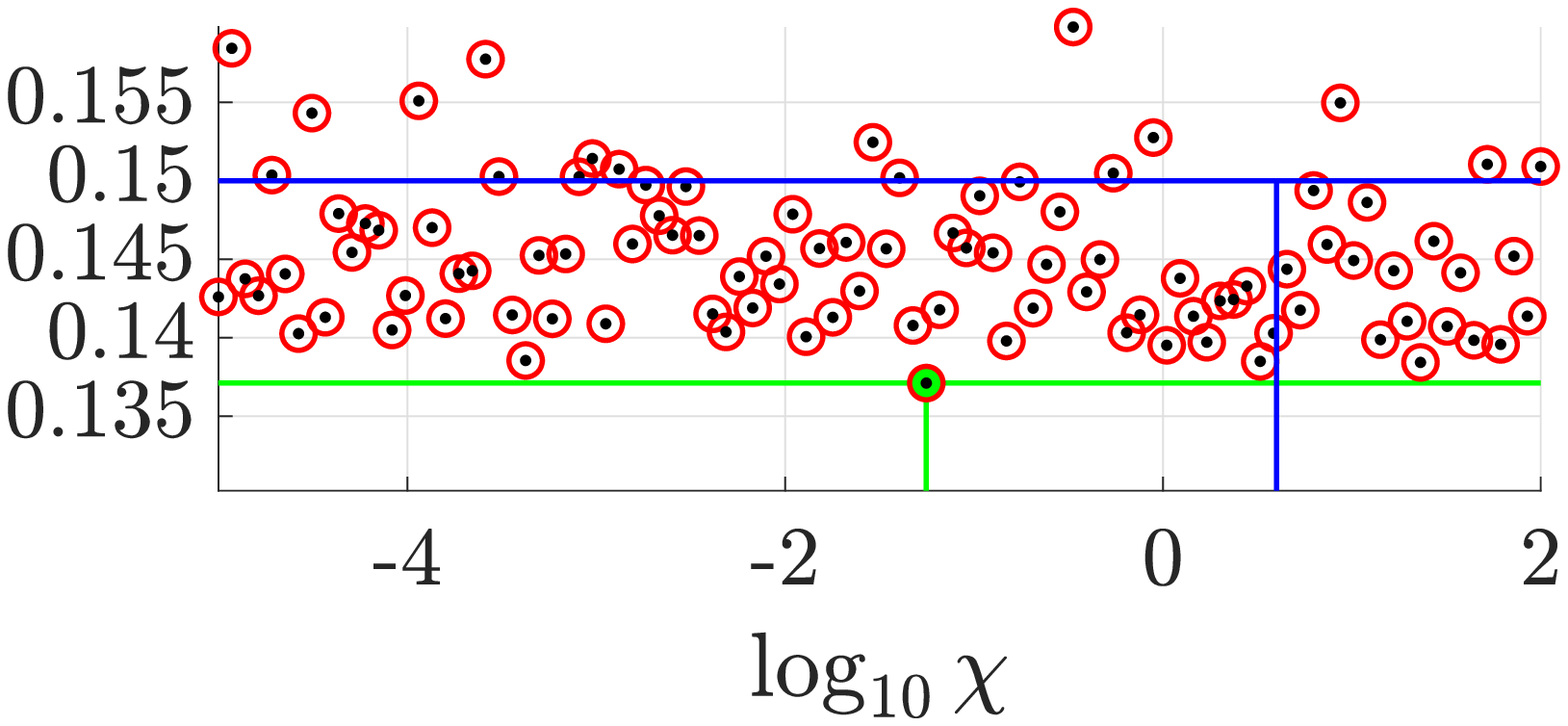}{Th. max epipolar error F}{white}
    \highlightedsubfigure{0.220\textwidth}{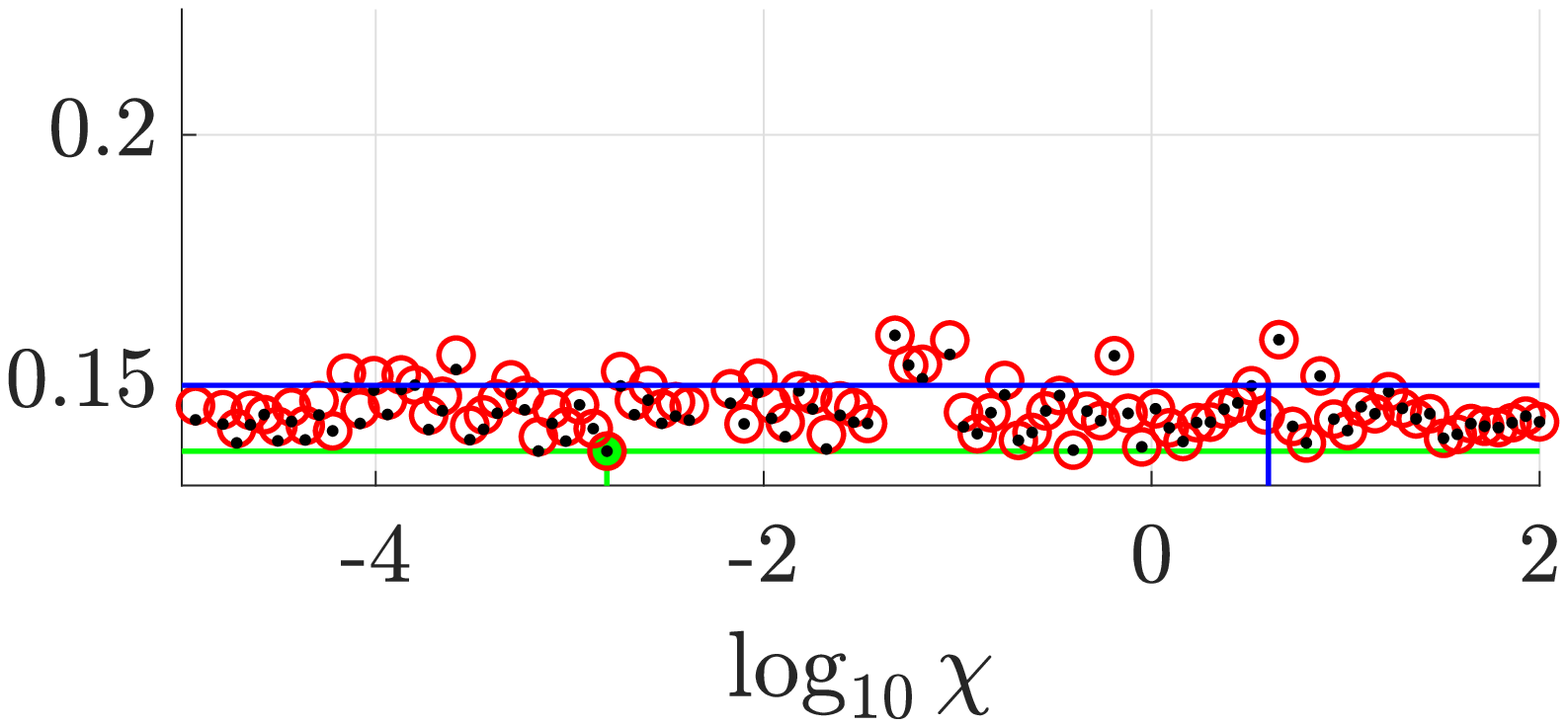}{Th. max epipolar error H}{white}
    \highlightedsubfigure{0.220\textwidth}{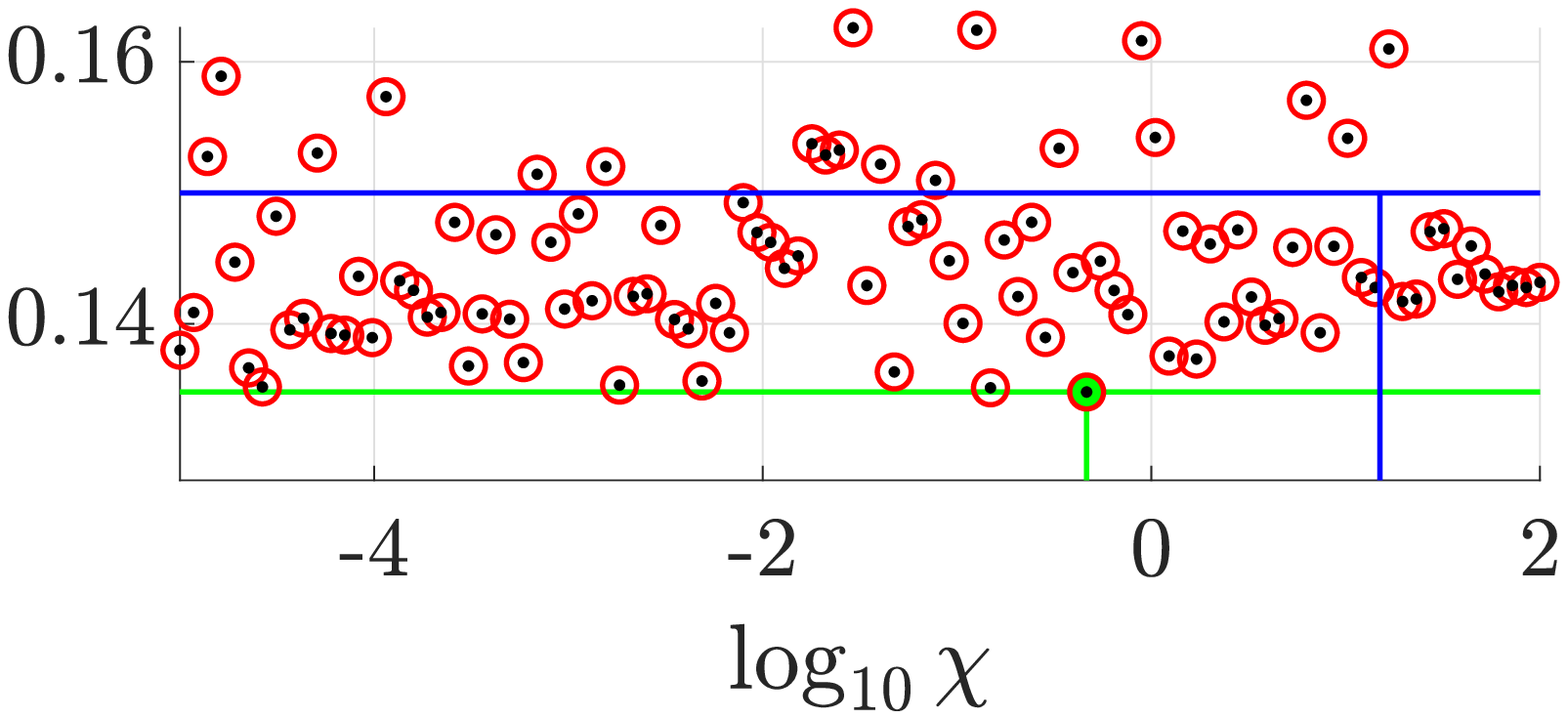}{Tr. complete max reproj error}{white}
    \highlightedsubfigure{0.220\textwidth}{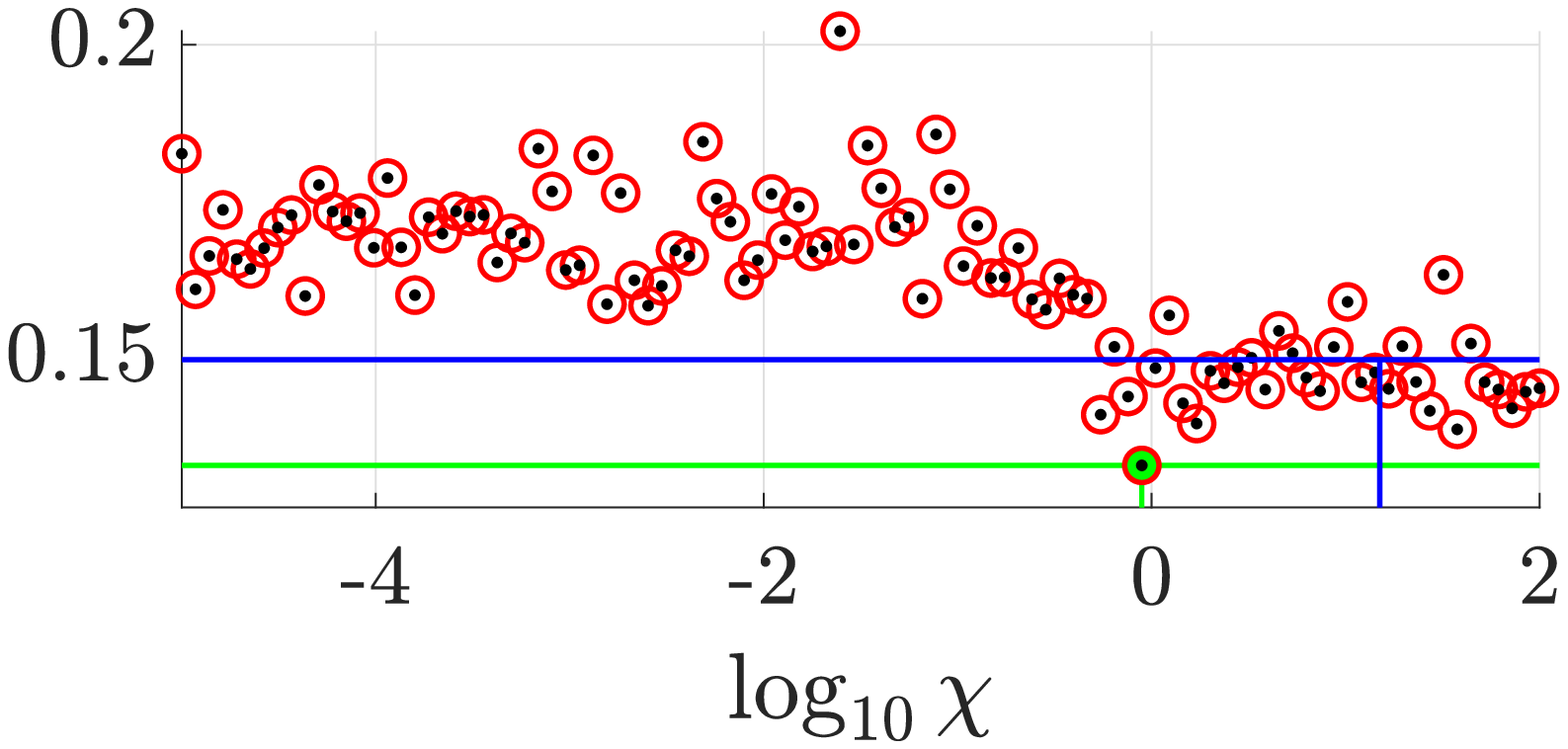}{Tr. merge max reproj error}{green}
    \highlightedsubfigure{0.220\textwidth}{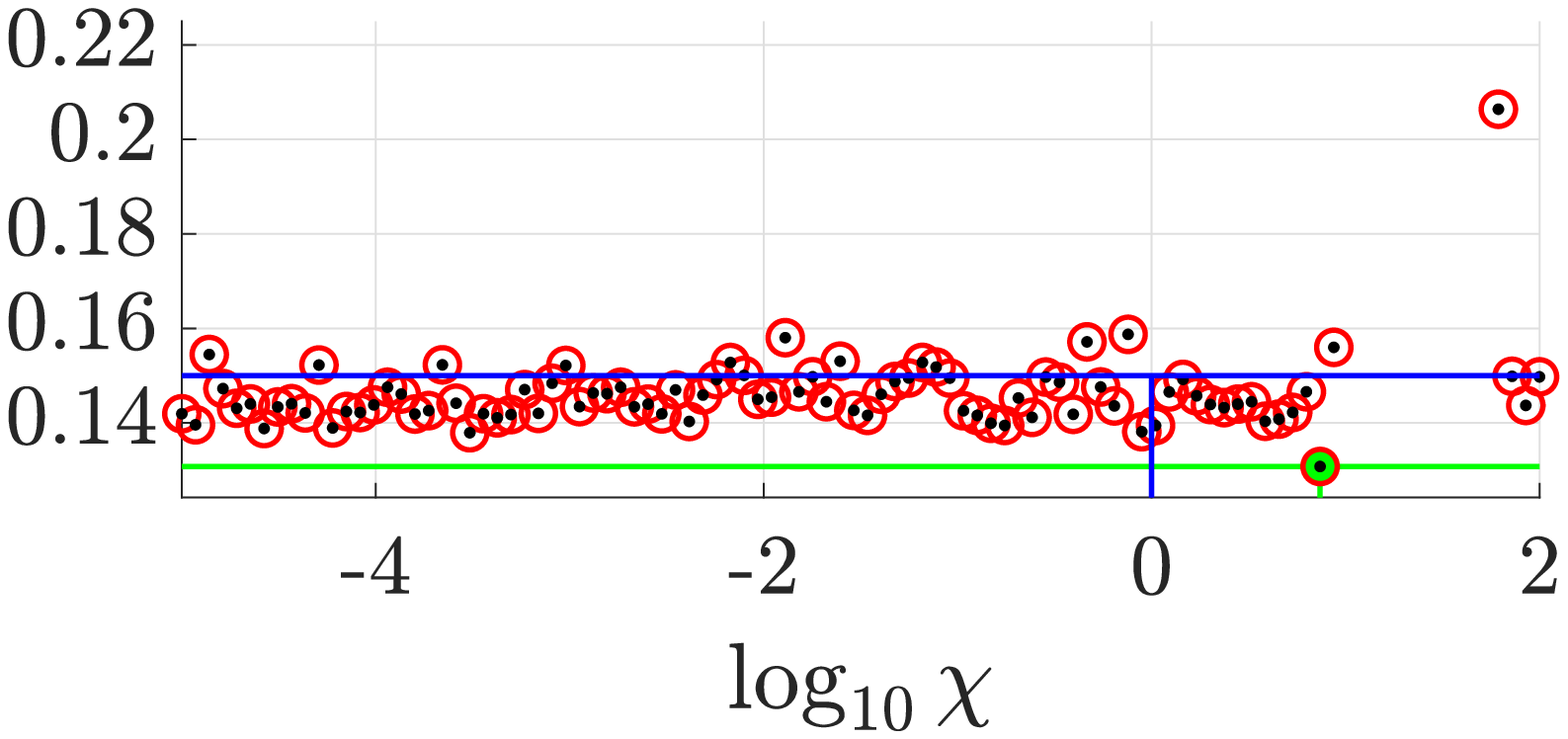}{Tr. min angle}{white}
    \highlightedsubfigure{0.220\textwidth}{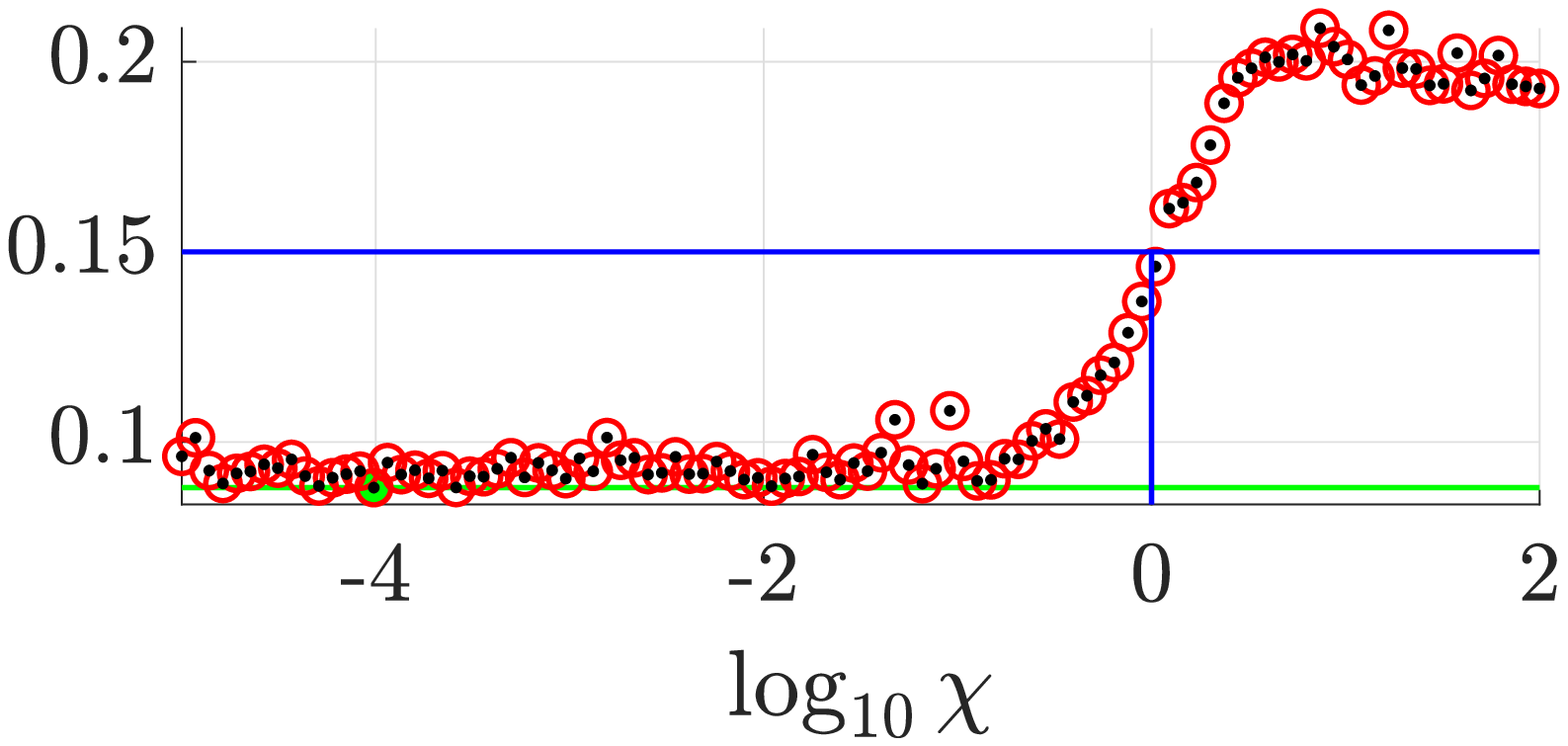}{BA th loss function}{green}
    \highlightedsubfigure{0.220\textwidth}{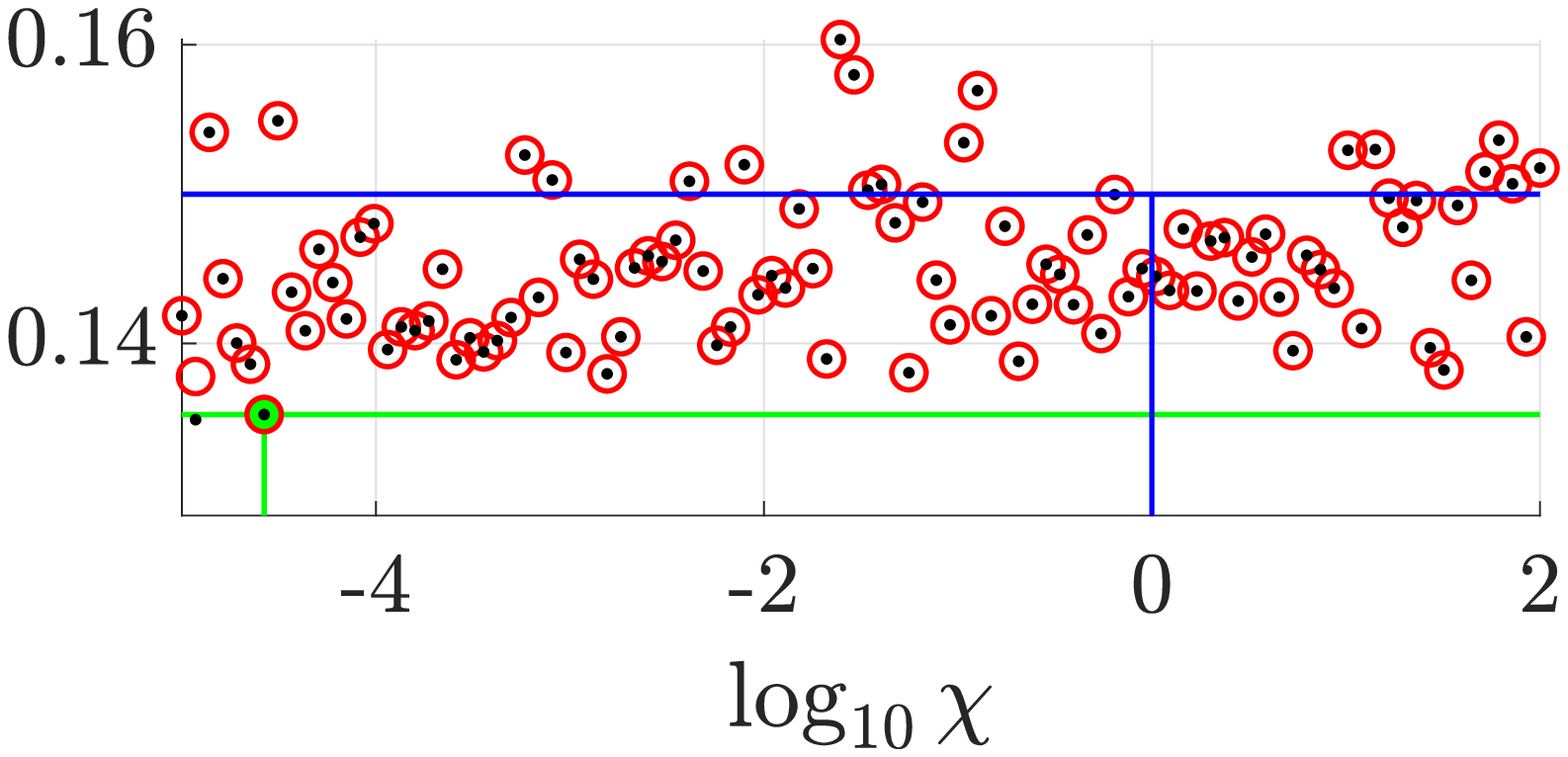}{RPE max epipolar error}{white}
    \highlightedsubfigure{0.220\textwidth}{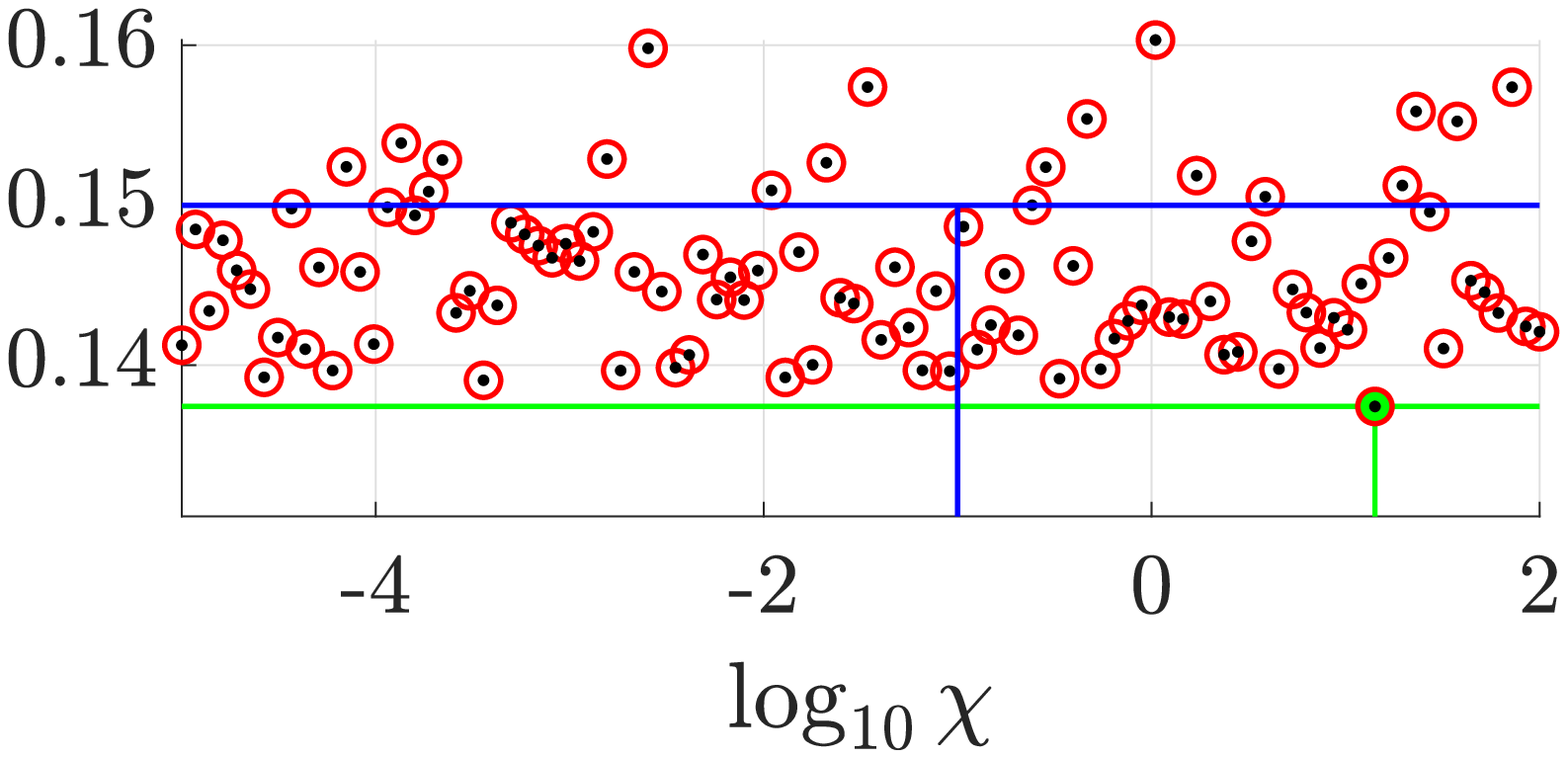}{GP th loss function}{white}
    \highlightedsubfigure{0.220\textwidth}{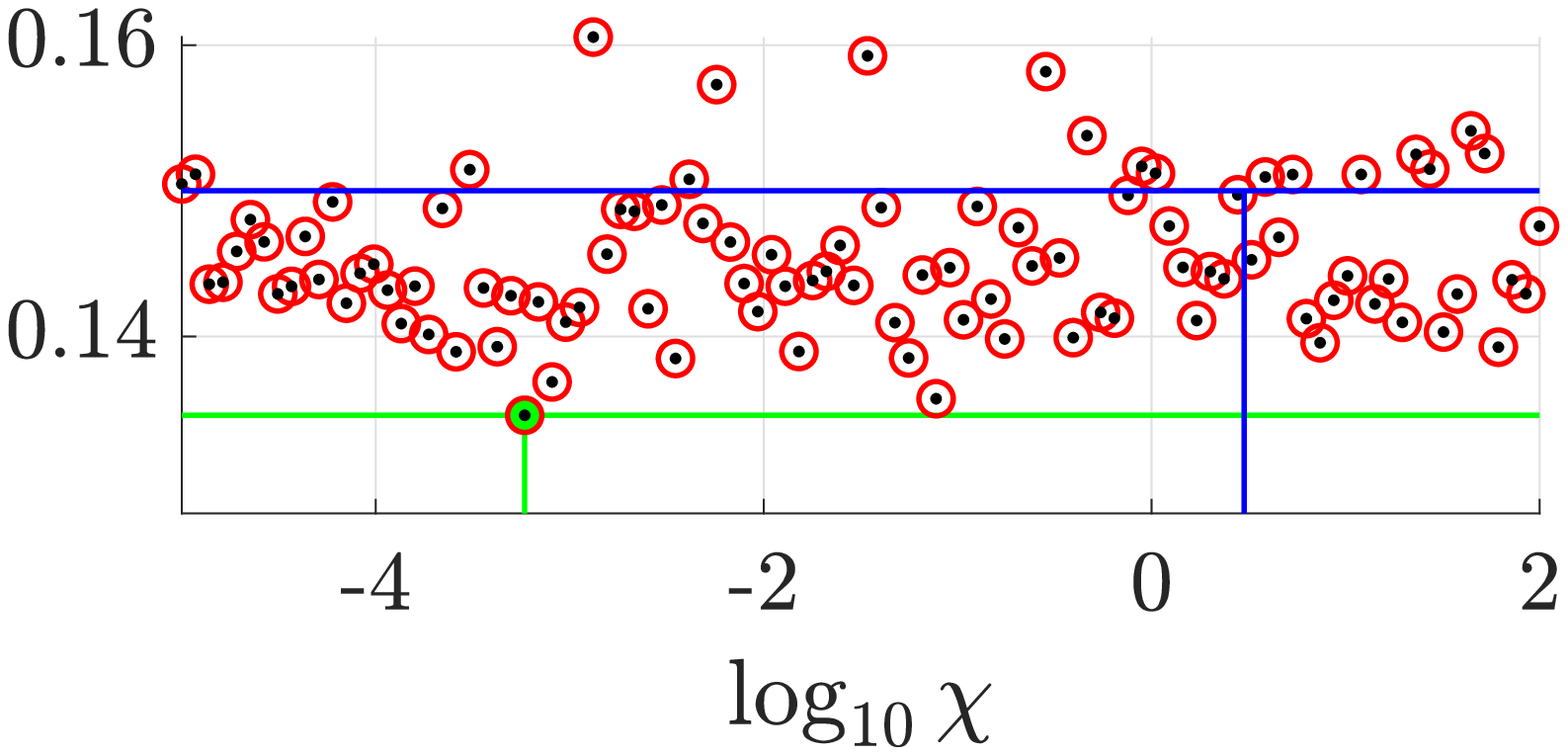}{Sift Ext. dsp max scale}{white}
    \highlightedsubfigure{0.220\textwidth}{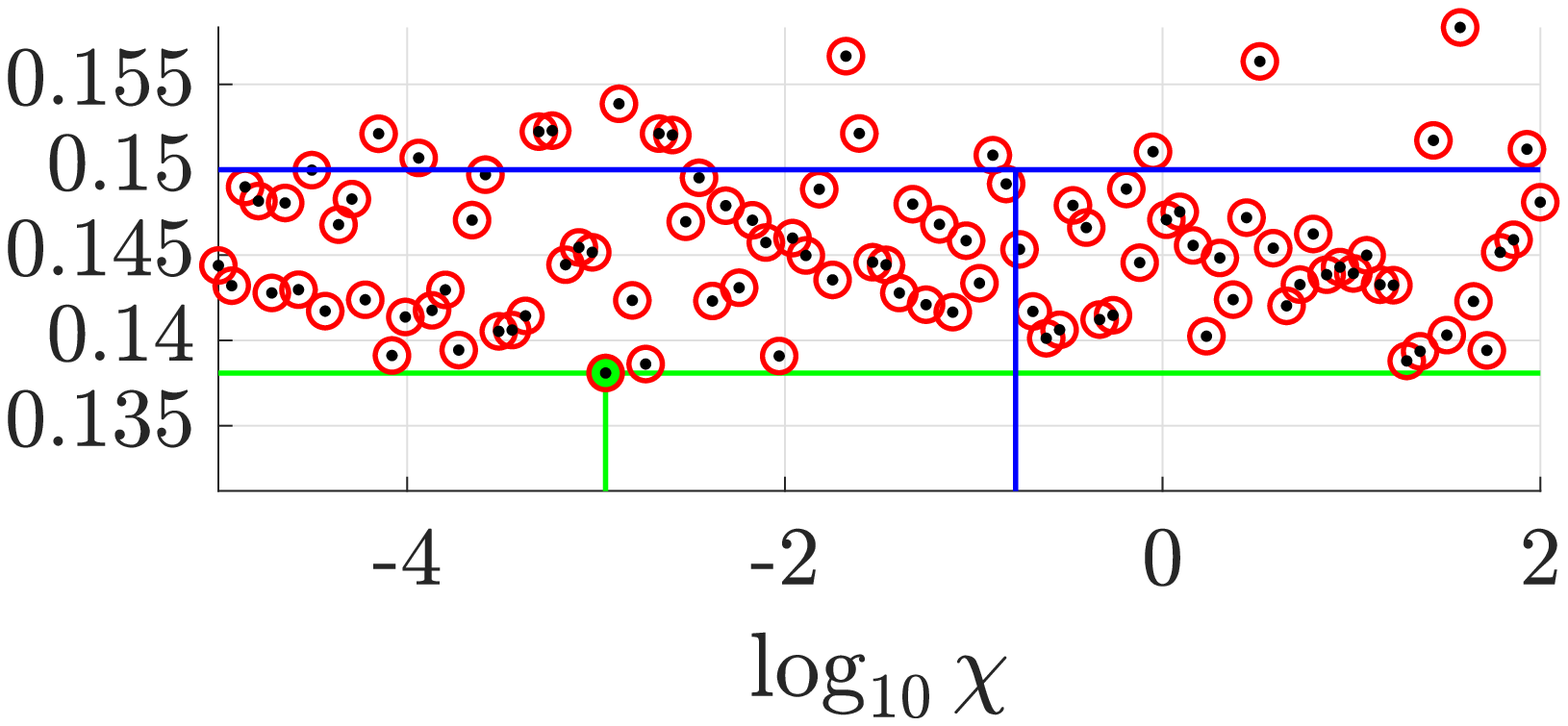}{Sift Ext. dsp min scale}{white}
    \highlightedsubfigure{0.220\textwidth}{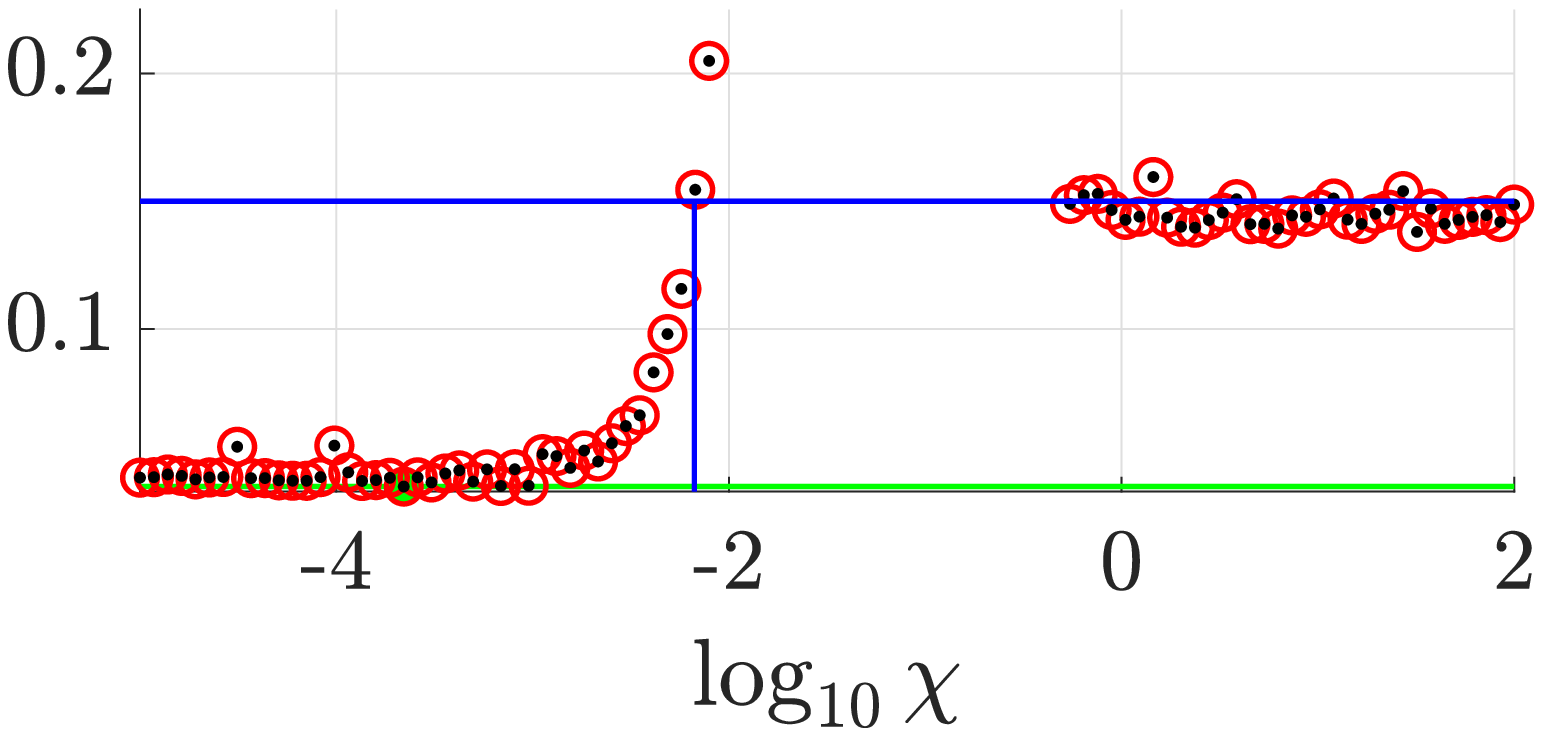}{Sift Ext. peak threshold}{green}
    \highlightedsubfigure{0.220\textwidth}{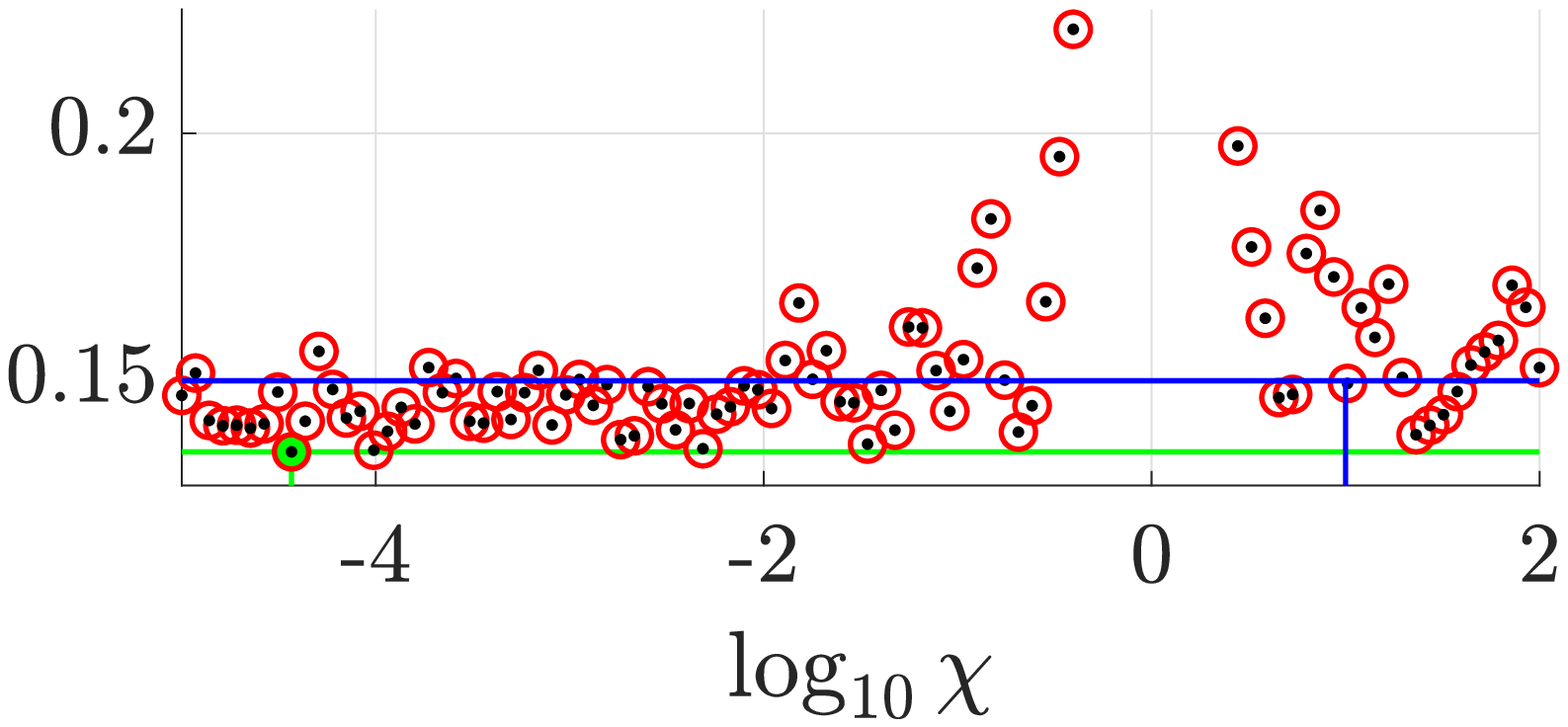}{Sift Ext. edge threshold}{white}
    \highlightedsubfigure{0.220\textwidth}{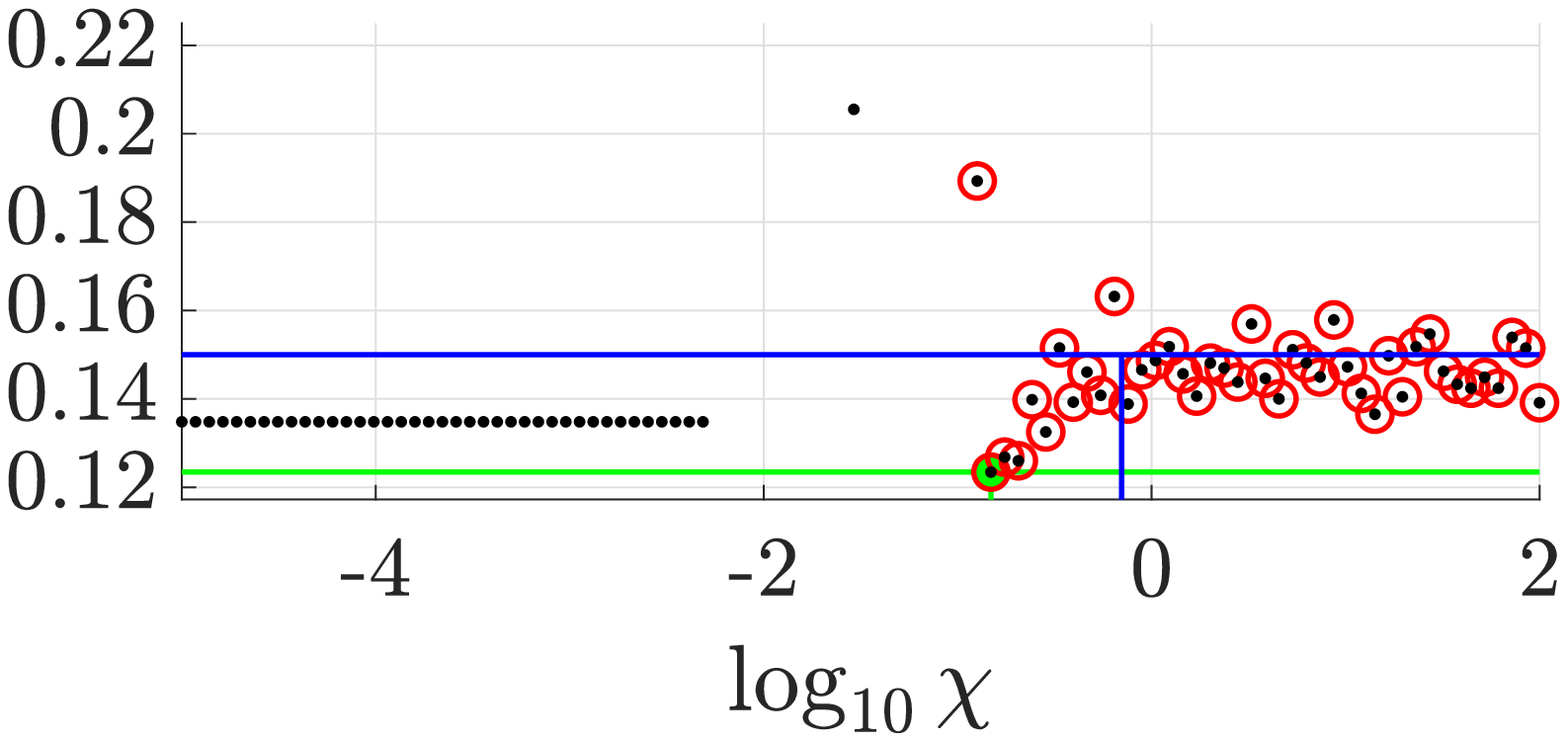}{Sift Matching max distance}{white}
    \highlightedsubfigure{0.220\textwidth}{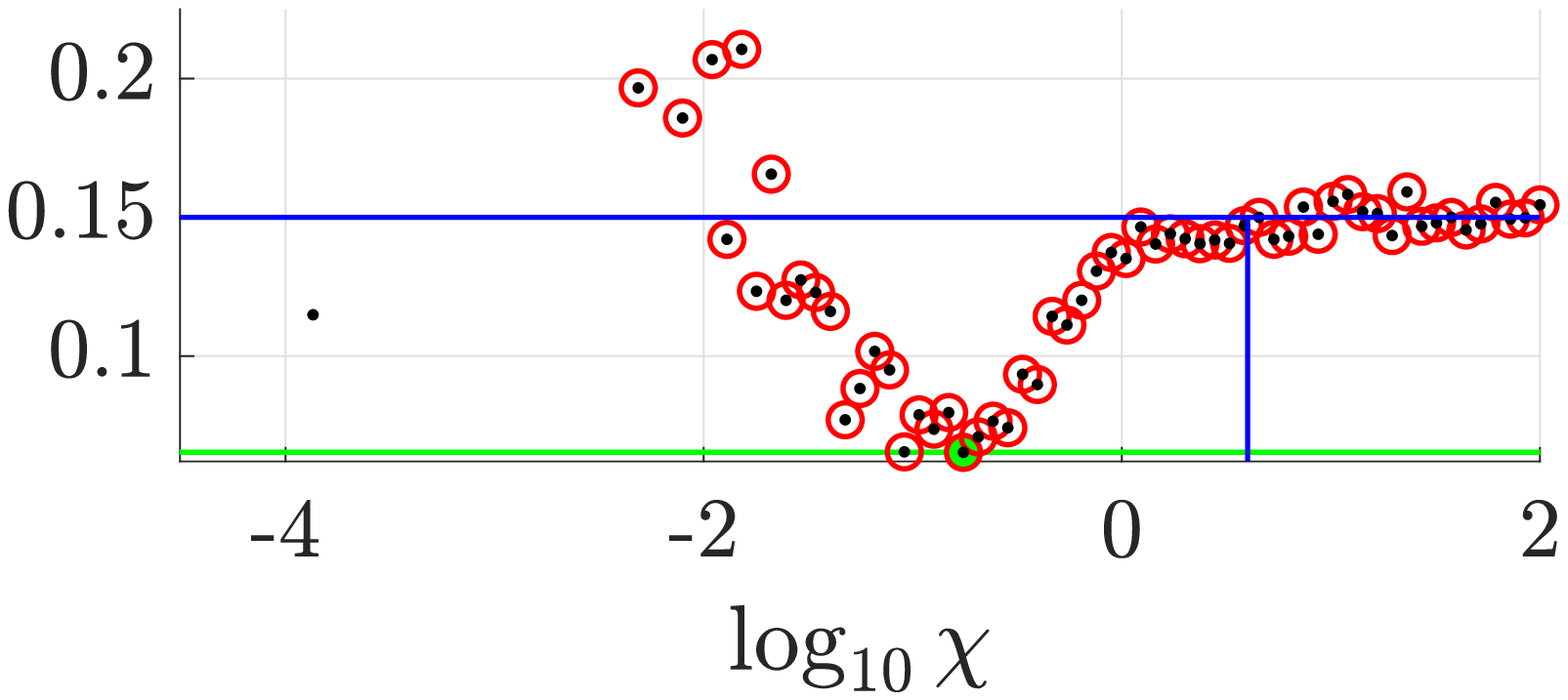}{2 View Geom. max error}{green}
    \highlightedsubfigure{0.220\textwidth}{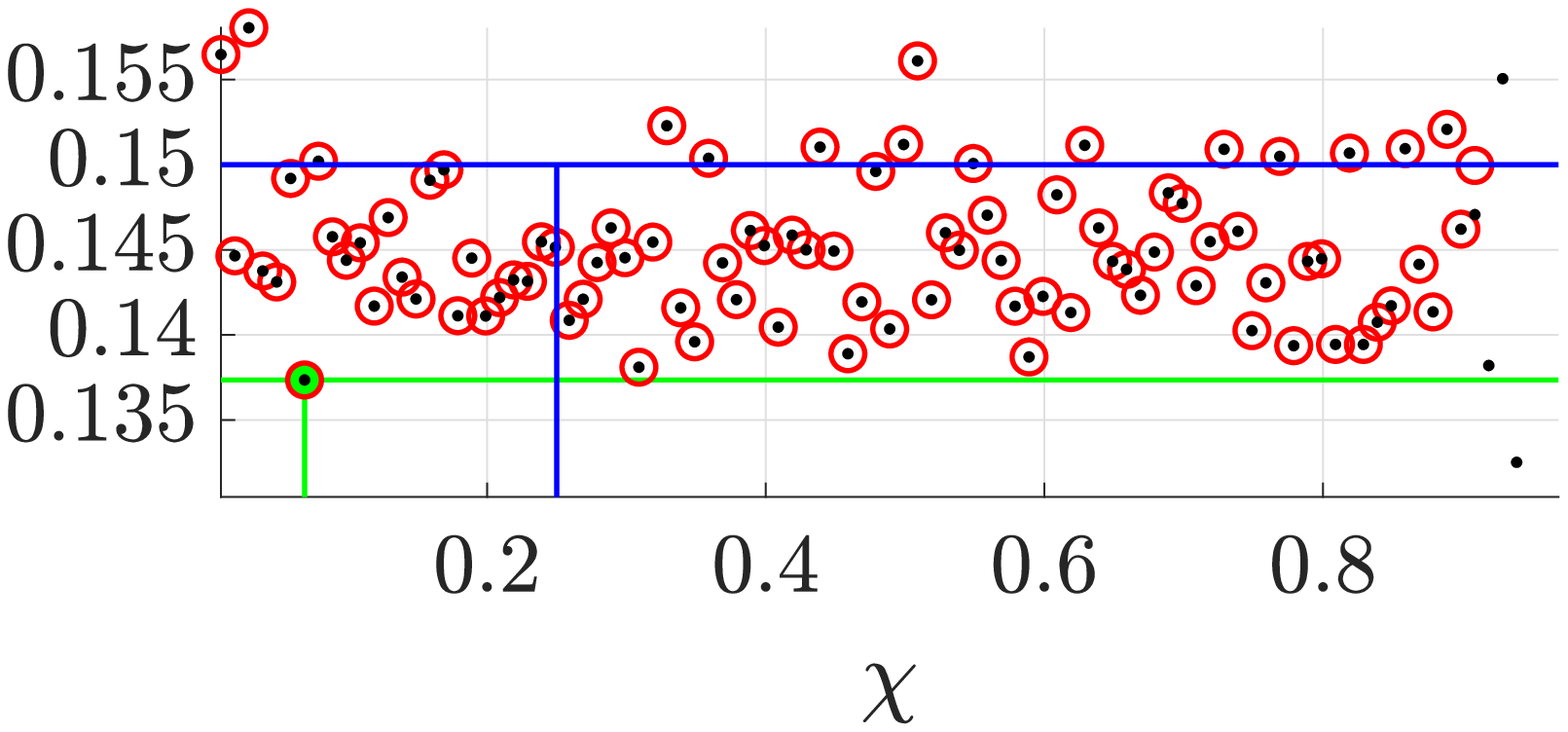}{Th. min inlier ratio}{white}
    \highlightedsubfigure{0.220\textwidth}{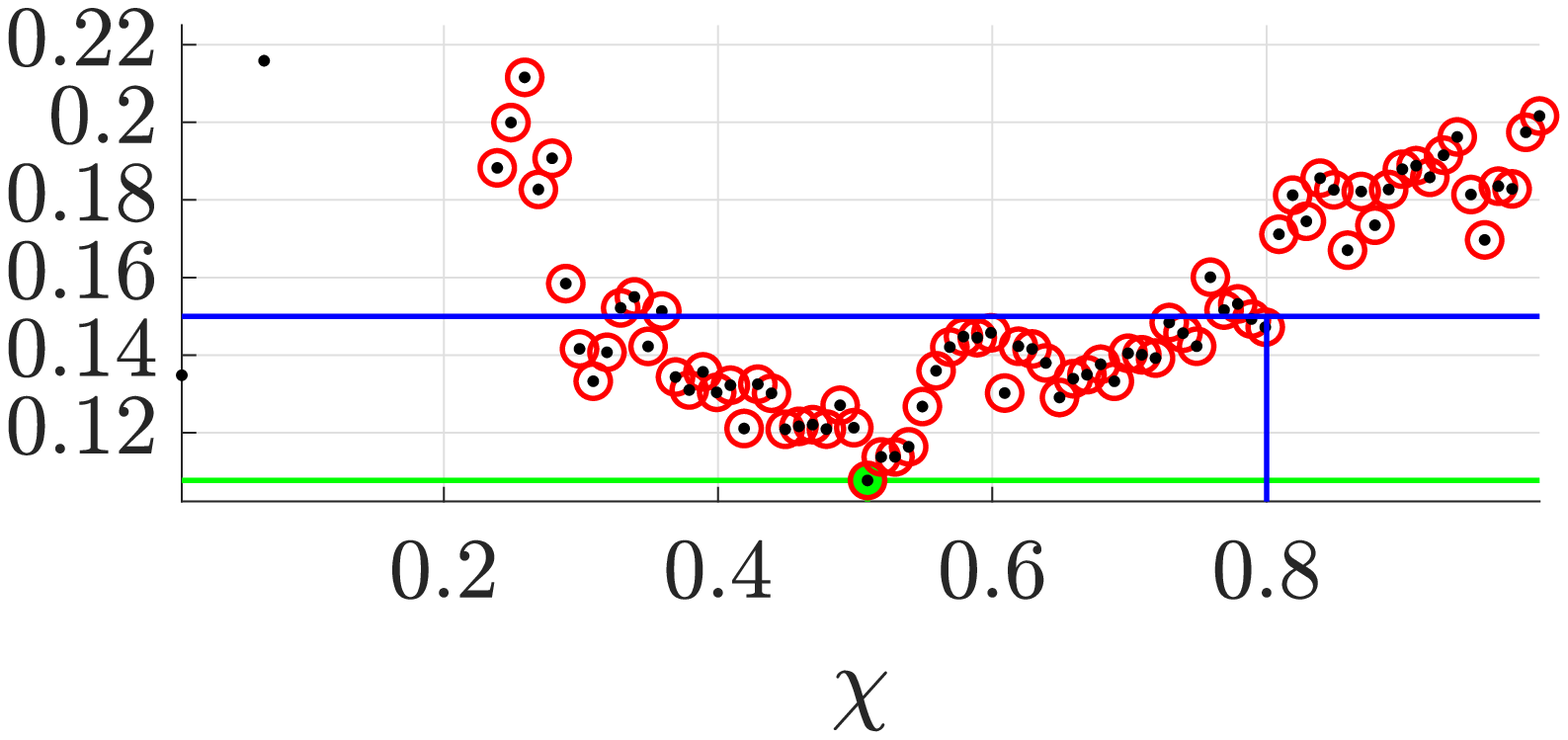}{Sift Matching max ratio}{green}    
    \highlightedsubfigure{0.220\textwidth}{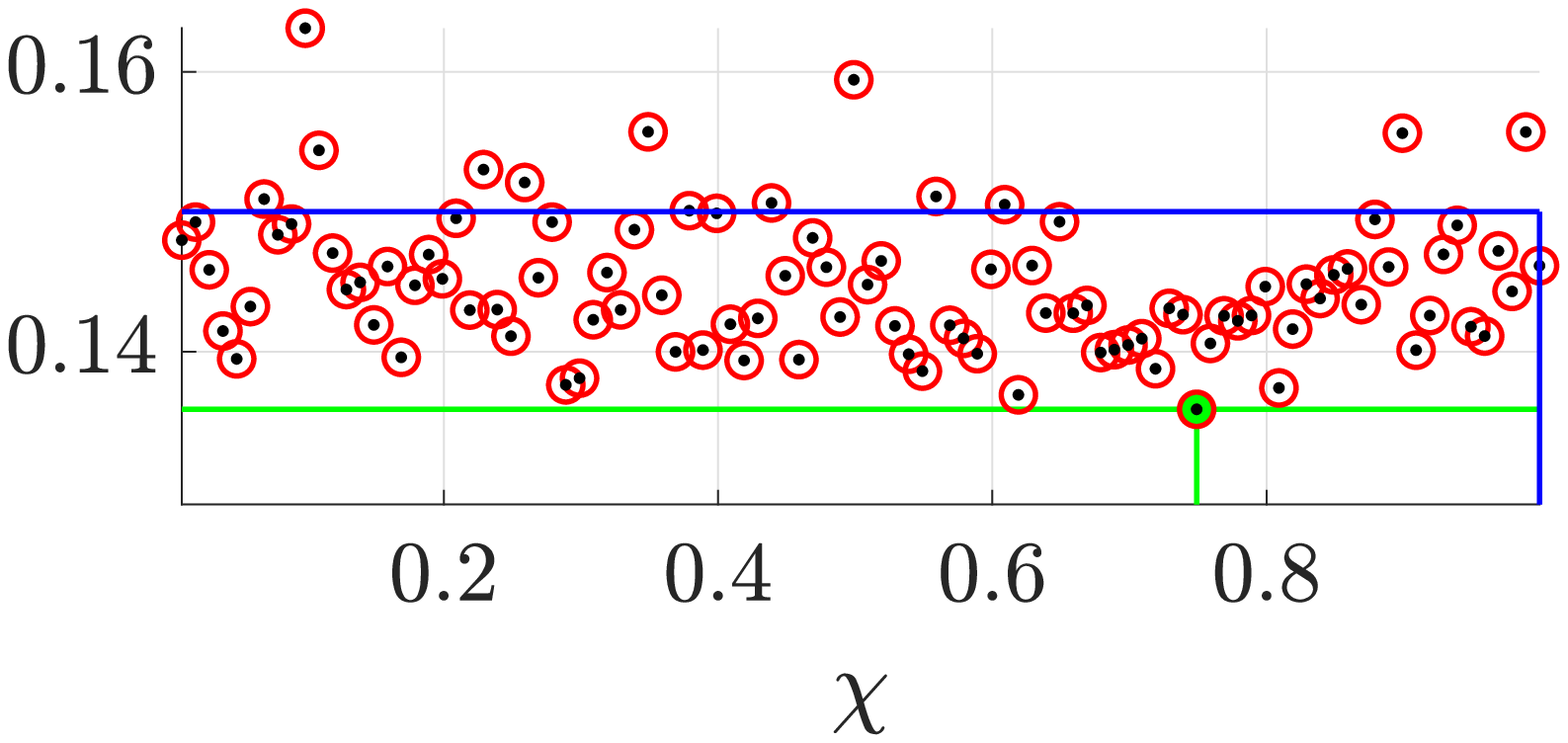}{2 View Geo. confidence}{white}   
    \highlightedsubfigure{0.220\textwidth}{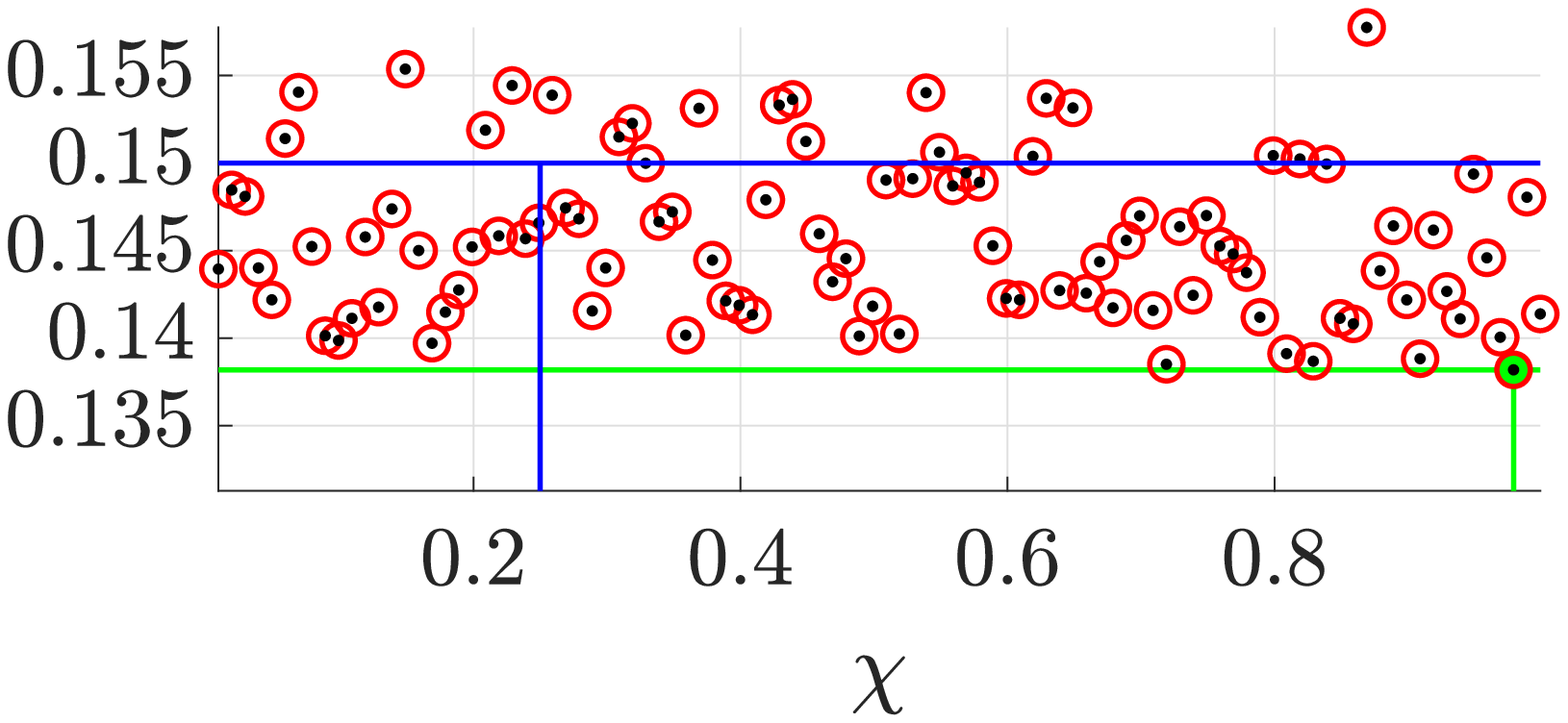}{2 View Geom. min inlier ratio}{white} 
    \highlightedsubfigure{0.220\textwidth}{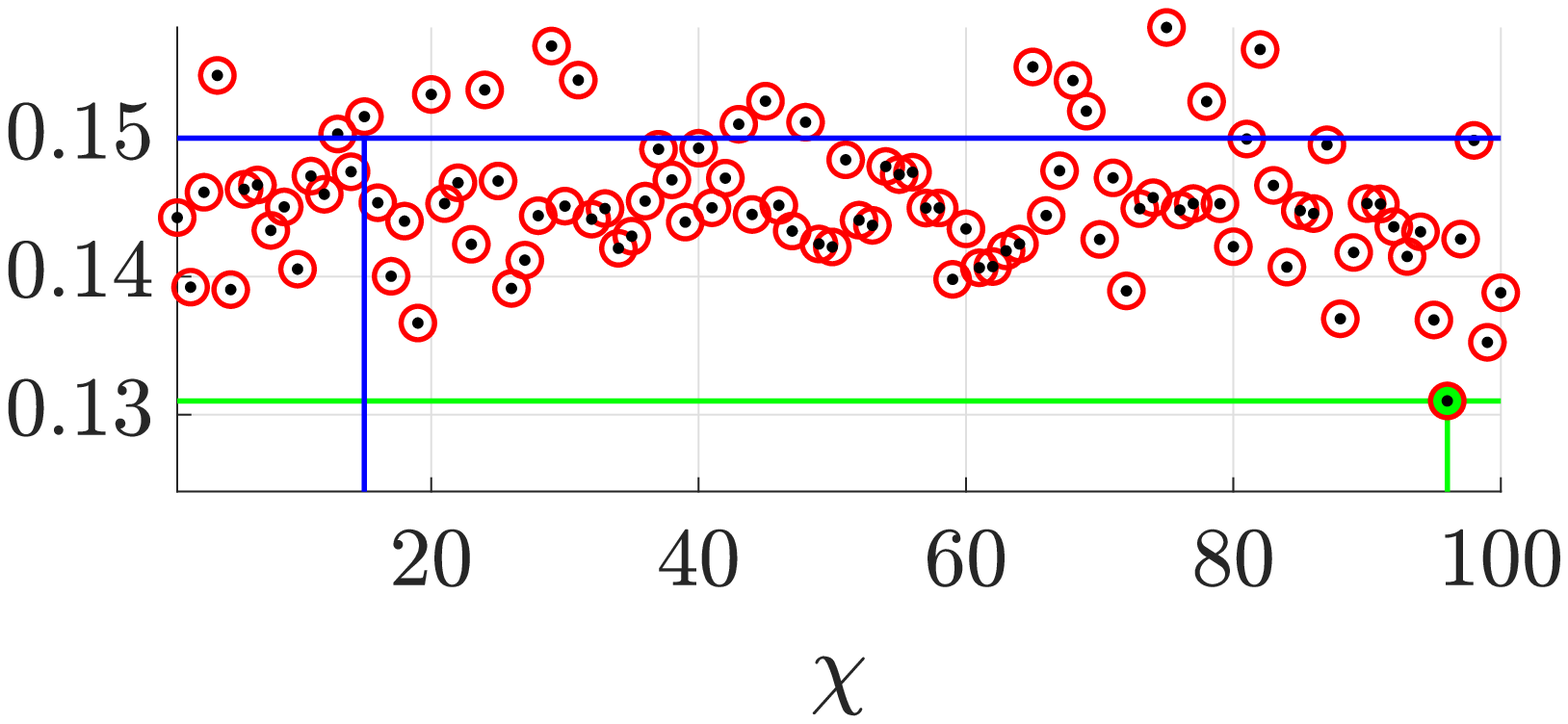}{Triang. min num matches}{white} 
    \highlightedsubfigure{0.220\textwidth}{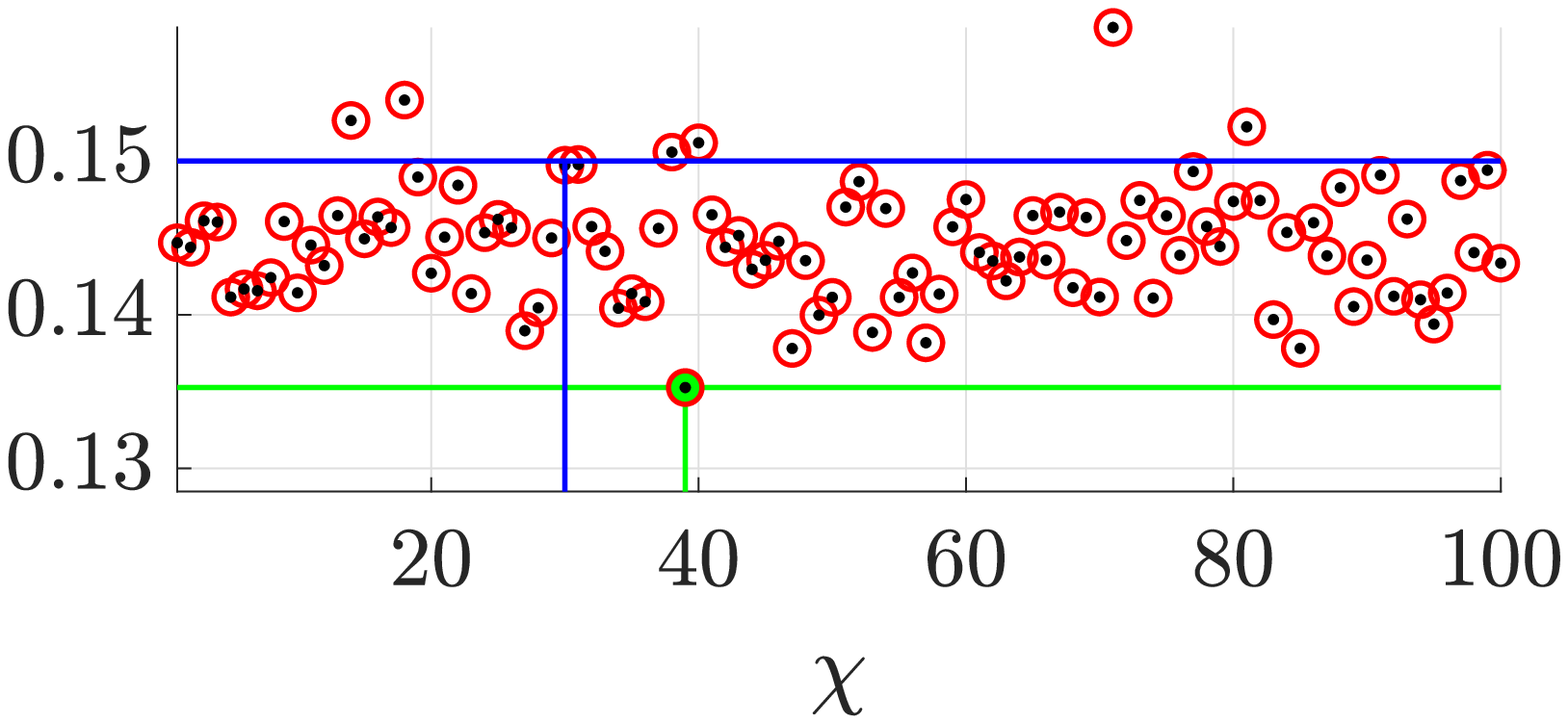}{Th. min inlier num}{white}
    \highlightedsubfigure{0.220\textwidth}{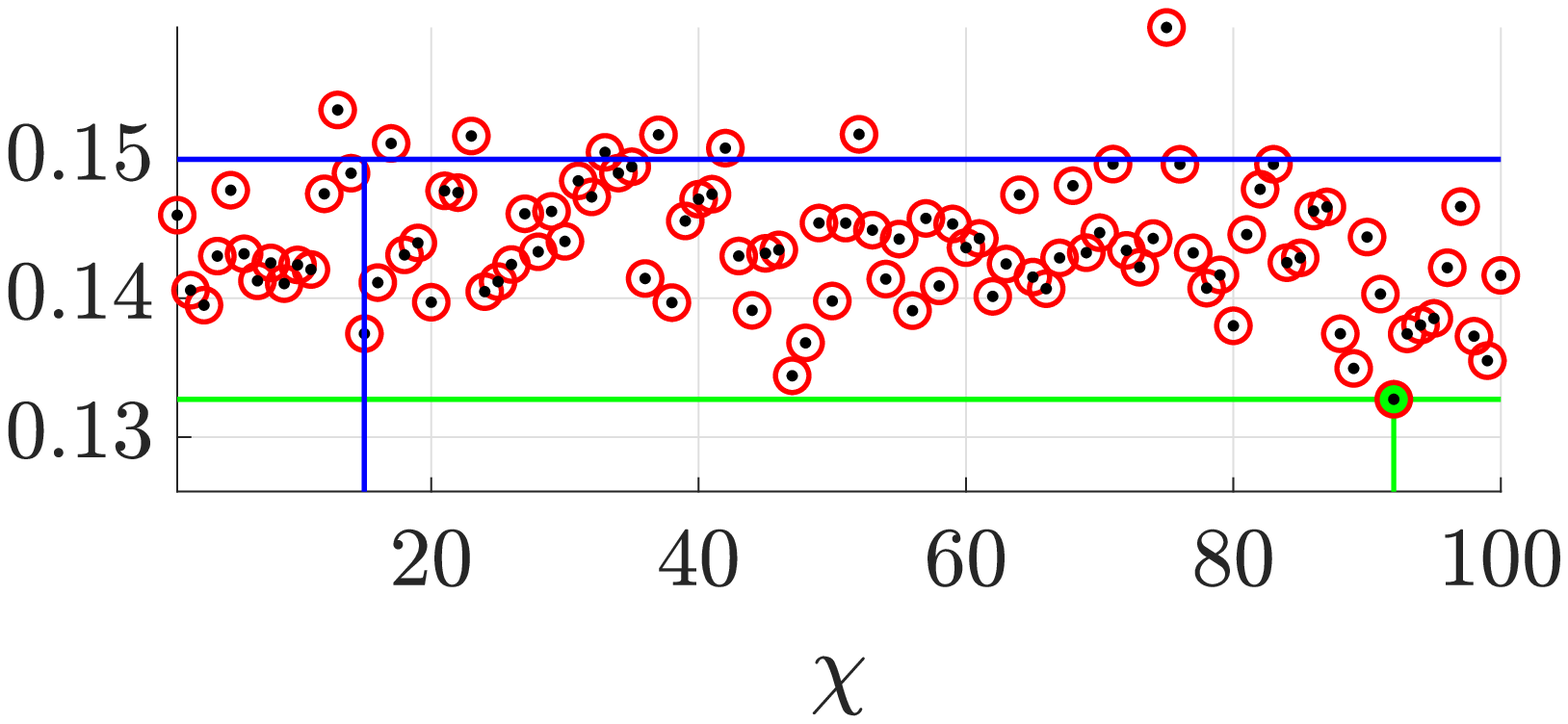}{2 View Geom. min num inliers}{white} 
    \caption{\textbf{GLOMAP Hyperparameter Ablation}. A one-dimensional brute-force search is conducted over all parameters, with the \ate\ represented on the Y-axis. Dataset: \textbf{REPLICA}; Sequence: \textbf{office0}; Number of Images: \textbf{50 / 2000}; Frame Rate: \textbf{3.06 Hz}.}
    \label{fig:brute_force_eval}
\end{figure*}

We detail below the hyperparameters used in the GLOMAP tuning experiment described in Section~\ref{subsec:finetuning_glomap}, with results shown in Figure~\ref{fig:fine_tuning_glomap} and Table~\ref{tab:glomap_ablation}. The most influential parameters were identified through an ablation study evaluating the sensitivity of accuracy to all parameters within GLOMAP, as illustrated in Figure~\ref{fig:brute_force_eval}. For further details, we refer readers to the original publications and publicly available  repositories~\cite{schoenberger2016sfm,pan2024global}. \newline

\noindent \textbf{SIFT Extraction Peak Threshold (Sift Ext. Peak)}: The parameter \textit{--SiftExtraction.peak\_threshold} (default: 0.0067) in COLMAP's feature extractor specifies the minimum contrast required to retain a keypoint. Increasing this value eliminates more low-contrast keypoints. 

\noindent \textbf{Maximum Bundle Adjustment Reprojection Error (Max. BA $\boldsymbol{e_r}$)}: The parameter \textit{--Thresholds.max\_reprojection\_error} (default: 0.01) in GLOMAP's feature mapper defines the maximum allowed reprojection error (in radians) for inliers during Bundle Adjustment.
\begin{figure}[!t]
    \centering
    \begin{subfigure}[b]{0.03\textwidth}
        \centering
        \rotatebox{90}{\parbox{4cm}{\centering \large \ate}}
    \end{subfigure}
    \begin{subfigure}[b]{0.34\textwidth}
         \centering
         \includegraphics[width=1.0\textwidth]{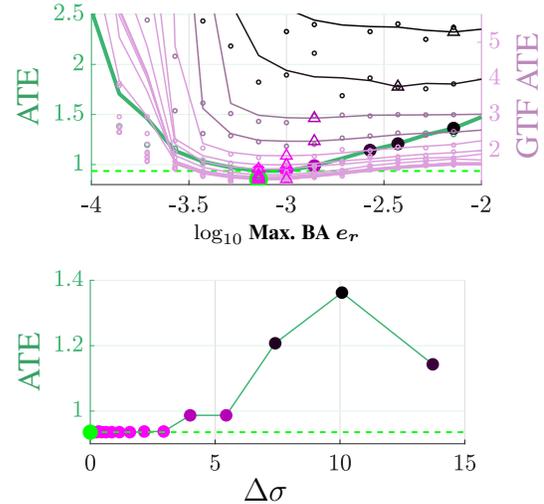} 
         \caption*{\centering \textbf{$\log_{10}$ Max. BA $\boldsymbol{e_r}$}}
    \end{subfigure}
    \begin{subfigure}[b]{0.03\textwidth}
        \centering
        \rotatebox{90}{\parbox{4cm}{\centering \large \gtf}}
    \end{subfigure}
    
    \vspace{-0.5cm}
    \hspace{-0.4cm}
    \begin{subfigure}[b]{0.03\textwidth}
        \centering
        \rotatebox{90}{\parbox{4cm}{\centering \large \ate}}
    \end{subfigure}
    \begin{subfigure}[b]{0.32\textwidth}
         \centering
         \includegraphics[width=1.0\textwidth]{figures/noise_ablation/noise_ablation_fig_2.eps} 
         \caption*{\centering \large $\Delta\sigma$}
    \end{subfigure}
    \begin{subfigure}[b]{0.03\textwidth}
        \centering
        \rotatebox{90}{\parbox{3cm}{\centering \large}}
    \end{subfigure}
    
    \caption{\textbf{Top:} The \textcolor{groundtruth}{green} line represents the \textcolor{groundtruth}{Absolute Trajectory Error (ATE)} results of GLOMAP as a function of the maximum reprojection error for inliers in the Bundle Adjustment (Max. BA $e_r$) in radians. Our proposed \textcolor{proxy}{GTF ATE} curve, shown in \textcolor{proxy}{pink}, is estimated for varying magnitudes of input Gaussian noise $\Delta\sigma$, with darker shades of pink representing larger values of $\Delta\sigma$. \textbf{Bottom:} For each curve shown in the top plot, we present the corresponding minimum ATE identified using our proposed \textcolor{proxy}{GTF ATE}.}
    \label{fig:noise_suppl}
\end{figure}
\noindent \textbf{Bundle Adjustment Huber Loss (BA Huber Loss)}: The parameter \textit{--BundleAdjustment.thres\_loss\_function} (default: 0.1) in GLOMAP's feature mapper sets the length scale for the robustification of the reprojection error (in pixels) in Bundle Adjustment, controlling the sensitivity to outliers.

\noindent \textbf{SIFT Matching Maximum Ratio (Sift Match. Max. ratio)}: The parameter \textit{--SiftMatching.max\_ratio} (default: 0.8) in COLMAP's matcher controls the maximum allowable ratio between the distances of the best and second-best matches.

\noindent \textbf{Two-View Geometry Maximum Error (2V Geo. Max. $e_r$)}: The parameter \textit{--TwoViewGeometry.max\_error} (default: 4.0) in COLMAP's matcher specifies the maximum allowable error (in pixels) for two-view geometry estimation during the initial image pair matching. 

\subsection{Hyperparameter Tuning VSLAM}
Similarly, we outline the hyperparameters used in the DROID-SLAM tuning experiment described in Section~\ref{subsec:finetuning_droid}, with results shown in Figure~\ref{fig:fine_tuning_droid} and Table~\ref{tab:droidslam_ablation}. For additional details, please refer to~\cite{teed2021droid}. \newline

\noindent \textbf{Beta}: The parameter \textit{Beta} (default: 0.3) in DROID-SLAM determines the weight assigned to the translation and rotation components of the optical flow.

\noindent \textbf{Keyframe Threshold}: The parameter \textit{keyframe\_thresh} (default: 4.0) defines the threshold  (in pixels) used to decide when a new keyframe should be created.

\noindent \textbf{Frontend Threshold}: The parameter \textit{frontend\_thresh} (default: 16.0) specifies the distance (in pixels) within which edges are added between frames in the frontend of DROID-SLAM.

\section{Input Noise Magnitude }
Figure~\ref{fig:noise_suppl} presents the complete ablation study described in Section~\ref{subsec:noise}. This study evaluates different noise magnitudes to determine the configuration that achieves the strongest correlation with real ground truth. As shown in Figure~\ref{fig:noise_suppl}, our \gtf\ effectively identifies the optimal \ate\ without requiring ground truth data within a specific range of noise magnitudes. However, beyond this range, as the noise magnitude $\Delta\sigma$ increases, the noise begins to dominate the system response, causing the \gtf\ curves to flatten.

\newpage

\end{document}